\documentclass[final,12pt]{arxiv} 

\title[Causal Evaluation via Simulation-based Inference]{Improving Generative Methods for Causal Evaluation \\via Simulation-Based Inference}
\usepackage{times}

\clearauthor{%
 \Name{Pracheta Amaranath}\footnote{Corresponding Author} \Email{pboddavarama@umass.edu}\\
 \addr College of Information and Computer Sciences, University of Massachusetts Amherst, USA
 \AND
 \Name{Vinitra Muralikrisnan}
 \Email{vmuralikrish@umass.edu}\\
 \addr College of Information and Computer Sciences, University of Massachusetts Amherst, USA
 \AND
 \Name{Amit Sharma} \Email{amshar@microsoft.com}\\
 \addr Microsoft Research 
 \AND
 \Name{David Jensen}
 \Email{jensen@cs.umass.edu}\\
 \addr College of Information and Computer Sciences, University of Massachusetts Amherst, USA
}

\usepackage{bbm}
\usepackage{caption}
\usepackage{subcaption}
\usepackage{wrapfig}
\usepackage{mathtools}
\usepackage{booktabs}
\usepackage{graphicx}
\usepackage{xcolor}
\usepackage{algorithm}
\usepackage{algorithmic}
\usepackage{tcolorbox}

\newcommand{\knob}{DGP parameter}
\newcommand{\ourmethod}{SBICE} 

\newcommand{\xM}{\mathcal{M}}

\newcommand{\bE}{\mathbb{E}}
\newcommand{\xG}{\mathcal{G}}

\begin{document}

\maketitle

\begin{abstract}%
  Generating synthetic datasets that accurately reflect real-world observational data is critical for evaluating causal estimators, but it remains a challenging task. Existing generative methods offer a solution by producing synthetic datasets anchored in the observed data (\emph{source data}) while allowing variation in key parameters such as the treatment effect and amount of confounding bias. However, it is often unclear which generative methods to use and which values of parameters to choose when generating synthetic datasets. Moreover, existing methods typically require users to provide fixed point estimates of such parameters. This denies users the ability to express uncertainty over both generative methods and parameter values and removes the potential for posterior inference, potentially leading to unreliable estimator comparisons. We introduce \emph{simulation-based inference for causal evaluation ({\ourmethod})}, a framework that treats the generative method and its corresponding generative parameters as uncertain and infers their posterior distribution given a source dataset. Leveraging techniques in simulation-based inference, {\ourmethod} identifies suitable generative methods and infers distributions over its parameter configurations to produce synthetic datasets closely aligned with the source data distribution. Empirical results demonstrate that {\ourmethod} improves the reliability of estimator evaluations by generating realistic datasets whose causal estimates closely match the estimates of the source data, making it a robust and uncertainty-aware approach to selecting causal estimators.
\end{abstract}


\section{Introduction}
\label{sec:intro}

A central challenge in causal inference is evaluating the accuracy of causal estimators. This is a difficult task due to the fundamental problem of causal inference~\citep{holland1986}---we only observe the outcome under the assigned treatment group for each unit. In contrast to associational machine learning, where ground truth labels enable cross-validation with held-out data, causal inference almost always lacks direct access to counterfactual outcomes or ground truth estimates, making such cross-validation techniques infeasible.

These challenges are exacerbated when we wish to evaluate causal estimators for a specific, real-world dataset (referred to henceforth as the \emph{source} dataset). 
\begin{wrapfigure}[27]{R}{0.46\textwidth}
    \centering
    \includegraphics[clip=true,trim={10 150 10 160},width=\linewidth]{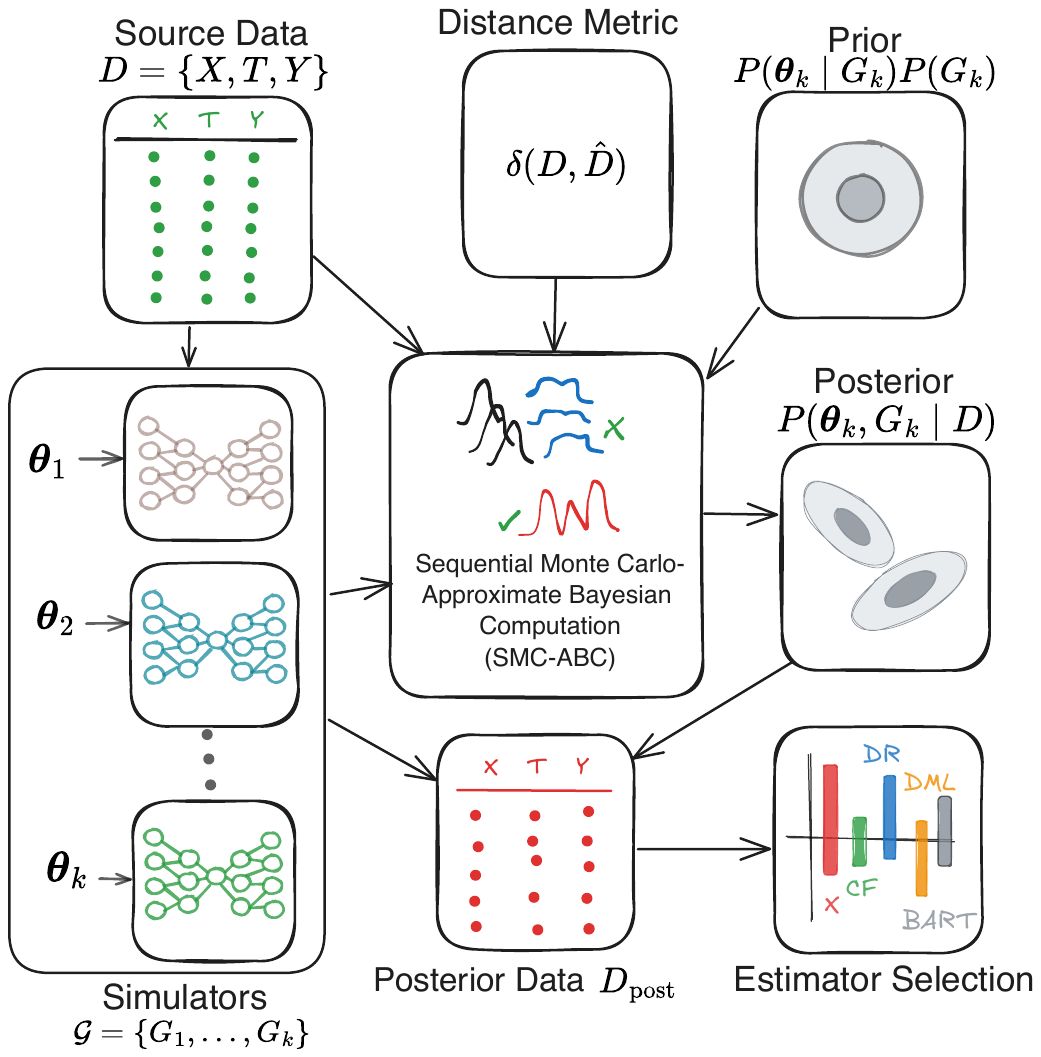}
    \caption{Workflow for simulation-based inference for causal evaluation ({\ourmethod}). Simulators corresponding to parametric/non-parametric generative models that generate data similar to the source dataset. Then, using SMC-ABC, we generate posterior datasets that are closer in distance to the source data. These datasets can then be used for reliable and robust benchmarking of causal estimators.}
    \label{fig:intro-figure}
\end{wrapfigure}
Existing approaches use synthetic and semi-synthetic datasets as benchmarks~\citep[e.g., ][]{dorie2019automated}, but they are often not designed to reflect the specific assumptions or characteristics of a given source dataset. Recent approaches use generative neural networks~\citep{parikh2022validating,neal2020realcause,de2024marginal,athey2021using} to approximate the distribution of the source data and potentially generate counterfactual outcomes and ground-truth estimates. We refer to such methods as \emph{generative methods} and focus on their use for generating synthetic datasets for causal evaluation. 

Generative methods aim to produce a distribution of datasets that correspond to: (1) what can be inferred from the source data; and (2) domain-specific prior knowledge specified via parameters that directly or indirectly influence confounding bias, selection bias, treatment propensity (overlap), unobserved confounding, and treatment heterogeneity in the generated datasets. We refer to these parameters as \emph{data generating process (DGP) parameters} henceforth, as they pertain to parameters that are specified as part of the data generating process.\footnote{To be precise, these parameters are the DGP \emph{model} parameters, as they pertain to the parameters that define the model that produces the data as opposed to the actual data generating process which may not conform to simplistic mathematical expressions} The specification of {\knob}s is central to evaluating the sensitivity of causal estimators and using a plausible range of values remains critical for robust evaluation.

However, it is often true that the distribution inferred from the source data using these generative methods is far from the true distribution or even that the {\knob}s are inconsistent with the observed distribution. Under such conditions, downstream tasks such as estimator selection and sensitivity analysis can yield biased conclusions.
These differences in the inferred distribution of the various generative methods can be attributed to a number of potential causes: the neural network architectures employed (VAEs~\citep{parikh2022validating}, GANs~\citep{athey2021using}, normalizing flows~\citep{neal2020realcause,de2024marginal}) have different convergence properties for the limited sample sizes of the source data, the factorization of the joint distribution and the set of parameters they incorporate are different. This raises an important question---which generative method should be used, and what parameter values should be chosen?

Moreover, existing generative approaches allow users to fix {\knob}s to point estimates informed by domain knowledge, from which synthetic datasets can be generated. However, these methods vary substantially in how they encode such assumptions, and how changes in {\knob}s propagate through the generation process. As we demonstrate in Section~\ref{sec:comparing-gen-methods}, incompatible choices of {\knob}s can distort the evaluation of causal estimators by inducing synthetic datasets that poorly represent the underlying source data. Moreover, most current methods do not explicitly correct for mismatches between the source data distribution and the data generated under a given {\knob} configuration, increasing the risk of drawing misleading conclusions about estimator performance. 

To address this issue, we advocate for a principled approach that incorporates Bayesian model selection to select generative methods and represent {\knob}s not as fixed values but as distributions that reflect user uncertainty. This perspective enables a more expressive form of sensitivity analysis, where identifiability of causal effects is treated as a continuum: users can encode varying degrees of uncertainty over different parts of the generative model, and some joint configurations of {\knob}s may become more or less probable given the source data.

Our approach anchors the choice of {\knob}s in the observed characteristics of the source dataset. Instead of relying on arbitrary or uniformly sampled parameter settings, our method uses simulation-based inference to estimate a posterior distribution over the generative method and the corresponding {\knob}s consistent with the source data. This posterior captures both the uncertainty in the assumptions and their compatibility with the observed distribution, allowing us to filter out implausible parameter configurations ({\knob}s that may lead to datasets that are unlike the source data, as measured qualitatively by examining the marginal/conditional distributions or quantitatively by computing a distance value). By generating datasets from this posterior rather than from a potentially uninformative prior, our approach provides more reliable sensitivity analyses and produces estimates of causal estimators closer to that of the source dataset.   
We formalize this approach in \emph{simulation-based inference for causal evaluation ({\ourmethod})\footnote{Our implementation is available at \url{https://github.com/PrachetaBA/sbi-causal-evaluation}}}, detailed in Section~\ref{sec:sbi}. {\ourmethod} encodes the user’s initial assumptions as a prior and using techniques in simulation-based inference~\citep{cranmer2020frontier} to infer a posterior over the generative method and its {\knob}s. Generative methods serve as simulators, enabling likelihood-free inference, as illustrated in Figure~\ref{fig:intro-figure}.
Importantly, {\ourmethod} acts as a wrapper around existing generative methods, enhancing their utility by grounding the generative process in data-informed posterior distributions over {\knob}s, rather than relying on fixed or arbitrary configurations. Admittedly, the improvements of {\ourmethod} rely heavily on the ability of the simulator to generate datasets that are close to the source, in the absence of which we see equivalent performance for the posterior and the prior.  

This approach offers several key advantages: (1) The generated data distribution more closely aligns to the source data distribution even in the absence of strong prior knowledge about true {\knob} values; (2) By sampling according to the posterior, our method avoids the naive assumption that all parameter values are equally plausible. In scenarios where the source dataset offers little information about {\knob}, our approach naturally falls back to a uniform prior, recovering standard sensitivity analysis practices; (3) The posterior distribution supports both marginal and joint sensitivity analyses over {\knob}; and (4) It facilitates the integration of datasets from multiple generative methods through Bayesian model selection. 
Our primary contributions are:
\begin{enumerate}
    \item \textbf{Empirical evaluation of generative methods for causal estimator selection}: We present a systematic evaluation of existing generative methods for the downstream task of causal estimator evaluation. Our study reveals that generative methods often yield inconsistent synthetic datasets, both with respect to each other and to the source dataset. This inconsistency can lead to divergent or misleading assessments of estimator performance. 
    \item \textbf{Improved estimator evaluation through posterior-weighted simulations (\ourmethod)}: We introduce {\ourmethod}, a novel simulation-based inference framework that addresses the inconsistency of generated datasets problem by learning a posterior distribution over generative methods and their corresponding {\knob} values consistent with the source data. This approach enables principled filtering of implausible parameter settings and facilitates downstream tasks such as estimator selection and robustness assessment across {\knob} variations.
\end{enumerate}

\section{Prior Work}
\label{sec:prior-work}

Early generative approaches directly estimated causal effects by approximating the counterfactual outcome distributions~\citep{yoon2018ganite,shi2019adapting}. Recent frameworks instead use generative models to evaluate causal estimators by incorporating {\knob}s that encode prior knowledge. {\ourmethod} extends this paradigm by treating {\knob}s as uncertain, enabling systematic assessment of estimator performance across plausible scenarios rather than at fixed parameter values. 

A complementary perspective comes from \textit{Bayesian causal inference}~\citep{li2023bayesian,linero2023and,hahn2020bayesian,oganisian2021practical,mccandless2007}, which propagates uncertainty directly into the causal effect estimates through explicit probabilistic models for the covariate, treatment and outcome distributions. However, this requires explicit specification of independence assumptions and parametric forms that may not scale to high-dimensional settings. {\ourmethod} instead leverages generative neural networks as flexible non-parametric representations of the data-generating process, reducing the need for such assumptions, while still enabling posterior inference.

A related body of work is the literature on source distribution estimation~\citep{vetter2024sourcerer} which identifies distributions compatible with source data by selecting the maximum-entropy solution. {\ourmethod} pursues a different objective: rather than recovering a single distribution, it extracts the full set of plausible generative processes that yield diverse yet source-consistent datasets. 
Existing generative frameworks each address part of this problem. Credence~\citep{parikh2022validating} incorporates {\knob}s but constrains them through training-time hyperparameters, preventing posterior refinement. WGANs~\citep{athey2021using} learn the source distribution flexibly but do not support controlled parameter variation for sensitivity analysis. {\ourmethod} bridges these approaches: it allows users to specify priors over multiple generative methods and parameters, infers posteriors reflecting the data fit, and supports both domain-informed simulators and data-driven generative models.



\section{Inconsistencies in Generative Methods for Causal Evaluation}
\label{sec:comparing-gen-methods}

In this section, we highlight a key obstacle to reliable causal evaluation using generative methods: the lack of principled approaches for selecting among generative methods and specifying {\knob} values. We consider source datasets with binary treatment $T$, outcome $Y$ and pre-treatment covariates $X$. The causal estimand is the average treatment effect (ATE), defined as $\tau = \mathbb{E}[Y(1) - Y(0)]$, where $Y(t)$ denotes the potential outcome under treatment $t \in [0, 1]$.

Our goal is to generate synthetic datasets that can support downstream tasks such as estimator ranking and selection, hyperparameter tuning for causal machine learning models, and sensitivity analysis. The reliability of these tasks hinges on whether the generated datasets faithfully reflect the source data distribution and whether causal estimates from synthetic data align with those from the source. Without this alignment, downstream analyses risk producing misleading conclusions about estimator performance. Our analysis addresses two central questions: (1) How well do generative methods faithfully encode prior knowledge? and (2) How accurately do they infer the probability distribution over possible datasets that align with source data and prior knowledge? 

We investigate the differences between generative methods across synthetic and observational datasets, highlighting our findings using the Lalonde~\citep{lalonde1986evaluating} dataset. This dataset is particularly well-suited for our analysis because it includes a corresponding randomized controlled trial (RCT) arm, providing access to a ground-truth ATE. We evaluate four generative methods---Credence~\citep{parikh2022validating}, a modified version of Credence with a different factorization of the joint distribution (mod-Credence), Realcause~\citep{neal2020realcause}, and FrugalFlows~\citep{de2024marginal}—--each representing distinct approaches to encoding causal assumptions and DGP parameters. Detailed descriptions of each method are included in Appendix~\ref{app:gen-methods-summary}. Additional experiments for other datasets are included in Section~\ref{app:comparing-gen-methods-expts}. 

To quantify inconsistencies across generative methods, we generate 50 synthetic datasets from each method and compute the bias of a set of causal estimators for three settings of the {\knob}s. The bias of an estimator is defined as the estimated ATE minus the ground-truth ATE. We use the following set of estimators: X-Learner~\citep{kunzel2019metalearners} with linear and gradient boosted tree (GBT) models, Double Machine Learning~\citep{chernozhukov21w} with linear and GBT models, Doubly Robust~\citep{dudik2014dr} (labeled as DR (Lin)), BART~\citep{hill2011} and TMLE~\citep{van2006targeted}.

\begin{figure}[htb]
    \centering
    \subfigure[Flexible ATE\label{fig:lalonde-obs-flexible}]{
        \includegraphics[width=0.31\textwidth,trim=0 1.1cm 0 0,clip]{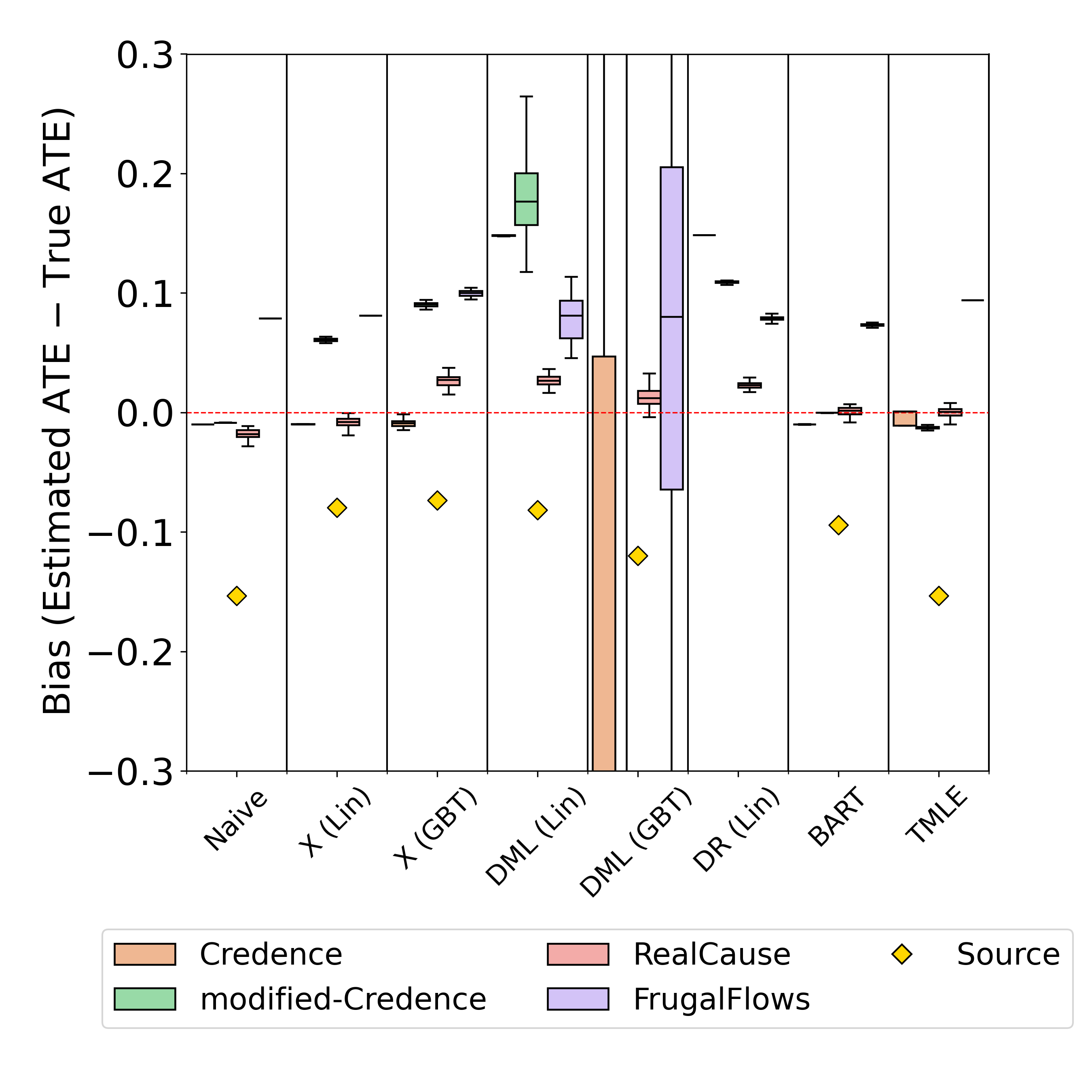}
    }
    \subfigure[True ATE\label{fig:lalonde-obs-true}]{
        \includegraphics[width=0.31\textwidth,trim=0 1.1cm 0 0,clip]{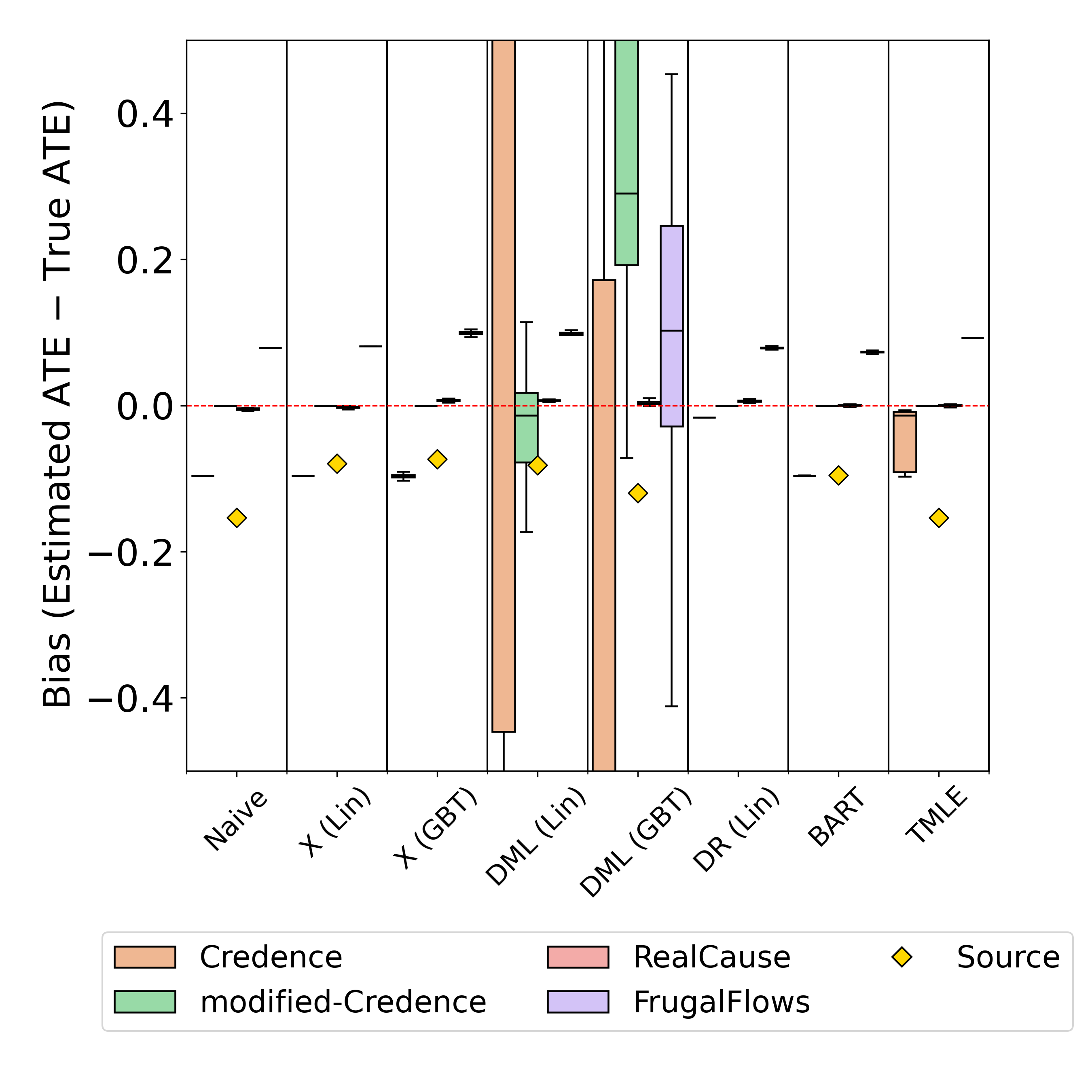}
    }
    \subfigure[Incorrect ATE\label{fig:lalonde-obs-incorrect}]{
        \includegraphics[width=0.31\textwidth,trim=0 1.1cm 0 0,clip]{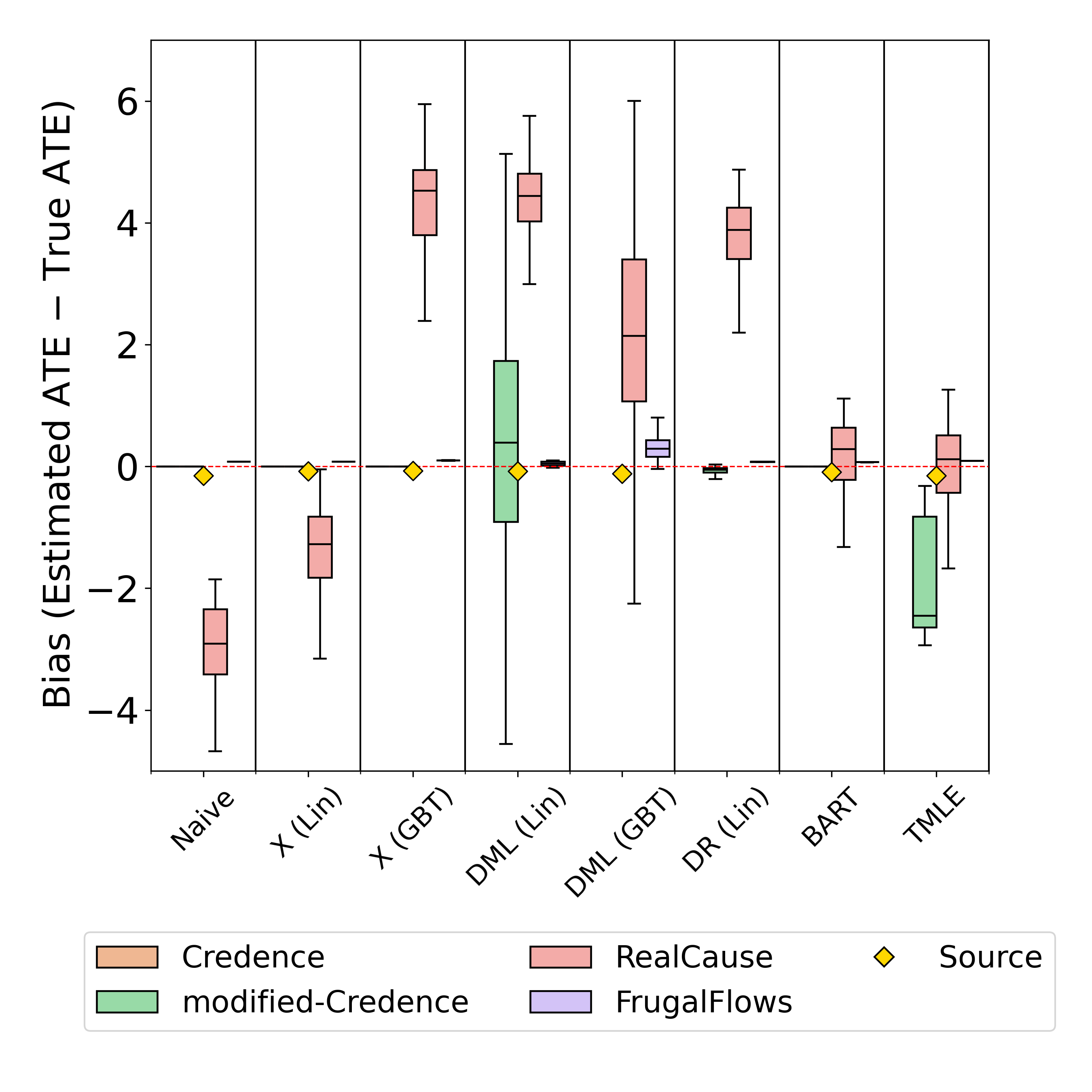}
    }
    \caption{Boxplots of estimator bias (estimated ATE $-$ true ATE) across four generative methods for the Lalonde dataset under three {\knob} setting: (a) Learned ATE, (b) True ATE, and (c) Incorrect ATE. Estimator performance varies substantially across both generative methods and parameter settings, highlighting the sensitivity of causal benchmarking to these choices.}
    \label{fig:lalonde-obs}
\end{figure}

Figure~\ref{fig:lalonde-obs-flexible} presents results for the flexible ATE setting, where the treatment effect depends on what the generative model converges to during training. 
Even without imposing constraints, we observe substantial variation in estimator bias across generative methods, indicating that different methods learn different representations of the source distribution. Figure~\ref{fig:lalonde-obs-true} shows results when the ATE parameter is set to the true value obtained from the RCT. 
Despite this informed specification, notable disagreement in estimator performance persists, suggesting that differences in how methods encode this constraint lead to divergent synthetic datasets. 
Figure~\ref{fig:lalonde-obs-incorrect} depicts the incorrect ATE setting, where the ATE is fixed to an arbitrary value (10.0) 
Here, inconsistencies are most pronounced: some methods produce large estimator biases while others yield moderate deviations, reflecting differences in how each method reconciles an incompatible constraint with the source.

\begin{table}[hbt]
\centering
\small
\begin{tabular}{@{}lcccc@{}}
\toprule
 & Credence & mod-Credence & Realcause & FrugalFlows \\ \midrule
Flexible ATE & $0.254\pm0.010$ & $0.015\pm0.001$ & $0.394\pm0.029$ & $0.573\pm0.038$ \\
True ATE & $0.254\pm0.012$ & $0.066\pm0.005$ & $0.388\pm0.028$ & $0.479\pm0.021$ \\
Incorrect ATE & $0.856\pm0.080$ & $0.876\pm0.076$ & $10.190\pm1.022$ & $0.882\pm0.071$ \\ \bottomrule
\end{tabular}
\caption{Mean sliced Wasserstein distance (and the corresponding standard deviation) between the source (Lalonde) and 50 generated datasets for each generative method and {\knob} setting.}
\label{tab:lalonde-obs-swd}
\end{table}

Beyond estimator performance, we also assess distributional alignment by computing the mean sliced-Wasserstein distance (SWD)~\citep{nadjahi2020approximate} between generated and source datasets for each generative method and parameter setting (Table~\ref{tab:lalonde-obs-swd}). Notably, the distributional distance also increases for incorrect ATE values, reflecting the tension between satisfying user-imposed constraints and maintaining fidelity to the source data. However, even methods with similar SWD values can yield divergent estimator biases as generative methods encode and enforce {\knob}s differently. For example, Credence and Realcause achieve similar SWD values in the flexible and true ATE settings (Table~\ref{tab:lalonde-obs-swd}), yet produce markedly different estimator biases (Figure~\ref{fig:lalonde-obs}).
 
Across all three settings, we find that estimator rankings are not stable across generative methods---an estimator that appears to perform well under one method may perform poorly under another. This inconsistency poses a fundamental challenge for practitioners using synthetic data for estimator selection or benchmarking, as conclusions may depend heavily on the choice of generative method and parameter specification. We hypothesize two primary causes. First, limited sample sizes may lead to different convergence properties across generative methods, causing each method to learn subtly different approximations of the source distribution. Second, when the standard assumptions of causal identifiability do not hold for the source data---a plausible scenario in many real-world settings---multiple {\knob} configurations may be consistent with the observed distribution, yet different generative methods may implicitly select different configurations. We formalize this intuition in the following proposition.

\begin{proposition}
\label{thm:inconsistent-theta}
\textbf{Incompatibility of fixed {\knob}s under generative modeling:} Let $P(D;\tau^*)$ denote the true distribution of the source data $D = \{X, T, Y\}$, for a binary treatment $T$, consistent with the {\knob}: $\tau^*$. Let $\hat{P}(D; \tau)$ represent the distribution of datasets generated under a fixed parameter value $\tau$. (i) Under assumptions of identifiability: If $\hat{P}(D; \tau) \overset{d}{=} P(D;\tau^*)$, then there exists a unique value $\tau = \tau^*$, and any constraints imposed on the values of $\tau \neq \tau^*$ results in incompatible generated data distributions: $\hat{P}(D;\tau) \not\overset{d}{=} P(D;\tau^*)$. (ii) Under violations of identifiability (e.g., due to unobserved confounding): There may exist a set of parameter values: $\mathcal{T} = \{\tau^*, \tau_1..\tau_k\}$ such that for all \(\tau \in \mathcal{T}\), \(\hat{P}(D; \tau) \overset{d}{=} P(D;\tau^*)\), while any value outside this set (or interval), i.e., \(\tau \not\in \mathcal{T}\), will produce data distributions that are not consistent with \(P(D;\tau^*)\).
\end{proposition}

\begin{proof} 
    We provide a proof sketch here and include the full proof in Appendix~\ref{app:proofs}. Under the standard assumptions of identifiability, a perfect generative method will match the true distributions $P(X)$ and $P(Y \mid T, X)$, uniquely determining $\mathbb{E}[Y\mid T,X]$ and therefore the ATE as $\tau^*$. If a strong constraint $\tau \neq \tau^*$ is enforced, the resulting conditional distribution $\hat{P}(Y \mid T,X)$ must diverge from the true distribution to satisfy the constraint. We demonstrate this by including an example of two different distributions $P_1(Y \mid T, X)$ and $P_2(Y\mid T, X)$ that have the same value of $\tau$. 
    In contrast, when identifiability is violated, even with a perfect generative method, multiple {\knob}s configurations: $\boldsymbol{\theta} = \{\tau, \rho\}$ (with $\rho$ being the unobserved confounding parameter) exist that may be consistent with the joint distribution $\hat{P}(D; \boldsymbol{\theta}) \overset{d}{=} P(D)$. 
\end{proof}

Simply put, this proposition states that there exists some {\knob}s that are inconsistent with the source data distribution. This can arise due to (i) finite-sample estimation error when learning the source data distribution even when the causal effects are identifiable, or (ii) multiple parameter values producing similar datasets due to the non-identifiability induced by unobserved confounders. This is 
compounded by how different generative methods model unobserved confounding: Credence balances constraint satisfaction and fidelity to the source data and models unobserved confounding by sampling from a learned counterfactual distribution; FrugalFlows introduces correlation between covariates and the causal effect distribution; Realcause does not model unobserved confounding, instead adjusting ATE, overlap and treatment heterogeneity post hoc. These varying approaches can produce incompatible datasets. 

These observations motivate a principled approach that treats both the generative method and {\knob}s as uncertain quantities to be inferred from the source data. Rather than relying on fixed specifications that may be inconsistent with the observed distribution, we seek a framework that identifies plausible parameter configurations and weights them by their compatibility with the source data. We develop this approach in the following section. 

\section{Simulation-Based Inference for Causal Evaluation ({\ourmethod})}
\label{sec:sbi}
We propose {\ourmethod}, a framework that models both the choice of generative method and {\knob}s as uncertain quantities to be inferred from the source data. Unlike existing approaches that require fixed specifications---leading to the inconsistencies observed in Section~\ref{sec:comparing-gen-methods}---{\ourmethod} estimates a posterior distribution over generative methods and their corresponding parameters consistent with the observed data. 
When all prior configurations are equally consistent with the source data, {\ourmethod} reduces to a uniform distribution over the prior, making it a generalization of existing practices.

Let $D = \{X, T, Y\}$ denote the source dataset, $\xG=\{G_1,\ldots,G_K\}$ a set of candidate generative methods, and $\boldsymbol{\theta}_k$ the {\knob}s associated with method $G_k$. {\ourmethod} requires the specification of four components: simulators, priors, a distance metric and a posterior inference algorithm, which we describe below. 


\paragraph{Simulators.} Each generative method $G_k \in \xG$ serves as a simulator that produces synthetic datasets $\hat{D} \sim G_k(\boldsymbol{\theta}_k)$ given {\knob} values $\boldsymbol{\theta}_k$. Simulators may be parametric, with explicit functional forms and parameters encoded as distributional parameters, or non-parametric, using generative neural networks to model the source data. Examples of non-parametric simulators include FrugalFlows and Realcause, which first learn a flexible generative model from the source data and then generate datasets conditioned on user-specified {\knob}s.

\paragraph{Priors.} The prior distribution encodes the user's beliefs about plausible generative methods and parameter values. The method prior $P(G_k)$ may be uniform across candidate methods or weighted based on domain knowledge. The parameter prior $P(\boldsymbol{\theta}_k \mid G_k)$ specifies plausible ranges for DGP parameters such as the treatment effect $\tau$, unobserved confounding $\rho$, degree of heterogeneity $\xi$, propensity of treatment $\pi$, depending on the generative method or parametric simulator. Priors may be independent or joint distributions, and can be non-informative (uniform) or informative when domain knowledge is available.

\paragraph{Distance metric.} To compare generated and source datasets without explicit likelihood computation, we use a distance metric $\delta(\hat{D},D)$ that quantifies distributional similarity. We employ the sliced-Wasserstein distance~\citep{nadjahi2020approximate} due to its computational efficiency and suitability for high-dimensional, mixed-type data. The distance metric guides parameter proposals during inference and determines which configurations are retained. Alternatively, a composite metric disentangling covariate, treatment, and outcome distributions~\citep{amad2025improving} may offer a more structured basis for measuring fidelity to the source dataset.

\paragraph{Posterior inference via SMC-ABC.} Our trained simulators can be viewed as neural likelihood estimators (NLEs): once fitted to the source data, it implicitly defines a likelihood $P(D \mid G_k, \boldsymbol{\theta}_k)$ over datasets. Following standard practice in simulation-based inference~\citep{cranmer2020frontier}, we pair these NLEs with Sequential Monte Carlo Approximate Bayesian Computation (SMC-ABC)~\citep{toni2010simulation} to approximate the posterior. While neural posterior estimation (NPE)~\citep{lueckmann2021benchmarking} offers an alternative by directly learning the posterior, it poses challenges in our setting: NPE requires training an additional neural network on limited source data, risking compounded misspecification, and struggle with small sample sizes and mixed discrete-continuous data. SMC-ABC avoids these issues while offering flexibility through user-defined distance metrics. It also naturally flags model misspecification---if no simulator can generate data close to the source distribution, the distance metric remains high. We acknowledge that SMC-ABC can be computationally intensive for higher dimensional datasets and sensitive to the choice of the distance metric. We leave the incorporation of NPE to {\ourmethod} to future work. We include a computational complexity analysis of {\ourmethod} with SMC-ABC in Appendix~\ref{app:computational-complexity-sbice}.

We now provide an overview of how the components combine to form the {\ourmethod} algorithm. Our goal is to infer a joint posterior distribution over generative methods and their parameters:
\begin{equation}
    P(G_k, \boldsymbol{\theta}_k \mid D) \propto P(D \mid G_k, \boldsymbol{\theta}_k) \, P(\boldsymbol{\theta}_k \mid G_k) \, P(G_k)
\end{equation}
Since the likelihood $P(D \mid G_k, \boldsymbol{\theta}_k)$ is typically intractable for complex generative models, we employ likelihood-free inference via SMC-ABC. The algorithm proceeds in three phases and is outlined in detailed in Appendix~\ref{app:sbice-algorithm}. This process is also visually demonstrated in Figure~\ref{fig:intro-figure}. 

\paragraph{Phase 1: Train simulators.} Each generative method $G_k \in \xG$ is trained on the source dataset $D$. For non-parametric simulators, this involves fitting a generative model to approximate $P(X,T,Y)$; for parametric simulators, this involves estimating parameters of pre-specified functional forms. Once trained, each simulator can generate synthetic datasets $\hat{D} \sim G_k(\boldsymbol{\theta}_k)$. 

\paragraph{Phase 2: SMC-ABC Posterior Inference.} We initialize $N$ samples $(G^{(i)}, \boldsymbol{\theta}^{(i)})$ by sampling from the prior: $P(G_k) P(\boldsymbol{\theta}_k,G_k)$. The algorithm iterates through decreasing tolerances $\epsilon_1 > \epsilon_2 > \cdots > \epsilon_T$, where at each iteration samples are perturbed, synthetic datasets are generated and compared to the source via the distance metric, and samples with $\delta< \epsilon_t$ are retained. This concentrates posterior mass on configurations that generate data similar to the source.

\paragraph{Phase 3: Generate posterior datasets and compute causal estimates.} We sample configurations from the posterior and generate datasets $\hat{D}^{(i)}_{\text{post}} \sim G^{(i)}(\boldsymbol{\theta}^{(i)})$ from the selected and retained samples (posterior distribution). For each dataset $i$ and causal estimator $\xM_j$, we compute causal estimates $\tau_{\xM_j}^{(i)} = \xM_j(\hat{D}^{(i)}_{\text{post}})$. 

The posterior distribution enables several downstream applications. First, it provides
\textbf{principled model selection}: methods that consistently generate datasets close to the source receive higher posterior weight while others are down-weighted or excluded, addressing the inconsistency problem identified in Section~\ref{sec:comparing-gen-methods}. Second, it supports \textbf{sensitivity analysis} by allowing users to evaluate how causal estimates vary across plausible DGP configurations, focusing effort on parameter regions consistent with the observed data. Third, it enables \textbf{robust estimator evaluation}: causal estimators assessed on posterior-weighted datasets receive more reliable performance assessments than under arbitrary parameter settings. When multiple estimators exhibit similar posterior performance distributions, {\ourmethod} recommends confidence-aware evaluation rather than deterministic ranking. 

\section{Experiments}
\label{sec:sbi-emp-expts}

We evaluate {\ourmethod} on its ability to generate posterior-weighted datasets useful for downstream causal benchmarking. Our evaluation focuses on a key criterion: consistency of causal estimator performance across generated and source datasets, measured using the \emph{bias squared error (BSE)}.

\paragraph{Bias squared error (BSE).} The BSE quantifies whether a causal estimator exhibits similar performance on a generated dataset as it does on the source dataset. Intuitively, if the generated dataset faithfully represents the source, an estimator should produce comparable errors on both. For a causal estimator $\xM$, let $\tau_{\xM}$ denote its ATE estimate and $\tau^*$ the true ATE (which is the specific {\knob} for a generated dataset or the ground-truth value for a source dataset). The estimator's bias is $\tau_{\xM} - \tau^*$. We compute this bias on both the generated dataset $\hat{D}^{(i)}$ and the source dataset $D$, then measure their squared difference:
\begin{equation}
    \text{BSE}_{\xM; d}^{(i)} = \left[ \underbrace{\left( \tau_{\xM; d}^{(i)} - \tau_{d}^{*(i)} \right)}_{\text{bias on generated data}} - \underbrace{\left( \tau_{\xM;\text{source}} - \tau^* \right)}_{\text{bias on source data}} \right]^2, \quad d \in \{\text{post}, \text{prior}\}
\end{equation}
where the subscript $d$ indicates whether the generated dataset is drawn from the posterior ($\boldsymbol{\theta}_{\text{post}}$) or prior ($\boldsymbol{\theta}_{\text{prior}}$) distribution. A BSE of zero indicates that the estimator exhibits identical bias on the generated and source datasets---the ideal outcome for reliable benchmarking.

\paragraph{Aggregating BSE across datasets.} Since {\ourmethod} produces a distribution over DGP parameters, we generate $N$ datasets from both the posterior ($\boldsymbol{\theta}_{\text{post}}$) and prior ($\boldsymbol{\theta}_{\text{prior}}$) distributions of parameter values and aggregate BSE values to obtain a summary measure. We report the mean BSE:
\begin{equation}
    \text{Mean BSE}_{\xM; d} = \frac{1}{N} \sum_{i=1}^{N} \text{BSE}_{\xM; d}^{(i)}, \quad d \in \{\text{post}, \text{prior}\}
\end{equation}

The mean can be replaced by other summary statistics (e.g., median or quantiles) depending on the evaluation objective. For synthetic source datasets with multiple draws, we compute dataset-specific comparisons using $\tau_{\xM;\text{source}}^{(i)}$. A lower mean BSE for the posterior ($\text{mean BSE}_{\xM; \text{post}} \leq \text{mean BSE}_{\xM; \text{prior}}$) indicates that posterior-weighted datasets yield estimator performance more consistent with the source data when compared to the prior-weighted datasets. 

The BSE can be interpreted as a task-specific posterior predictive check~\citep{gelman1996posterior}. Standard posterior predictive checks assess model fit by comparing summary statistics of observed and simulated data. BSE applies this principle to causal benchmarking, comparing estimator bias on generated datasets to that on the source. A low (mean) BSE indicates that the posterior generates datasets on which causal estimators behave similarly to the source---precisely the property required for reliable benchmarking.

We evaluate {\ourmethod} across a suite of synthetic datasets and multiple classes of simulators—both parametric and non-parametric. Our empirical study is designed to answer the following questions: (1) How can {\ourmethod} improve upon existing generative methods for benchmarking causal estimators for a given source dataset? (2) Under what conditions does the estimated posterior distribution over {\knob}s provide meaningful insights for benchmarking causal estimators?

\paragraph{Datasets} We evaluate {\ourmethod} using synthetic and real-world datasets. These datasets are categorized into three types, referred to as LinearParam, Frugal and Real DGPs (data-generating processes). LinearParam DGPs use linear, parametric data-generating functions with a single observed covariate $X$, an unobserved covariate $Z$, binary treatment $T$ and outcome $Y$. They are mainly used to study the effects of simulator misspecification on {\ourmethod}. Frugal DGP  datasets are based on the synthetic datasets designed using the frugal parameterization~\citep{evans2024parameterizing} method~\cite[see][Appendix D]{de2024marginal}. These datasets introduce more complexity, with 4-10 covariates, non-linear functions and model unobserved confounding. Real DGP datasets are commonly benchmarks in the causal inference literature, limited by sample size and often lacking a ground-truth ATE. We focus on two types of real datasets: (1) with an experimental arm, that can be used to compute the ground-truth ATE, or (2) derived from real-world systems where it is possible to intervene on the treatment and observe both potential outcomes. For the second type of dataset, we sample treatments using importance sampling to create observational datasets~\citep{gentzel2021and}.   

\paragraph{Simulators} Our experiments use three simulator types: LinearParam simulators (based on parametric, linear models), FrugalParam simulators (based on frugal parameterization), FrugalFlows and Realcause (non-parametric generative methods). 
An extensive outline of the results for each dataset and simulator type is provided in Appendices~\ref{app:sbice-parametric-dgps},~\ref{app:sbice-frugal-dgps}, and~\ref{app:sbice-real-dgps}.

\subsection{Results}
\label{sec:sbi-results}
\paragraph{{\ourmethod} produces informative posteriors and enables principled model selection when simulators are sufficiently flexible to approximate the joint distribution of the source data, and the source data is informative of the {\knob}s} When multiple generative methods are available, {\ourmethod} automatically identifies which methods best approximate the source distribution by assigning higher posterior weight to methods that generate data consistent with the source. Table~\ref{tab:mean-bse-real} demonstrates this for three real-world datasets---Lalonde, Postgres, and Twins---where we evaluated two candidate simulators: Realcause and FrugalFlows. In all three cases, {\ourmethod} sampled datasets exclusively from Realcause based on the empirical fit (lower distance metric values), and estimated a posterior distribution over its corresponding {\knob}s. The resulting posterior-weighted datasets yield substantially lower BSE than the prior across nearly all causal estimators, indicating that estimator performance on posterior datasets closely matches performance on the source data. Additional details are included in Appendix~\ref{app:sbice-real-dgps}.  In addition, to verify that the results produced by {\ourmethod} is not overly sensitive to the choice of the distance metric, we re-ran some experiments with an additional distance metric, maximum mean discrepancy (MMD), which compares probability distributions by computing the distance between embeddings in an high-dimensional space, and found  that the learned posterior distributions were similar to those produced when using the sliced-Wasserstein distance. These results are listed in Appendix~\ref{app:sbice-mmd}. 

\begin{figure}[htb]
    \centering
    \subfigure[Dataset: LinearParam DGP1, Simulator: LinearParam Sim1\label{fig:parametric-linear}]{
        \includegraphics[width=0.3\textwidth]{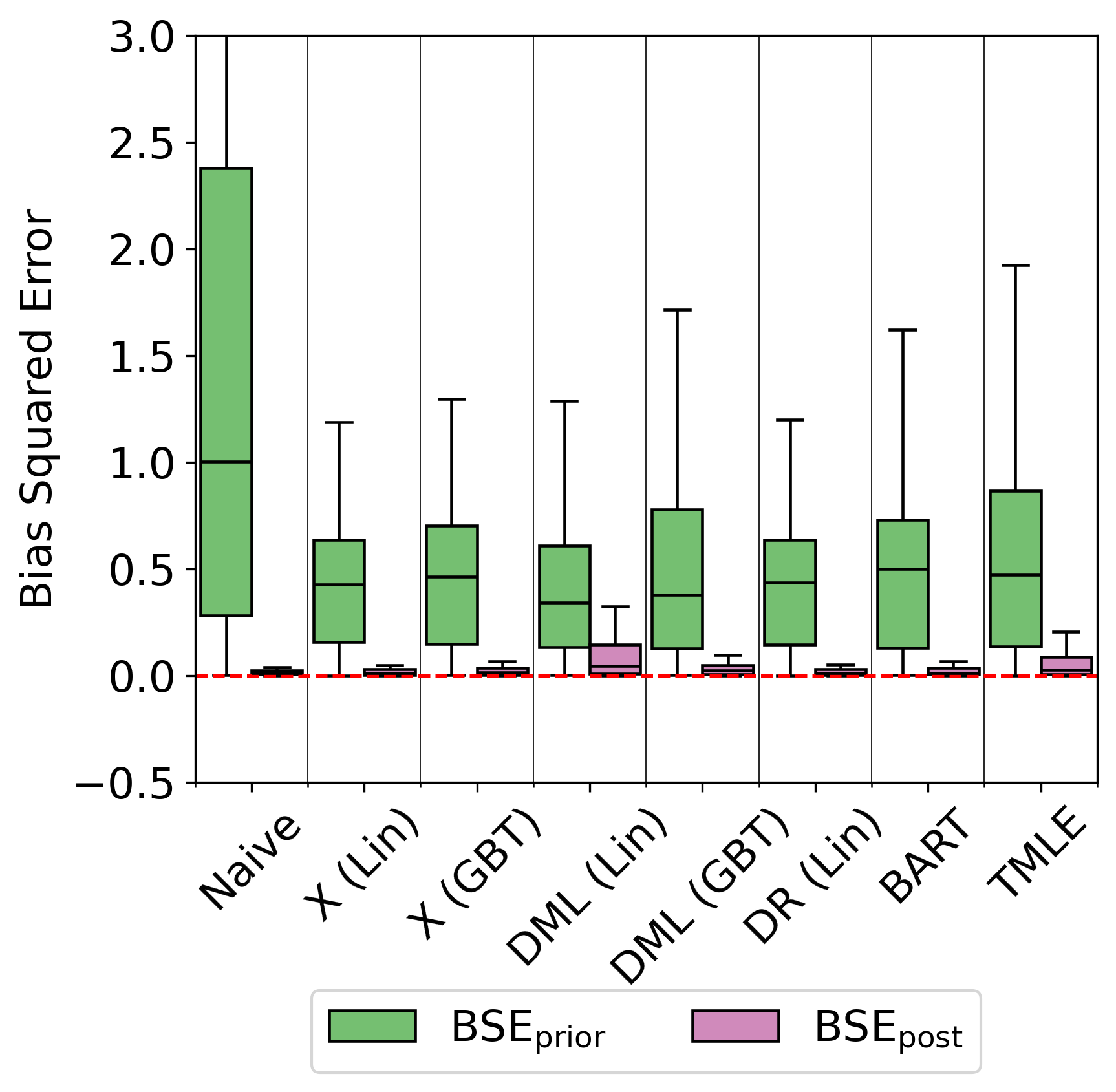}
    }
    \hspace{0.1em}
    \subfigure[Dataset: Frugal DGP4, Simulator: FrugalParam Sim4(u)\label{fig:frugal-param-syndata}]{
        \includegraphics[width=0.3\textwidth]{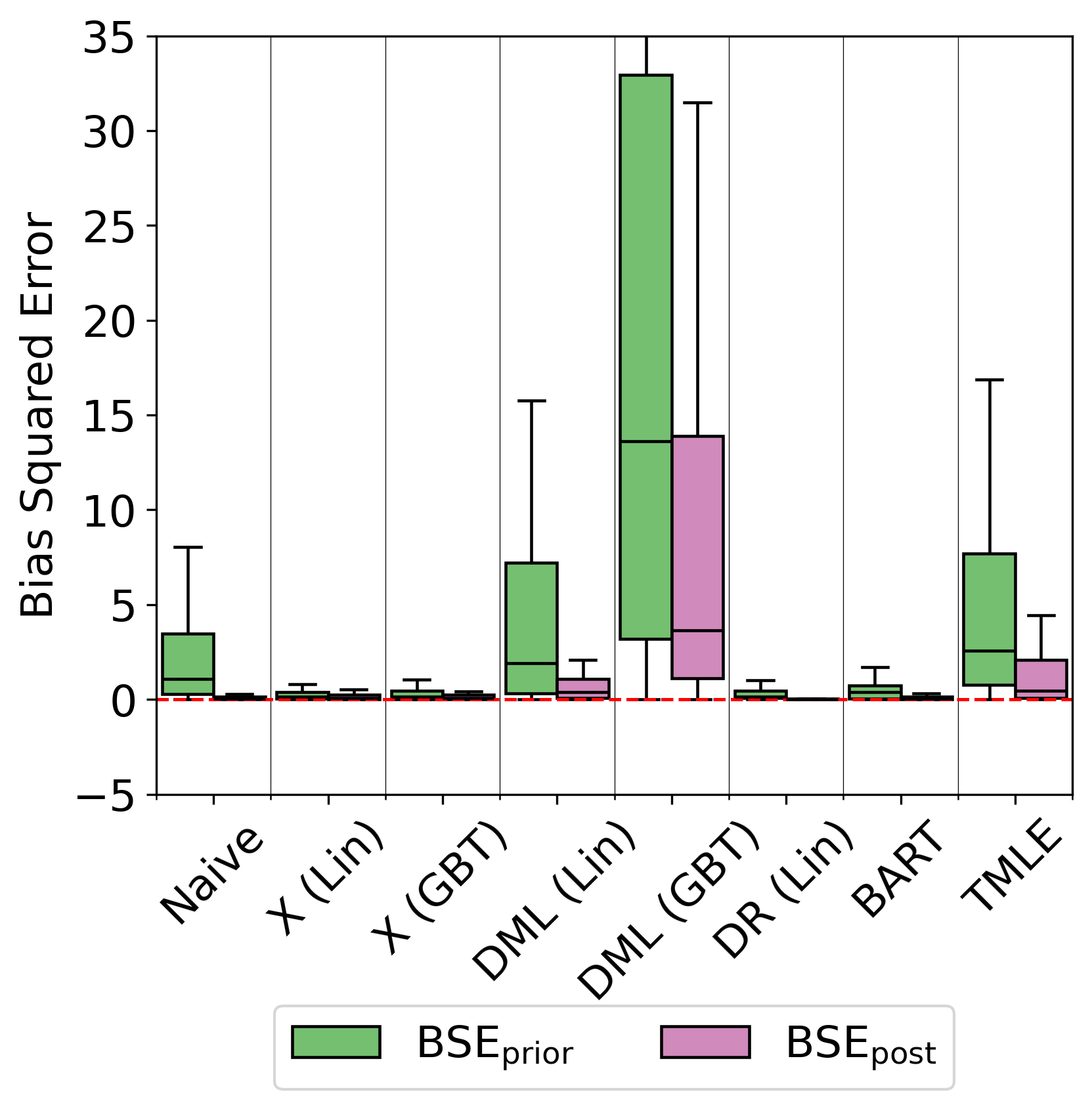}
    }
    \hspace{0.1em}
    \subfigure[Dataset: Frugal DGP4, Simulator: FrugalFlows Sim4(u)\label{fig:frugal-flows-syndata}]{
        \includegraphics[width=0.3\textwidth]{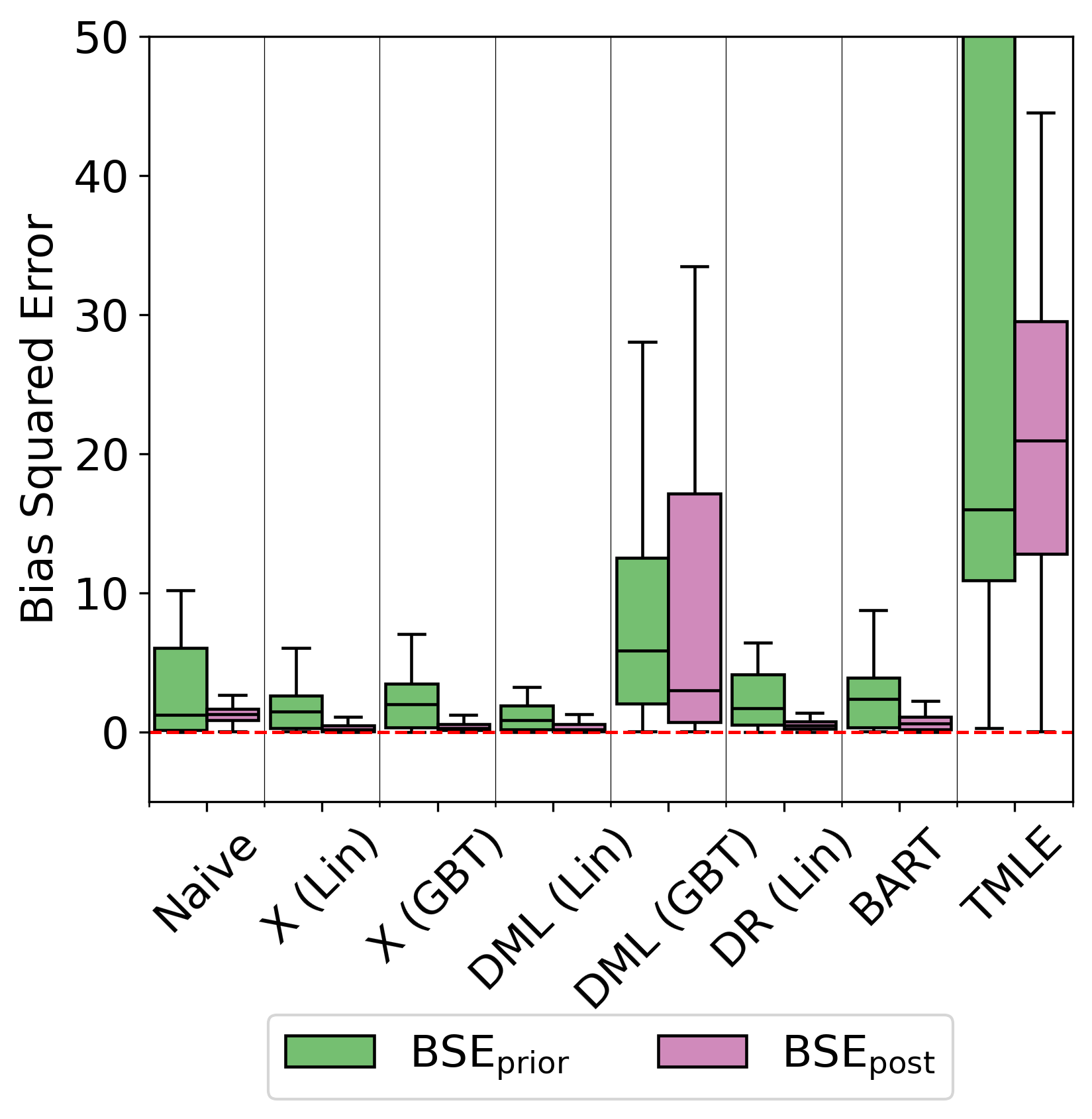}
    }
    \caption{BSE for a set of causal estimators for the datasets in Table~\ref{tab:mean-bse-synthetic}. Lower BSE values for the posterior indicate similarity to the source dataset.} 
    \label{fig:bse-plot-synthetic}
\vspace{-4mm}
\end{figure}

We also evaluate {\ourmethod} on synthetic datasets where the data-generating process is known (Table~\ref{tab:mean-bse-synthetic}, Figure~\ref{fig:bse-plot-synthetic}). When the generated datasets closely reflect the true {\knob}s and the simulator is well-specified, the $\text{mean BSE}_{\text{post}} < \text{mean BSE}_{\text{prior}}$ for all causal estimators. To assess the robustness of {\ourmethod}, we systematically introduced model misspecification---such as noise, non-linear functional forms, and changes in parameter identifiability or prior informativeness. We found that, as long as the misspecified simulator could still generate datasets resembling the source, the posterior estimates remained informative and outperformed fixed {\knob} baselines (see Appendices~\ref{app:sbice-frugal-dgps} and~\ref{app:sbice-parametric-dgps} for details). 

\begin{table}[thb]
\centering
\small
\begin{tabular}{@{}lcccccc@{}}
\toprule
DGP & \multicolumn{2}{c}{LinearParam DGP1} & \multicolumn{2}{c}{Frugal DGP4} & \multicolumn{2}{c}{Frugal DGP4} \\ 
Simulator & \multicolumn{2}{c}{LinearParam Sim1} & \multicolumn{2}{c}{FrugalParam Sim4(u)} & \multicolumn{2}{c}{FrugalFlows Sim4(u)} \\ 
DGP Param(s)  & \multicolumn{2}{c}{$\boldsymbol{\theta} = \{\beta, \tau, \rho\}$} & \multicolumn{2}{c}{$\boldsymbol{\theta} = \{\tau, \rho\}$} & \multicolumn{2}{c}{$\boldsymbol{\theta} = \{\tau, \rho\}$} \\ 
\midrule
 & Prior & Posterior & Prior & Posterior & Prior & Posterior \\ \midrule
\textbf{SWD} & $0.52\pm0.22$ &$0.09\pm0.02$ &$4.84\pm2.38$ &$2.26\pm0.08$ & $3.33\pm1.75$ & $0.30\pm0.03$ \\ \midrule 
\textbf{Mean BSE} & & & & & & \\
X (Lin) & $0.574$ & $0.015$ & $0.277$ & $0.196$ & $1.970$ & $0.315$\\
X (GBT) & $0.586$ & $0.019$ & $0.304$ & $0.219$ & $2.187$ & $0.414$ \\
DML (Lin) & $0.550$ & $0.082$ & $5.144$ & $0.711$ & $1.225$ & $0.438$ \\
DML (GBT) & $0.564$ & $0.028$ & $34.767$ & $10.531 $& $16.568$ & $12.743$ \\
DR (Lin) & $8\text{e}5$ & $0.016$ & $0.374 $& $0.040 $& $1\text{e}6$ & $0.628$ \\
BART & $0.590$ & $0.018$ & $0.585$ & $0.103$ & $2.447$ & $0.708$ \\
TMLE & $0.624$ & $0.064$ & $4.809$ & $2.554$ & $48.568$ & $21.787$ \\
\bottomrule
\end{tabular}
\caption{Mean BSE scores and sliced Wasserstein distances (mean $\pm$ standard deviation) for three synthetic datasets using different types of DGPs and simulators.}
\label{tab:mean-bse-synthetic}
\end{table}

\begin{table}[htb]
\centering
\small
\begin{tabular}{@{}lcccccc@{}}
\toprule
DGP & \multicolumn{2}{c}{Lalonde} & \multicolumn{2}{c}{Postgres} & \multicolumn{2}{c}{Twins} \\ 
Simulator(s) & \multicolumn{2}{c}{Realcause Lalonde,} & \multicolumn{2}{c}{Realcause Postgres,} & \multicolumn{2}{c}{Realcause Twins} \\ 
& \multicolumn{2}{c}{FrugalFlows Lalonde} & \multicolumn{2}{c}{FrugalFlows Postgres} & \multicolumn{2}{c}{FrugalFlows Twins} \\ 
DGP Param(s)  & \multicolumn{2}{c}{$\boldsymbol{\theta} = \{\tau, \xi, \pi, \rho \}$} & \multicolumn{2}{c}{$\boldsymbol{\theta} = \{\tau, \xi, \pi, \rho \}$} & \multicolumn{2}{c}{$\boldsymbol{\theta} = \{\tau, \xi\}$} \\ 
\midrule
 & Prior & Posterior & Prior & Posterior & Prior & Posterior \\ \midrule
\textbf{SWD} & $1.87\pm1.19$ &$0.16\pm0.01$ &$1.62\pm1.03$ &$0.02\pm0.001$ & $1.01\pm0.62$ & $0.01\pm0.00$ \\ \midrule 
\textbf{Mean BSE} & & & & & & \\
X (Lin) & $3.520$ & $0.016$ & $9.130$ & $3.23\text{e-}4$ & $0.044$ & $0.003$\\
X (GBT) & $8.580$ & $0.017$ & $10.055$ & $3.20\text{e-}4$ & $0.063$ & $0.001$\\
DML (Lin) & $8.332$ & $0.022$ & $7.827$ & $3.56\text{e-}4$ & $0.039$& $0.002$\\
DML (GBT) & $6.193$ & $0.046$ & $9.844$ & $5.77\text{e-}4$& $0.048$& $0.001$\\
DR (Lin) & $4\text{e}3$ & $4\text{e}3$ & $1\text{e}7$& $590 $& $0.062$ & $0.002$\\
BART & $4.631$ & $0.017$ & $8.738$ & $3.23\text{e-}4$ & - & - \\
TMLE & $4.291$ & $0.035$ & $27.045$ & $2.18\text{e-}3$ & - & - \\
\bottomrule
\end{tabular}
\caption{Mean BSE and SWD (mean $\pm$ standard deviation) for three real datasets using Realcause and FrugalFlows simulators. 
For the Twins dataset, we find that the default setting of both the BART and TMLE estimators results in high errors, due to which we omit computing their BSE. }
\label{tab:mean-bse-real}
\vspace{-5mm}
\end{table}

\paragraph{Prior and posterior BSE values converge when estimators are insensitive to {\knob}, the source dataset is non-informative, or the simulator poorly approximates the source distribution}
{\ourmethod} is most effective when the source data provides a clear signal about the {\knob} values and at least one simulator can accurately approximate the source distribution. However, when these conditions are not met, or the priors are too narrow or misaligned with the true values, the posterior and prior can yield similar estimator performance, highlighting the need for flexible simulators and appropriately broad priors.

Table~\ref{tab:sbice-limitations-results} presents three illustrative cases. In Frugal DGP3 and DGP5, a narrow prior centered on the true treatment effect $\tau^*$ lead to nearly identical mean BSEs for both posterior and prior datasets. For the Lalonde dataset with the FrugalFlows simulator, we found that the source dataset was not informative of the {\knob} values, leading to identical SWD and BSE values for both the posterior and the prior. This demonstrates that when no candidate simulator fits the source data well, {\ourmethod} appropriately reverts to the prior rather than producing misleading posteriors---a useful diagnostic indicating that alternative simulators should be considered.

\begin{table}[htb]
\centering
\small
\begin{tabular}{@{}lcccccc@{}}
\toprule
Dataset & \multicolumn{2}{c}{Frugal DGP3} & \multicolumn{2}{c}{Frugal DGP5} & \multicolumn{2}{c}{Lalonde} \\ 
Simulator & \multicolumn{2}{c}{FrugalFlows Sim3} & \multicolumn{2}{c}{FrugalFlows Sim5} & \multicolumn{2}{c}{FrugalFlows Lalonde}\\
DGP Param(s)  & \multicolumn{2}{c}{$\boldsymbol{\theta} = \{\tau\}$} & \multicolumn{2}{c}{$\boldsymbol{\theta} = \{\tau\}$} & \multicolumn{2}{c}{$\boldsymbol{\theta} = \{\tau\}$} \\ 
\midrule
 & Prior & Posterior & Prior & Posterior & Prior & Posterior \\ \midrule
\textbf{SWD} & $1.18\pm0.74$ &$0.20\pm0.03$ & $2.68\pm1.48$& $0.32\pm0.03$ & $0.28\pm0.05$ & $0.29\pm0.08$ \\ \midrule 
\textbf{Mean BSE} & & & & & & \\
X (Lin) & $0.001$ & $0.001$ & $0.064$ & $0.064$ & $0.805$ & $0.805$ \\
X (GBT) & $0.002$ & $0.002$ & $0.040$ & $0.040$ & $0.843$ & $0.845$ \\
DML (Lin)& $0.002$ & $0.002$ & $\mathbf{0.319}$ & $\mathbf{0.178}$ & $2.622$ & $2.310$ \\
DML (GBT) & $0.003$ & $0.003$ & $\mathbf{2.172}$ & $\mathbf{1.876}$ & $248.36$ & $254.512$\\
DR (Lin) & $0.001$ & $0.001$ & $1249.36 $& $1249.37$ & $\approx 2\text{e}7$ & $\approx 2\text{e}7$\\
BART & $0.001$ & $0.001$ & $0.052$ & $0.052$ & $1.158$ & $1.158$\\
TMLE & $0.011$ & $0.011$ & $0.790$ & $0.789$ & $2.198$ & $2.445$\\ 
\bottomrule
\end{tabular}
\caption{Mean BSE and SWD (mean $\pm$ standard deviation) values for two synthetic datasets: Frugal DGP3 and Frugal DGP5 and a real-world dataset: Lalonde (Obs). For the synthetic data, $\text{BSE}_{\xM;\text{post}} = \text{BSE}_{\xM; \text{prior}}$ for nearly all the causal estimators $\xM$ with the exception of the DML estimator (\textbf{in bold}).}
\label{tab:sbice-limitations-results}
\vspace{-4mm} 
\end{table}

\section{Discussion}
\label{sec:discussion}

{\ourmethod} is a principled, practically deployable extension of existing generative approaches for causal evaluation. Under the assumption that generative methods are appropriate for constructing synthetic datasets, {\ourmethod} is strictly more informative: it enables explicit uncertainty quantification over {\knob}s and mitigates inconsistencies from fixed, point-estimated specifications, operating as a lightweight wrapper around existing generative models, as demonstrated in Section~\ref{sec:sbi-emp-expts}.
It is not restricted to neural generative models: any parametric or simulation-based family can be used---and if all candidate generators are equally compatible with the source data, the posterior reduces to the prior, ensuring {\ourmethod} does not degrade relative to existing methods.

We recommend {\ourmethod} in settings where SCMs follow a known functional structure, such as data from sensors, physical systems (e.g., the Postgres dataset), or datacenters, where the simulator can be parameterized to closely reflect the true DGP. When no such structure is available, {\ourmethod} remains applicable with neural simulators, as our experiments demonstrate. Existing simulators can also be replaced by tabular generative methods that do not explicitly model causal structure~\citep{ctgan,hollmann2025accurate,tabddpm}, which may be preferable when there is no prior signal on appropriate {\knob}s for the source data.

\section{Conclusion}
\label{sec:conclusion}
We addressed the problem of generating datasets that resemble a real-world dataset focusing on approaches that use non-parametric neural network-based simulators. We observed that existing methods often misrepresent the source data distribution under user-specified {\knob}s and lack explicit mechanisms to assess this mismatch. To address this, we introduced {\ourmethod}, a principled framework that incorporates uncertainty over {\knob}s through prior distributions and generates synthetic datasets that are better aligned with the source distribution, reducing the risk of biased or misleading conclusions for downstream tasks.
By simultaneously inferring  both the generative model and its parameters, we enable principled model selection across generative methods, producing richer, more diverse benchmark datasets for evaluating causal estimators.

\acks{We thank Purva Pruthi for her contributions on the early version of this paper. We also thank Harsh Parikh and Daniel de Vassimon Manela for their helpful discussions on the existing generative methods. We thank Andrew Zane, Sankaran Vaidyanathan, Kate Avery and Riddho R. Haque and the anonymous reviewers for their comments and suggestions which helped improve the paper. Pracheta Amaranath was supported by the Dissertation Writing Fellowship from Manning College of Information and Computer Sciences during Fall 2024. The computational resources for this work were provided by the Unity Research Computing Platform, a multi-institutional cluster lead by University of Massachusetts Amherst, the University of Rhode Island, and University of Massachusetts Dartmouth.}

\bibliography{references}

\appendix
\section{Inconsistency of arbitrary point estimates with source data distributions} 
\label{app:proofs}

In this section, we prove Proposition~\ref{thm:inconsistent-theta} for two different sets of assumptions on the source data: (i) under the standard assumptions of unconfoundedness, positivity and consistency; and (ii) under the assumption that there are hidden confounders in the data.

\begin{itemize}
    \item[(i)] In this setting, we assume that the standard causal identification assumptions hold: unconfoundedness (no unmeasured confounding), positivity ($0 < P(T|X) < 1$), and consistency. We prove the proposition in three steps: (1) we establish that the ATE is a deterministic functional of the observed distribution; (2) we show that matching the joint distribution implies matching the ATE; and (3) we demonstrate that fixing the ATE does not uniquely determine the conditional distributions, explaining why generative methods can produce inconsistent datasets even when constrained to the same parameter value. 

    \begin{proof}
    
    \textbf{Step 1:} The average treatment effect (ATE) is identified from the observed distribution. 
    The ATE $\tau^*$ for the source distribution is given by: 
        \[ \tau^* = \bE[Y(1) - Y(0)] \] 
    Using linearity of expectation, 
        \[ = \bE[Y(1)] - \bE[Y(0)]\]
    Applying the law of iterated expectations, 
        \[ = \bE_X[\bE[Y(1)|X] - \bE[Y(0)|X]]\] 
    By the assumptions of unconfoundedness, consistency and positivity, we have $\bE[Y(t)|X] = \bE[Y|T=t, X]$. Therefore, 
        \begin{equation}
        \label{eq:tau-identified}
        \tau^* = \bE_X[\bE[Y|T=1,X] - \bE[Y|T=0,X]] 
        \end{equation}
        \[ = \sum_x \left[ \left[ \sum_y P(y|t = 1,x) \right]  - \left[ \sum_y P(y|t = 0,x)\right] \right] \cdot P{(x)} \]

    This expression shows that $\tau^*$ is a deterministic functional of $P(X)$ and $P(Y\mid T, X)$, which are the components of the true data distribution $P(D)$. 

    \textbf{Step 2:} Matching the joint distribution implies a unique ATE. 

    \textit{Forward direction.} Suppose $\hat{P}(D; \tau) \equiv P(D;\tau^*)$, meaning the joint distributions are identical: $\hat{P}(X, T, Y) = P(X, T, Y)$ for all $(X, T, Y)$. By the laws of probability, the joint uniquely determines all marginals and conditionals, that is, $\hat{P}(X) \equiv P(X), \hat{P}(T\mid X) \equiv P(T \mid X), \hat{P}(Y \mid T, X) \equiv P(Y \mid T, X)$. For the generated data distribution $\hat{P}$, we have 

    \[
        \tau = \mathbb{E}_{\hat{P}(X)}\left[\mathbb{E}_{\hat{P}(Y|T,X)}[Y|T=1,X] - \mathbb{E}_{\hat{P}(Y|T,X)}[Y|T=0,X]\right] \\
    \]
    \[ =  \mathbb{E}_{P(X)}\left[\mathbb{E}_{P(Y|T,X)}[Y|T=1,X] - \mathbb{E}_{P(Y|T,X)}[Y|T=0,X]\right] \]
    \[ = \tau^* \]

    \textit{Contrapositive.} Suppose $\tau \neq \tau^*$. If a generative method enforces this constraint while generating data, then either $\hat{P}(X) \neq P(X)$ or $\hat{P}(Y|T,X) \neq P(Y|T,X)$ (or both). This is because any pair of distributions $(\hat{P}(X), \hat{P}(Y|T,X))$ that matches the source would necessarily yield $\tau^*$ via Equation~\eqref{eq:tau-identified}, contradicting the constraint $\tau \neq \tau^*$. Therefore, $\hat{P}(D; \tau) \not\equiv P(D;\tau^*)$.

    \textbf{Step 3:} Non-uniqueness of conditional distributions for fixed $\tau$.
    While Steps 1--2 establish that matching $P(D)$ implies a unique $\tau$, the converse does not hold: fixing $\tau$ does not uniquely determine $P(Y \mid T, X)$. Multiple conditional distributions can yield the same ATE. To see this, observe that the conditional average treatment effect (CATE) is:
    \begin{equation}
        \tau(X) = \mathbb{E}[Y \mid T = 1, X] - \mathbb{E}[Y \mid T = 0, X]
    \end{equation}
    
    and the ATE is its expectation over $X$:
    \begin{equation}
        \tau = \mathbb{E}_X[\tau(X)]
    \end{equation}
    
    Consider modifying the outcome by adding a function $h(X)$ that depends only on $X$:
    \begin{equation}
        \tilde{Y} = Y + h(X)
    \end{equation}
    
    The CATE under this modification is:
    \begin{align}
        \tilde{\tau}(X) &= \mathbb{E}[\tilde{Y} \mid T = 1, X] - \mathbb{E}[\tilde{Y} \mid T = 0, X] \\
        &= \mathbb{E}[Y + h(X) \mid T = 1, X] - \mathbb{E}[Y + h(X) \mid T = 0, X] \\
        &= \mathbb{E}[Y \mid T = 1, X] + h(X) - \mathbb{E}[Y \mid T = 0, X] - h(X) \\
        &= \tau(X)
    \end{align}
    
    The $h(X)$ term cancels because it does not depend on $T$. Therefore $\tilde{\tau}(X) = \tau(X)$ for all $X$, and consequently $\tilde{\tau} = \tau$. However, if $h(X)$ is not identically zero, the conditional distributions differ:
    \begin{equation}
        P(\tilde{Y} \mid T, X) \neq P(Y \mid T, X)
    \end{equation}
    
    This demonstrates that infinitely many conditional distributions $P(Y \mid T, X)$ are consistent with any given ATE value $\tau$. 
    \paragraph{Example.}
    
    We illustrate this with two concrete data-generating processes $C_1, C_2$ that share the same ATE but have different conditional outcome distributions.
    
    DGP $C_1$:
    \begin{align}
        X &\sim \mathcal{N}(0, 1) \\
        T &\sim \text{Bernoulli}(\sigma(X)) \\
        Y_1 &= X + T + \epsilon, \quad \epsilon \sim \mathcal{N}(0, 1)
    \end{align}
    
    DGP $C_2$: 
    \begin{align}
        X &\sim \mathcal{N}(0, 1) \\
        T &\sim \text{Bernoulli}(\sigma(X)) \\
        Y_2 &= X + T + (X^2 - 1) + \epsilon, \quad \epsilon \sim \mathcal{N}(0, 1)
    \end{align}
    
    Here, $h(X) = X^2 - 1$, which has mean zero since $\mathbb{E}[X^2] = 1$ for $X \sim \mathcal{N}(0, 1)$.
    
    ATE for $C_1$:
    \begin{align}
        \tau_1 = \mathbb{E}_X[\tau_1(X)] &= \mathbb{E}[Y_1 \mid T = 1, X] - \mathbb{E}[Y_1 \mid T = 0, X] \\
        &= (X + 1) - (X + 0) \\
        &= \mathbb{E}_X[1]
    \end{align}
    
    ATE for $C_2$:
    \begin{align}
        \tau_2 = \mathbb{E}_X[\tau_2(X)] &= \mathbb{E}[Y_2 \mid T = 1, X] - \mathbb{E}[Y_2 \mid T = 0, X] \\
        &= (X + 1 + X^2 - 1) - (X + 0 + X^2 - 1) \\
        &= \mathbb{E}_X[1]
    \end{align}
    
    However their conditional distributions $P(Y \mid T, X)$ differ, 
    \begin{align}
        P_{C_1}(Y \mid T, X) &= \mathcal{N}(X + T, \, 1) \\
        P_{C_2}(Y \mid T, X) &= \mathcal{N}(X + T + X^2 - 1, \, 1)
    \end{align}

    In practice, the learned conditional data distributions may not differ from each other drastically, but in cases when there is a conflict between the user-specified parameter value and the possible conditional distributions that satisfy that assignment, we may see the effects of such variation. 
    \end{proof}

    \item[(ii)] We now consider the setting where the causal effect is non-identifiable from the source data due to the presence of unobserved confounders. It is well-established in the causality literature that the treatment effect is not identifiable from the observed data in the presence of unobserved confounders~\citep{pearl2009causality,shpitser2006identification,neal2021introduction}, which implies that there are multiple values of $\tau$ that are consistent with the observed data distribution. For completeness, we outline this proof here. Note that the only assumption we make here is that the outcome is bounded to some finite value, without which the treatment effect can be any real value. 
    
    Essentially, in the presence of unobserved confounders, the potential outcomes $Y(t)$ are not independent of the treatment given the observed covariates, then, there exists an interval $\mathcal{T} = [\tau_{\ell}, \tau_{u}]$ such that for all $\tau \in \mathcal{T}$, there is a valid distribution over potential outcomes that is (i) consistent with the observed distribution, and (ii) has treatment effect in the interval $\mathcal{T}$. For any $\tau \notin \mathcal{T}$, no such distribution exists. Consequently, a generative method can produce $\hat{P}(D;\tau) \equiv P(D)$ for any $\tau \in \mathcal{T}$, while any $\tau \notin \mathcal{T}$ necessarily yields $\hat{P}(D;\tau) \not\equiv P(D)$. 

    \begin{proof}
        We prove this in two steps: (i) we derive bounds on the ATE and show that it is the interval $\mathcal{T}$, (ii) we show that within this interval, any $\tau$ admits a valid distribution consistent with the observed data. 

        \textbf{Step 1:} Bounds on the ATE
        The ATE is defined as 
        \[ \tau = \mathbb{E}[Y(1) - Y(0)]\]
        \[ = \mathbb{E}_X[\mathbb{E}[Y(1) \mid X] - \mathbb{E}[Y(0) \mid X]]\]
        We can decompose this into the observed and counterfactual distributions as follows
        \[ = \mathbb{E}[Y(1)\mid T= 1,X]P(T=1\mid X) + \mathbb{E}[Y(1)\mid T = 0,X]P(T = 0\mid X) \]
        \[ - \mathbb{E}[Y(0)\mid T = 1, X]P(T=1\mid X) - \mathbb{E}[Y(0)\mid T = 0, X]P(T=0\mid X)\]
        By consistency 
        \[ \mathbb{E}[Y \mid T= 1,X]P(T=1\mid X) + \mathbb{E}[Y(1)\mid T = 0,X]P(T = 0\mid X) \]
        \[ - \mathbb{E}[Y(0)\mid T = 1, X]P(T=1\mid X) - \mathbb{E}[Y\mid T = 0, X]P(T=0\mid X) \]
        Let us define $\pi(X) = P(T=1\mid X)$, the observed treated outcome $\mu_1(X) = \mathbb{E}[Y\mid T = 1,X]$, the observed control outcome $\mu_0(X) = \mathbb{E}[Y \mid T = 0,X]$. Let $\mu_{10}(X) = \mathbb{E}[Y(1) \mid T = 0,X]$ and $\mu_{01}(X) = \mathbb{E}[Y(0) \mid T = 1,X]$ as the counterfactual treated and control outcomes. 
        The ATE can be written as 
        \[ = \mathbb{E}_X[\pi(X)\{\mu_1(X)-\mu_{01}(X)\} + (1-\pi(X))\{ \mu_{10}(X)-\mu_0(X)\}\]
        Rearranging to separate the observed and counterfactual components, we have 
        \[ = \mathbb{E}_X[\mu_1(X) - \mu_0(X)] + \mathbb{E}_X[(1 - \pi(X))\{\mu_{10}(X) - \mu_1(X)\} + \pi(X)\{\mu_0(X) - \mu_{01}(X)\}] \]
        \[ = \tau_{\text{obs}} + \text{confounding bias}\]
        Under unobserved confounding $Y(t) \not\perp T \mid X$, we have that at least one of these equalities fail, i.e., 
        \[ \mu_{10}(X) \neq \mu_1(X) \] or 
        \[ \mu_{01}(X) \neq \mu_0(X) \]
        Crucially, the observed data $P(D)$ provides no information about these differences---they depend on the unobserved joint distribution of potential outcomes.

        Without additional assumptions, the counterfactual expectations $\mu_{10}(X)$ and $\mu_{01}(X)$ are constrained only by the support of $Y$. Assume $Y$ has bounded support: $Y \in [y_{\min}, y_{\max}]$.\footnote{If $Y$ is unbounded, then $\tau_{\min} \to -\infty$ and $\tau_{\max} \to +\infty$, so $\mathcal{T} = (-\infty, +\infty)$. This vacuous bound motivates parametric sensitivity analysis, which places structural assumptions on the confounding mechanism to obtain tighter bounds.}

        For each $X$:
        \begin{equation}
            y_{\min} \leq \mu_{10}(X) \leq y_{\max} \quad \text{and} \quad y_{\min} \leq \mu_{01}(X) \leq y_{\max}
        \end{equation}
        
        \textit{Lower bound on $\tau$:} To minimize $\tau$, set $\mu_{10}(X) = y_{\min}$ and $\mu_{01}(X) = y_{\max}$ for all $X$:
        \begin{equation}
            \tau_{\ell} = \mathbb{E}_X\left[\pi(X)\{\mu_1(X) - y_{\max}\} + (1-\pi(X))\{y_{\min} - \mu_0(X)\}\right]
        \end{equation}
        
        \textit{Upper bound on $\tau$:} To maximize $\tau$, set $\mu_{10}(X) = y_{\max}$ and $\mu_{01}(X) = y_{\min}$ for all $X$:
        \begin{equation}
            \tau_{u} = \mathbb{E}_X\left[\pi(X)\{\mu_1(X) - y_{\min}\} + (1-\pi(X))\{y_{\max} - \mu_0(X)\}\right]
        \end{equation}
        
        The set of consistent ATEs is therefore:
        \begin{equation}
            \mathcal{T} = [\tau_{\min}, \tau_{\max}]
        \end{equation}
        
        This interval contains all ATE values consistent with the observed distribution $P(X, T, Y)$ under any valid specification of the counterfactual distributions.
        
        \paragraph{Step 2: Constructing valid distributions for any $\tau \in \mathcal{T}$.}
        
        We show that for any $\tau \in \mathcal{T}$, there exists a valid joint distribution over potential outcomes consistent with the observed data.
        
        For a target $\tau \in [\tau_{\ell}, \tau_{u}]$, define:
        \begin{equation}
            \alpha = \frac{\tau - \tau_{\ell}}{\tau_{u} - \tau_{\ell}} \in [0, 1]
        \end{equation}
        
        For each $X$, set:
        \begin{align}
            \mu_{10}(X) &= (1-\alpha) \cdot y_{\min} + \alpha \cdot y_{\max} \\
            \mu_{01}(X) &= (1-\alpha) \cdot y_{\max} + \alpha \cdot y_{\min}
        \end{align}
        
        By construction, $\alpha = 0$ yields $\tau = \tau_{\ell}$, $\alpha = 1$ yields $\tau = \tau_{u}$, and intermediate values yield intermediate ATEs.
        
        Note the following is true.
        \begin{enumerate}
            \item By construction, $\mu_{10}(X), \mu_{01}(X) \in [y_{\min}, y_{\max}]$ for all $X$.
            
            \item A valid distribution exists: Given any mean $\mu \in [y_{\min}, y_{\max}]$, valid probability distributions on $[y_{\min}, y_{\max}]$ with that mean exist. 
            
            \item \textbf{Consistency with observed data:} The construction only specifies the counterfactual distributions $P(Y(1) \mid T=0, X)$ and $P(Y(0) \mid T=1, X)$. The observed distributions remain unchanged, preserving $P(X, T, Y)$.
            
            \item \textbf{Consistency axiom satisfied:} Under any realization, $Y = T \cdot Y(1) + (1-T) \cdot Y(0)$, matching the observed outcome by construction.
        \end{enumerate}
        
        Therefore, every $\tau \in \mathcal{T}$ admits a valid potential outcomes distribution consistent with $P(D)$. Conversely, for $\tau \notin \mathcal{T}$, the required counterfactual means would fall outside $[y_{\min}, y_{\max}]$, violating the support constraint. No valid distribution exists, and any generative method constrained to such a $\tau$ must distort the observed marginals.
        \end{proof}
        
        The key implications of this are
        \begin{enumerate}
            \item \textit{Set of multiple $\tau$ values: } The set $\mathcal{T}$ contains all ATE values consistent with $P(X, T, Y)$. The observed data alone cannot distinguish between them. 
            
            \item \textit{Generative methods implicitly resolve counterfactual ambiguity.} When learning from $P(X, T, Y)$, each method must implicitly specify the counterfactual quantities $\mu_{10}(X)$ and $\mu_{01}(X)$. Different methods---with different architectures, inductive biases, and training procedures---resolve this ambiguity differently, selecting different points in $\mathcal{T}$. This explains why different generative methods produce different ATE estimates from identical source data.
            
            \item \textit{User-specified ATEs outside $\mathcal{T}$ produce inconsistent data.} For $\tau \notin \mathcal{T}$, a generative method must distort the observed marginals to satisfy the constraint, producing $\hat{P}(D; \tau) \not\equiv P(D)$.
            
            \item \textit{The width of $\mathcal{T}$ quantifies fundamental causal uncertainty.} The interval $[\tau_{\min}, \tau_{\max}]$ reflects inherent ambiguity given only observational data. Narrower intervals require additional assumptions about confounding.
        \end{enumerate}
        
        This motivates {\ourmethod}: rather than arbitrarily selecting a single $\tau \in \mathcal{T}$, we infer a posterior distribution over consistent values, enabling principled uncertainty quantification in causal benchmarking. 
    
\end{itemize}

\section{Generative methods}
\label{app:gen-methods-summary}

In this section, we summarize the generative methods used this paper: Credence~\citep{parikh2022validating}, Realcause~\citep{neal2020realcause}, and FrugalFlows~\citep{de2024marginal}. While all aim to generate synthetic datasets that closely resemble the source data distribution, they differ in how they model the joint distribution $P(X, T, Y)$, the architectures they use, their training objectives, and the user-defined {\knob} they support. We outline these differences below.

\paragraph{Credence} Credence models the joint distribution as $P(X, T, Y) = P(T) \cdot P(X \mid T) \cdot P(Y \mid X, T)$ using conditional variational autoencoders (cVAEs). cVAEs encode the data into a latent space modeled by a user-defined distribution (normal or uniform) and recreates the data by sampling from a decoder that takes as input the latent variable as well as the variable being conditioned on. For binary treatment $T$, $P(T)$ is modeled as a Bernoulli distribution estimated from data, while $P(X \mid T)$ and $P(Y \mid X, T)$ are learned via the cVAE framework.

Credence supports two {\knob}s: treatment effect and selection bias (also representing unobserved confounding bias), which users define as functions of the covariates and treatment. These functions are incorporated into the loss through KL divergence to measure how closely the generated data matches the source, along with two regularization terms that enforce the specified {\knob}s, each controlled by a hyperparameter. This approach offers both benefits and drawbacks. On the one hand, the data guides the strength of enforcement: low regularization weights signal that the specified constraint is incompatible with the source data. On the other hand, users must manually specify multiple constraint values and retrain the model each time, making the process computationally expensive and less scalable. 

\paragraph{modified-Credence} We implement a variant of Credence that retains its original architecture (and set of implemented {\knob}s) but changes the factorization of the joint distribution. We factorize the data distribution as: $P(X, T, Y) = P(X) \cdot P(T \mid X) \cdot P(Y\mid T, X)$. This modification allows us to directly use the empirical distribution of covariates from the source data, avoiding the need to learn $P(X)$ from scratch. However, it increases the complexity of modeling $P(T\mid X)$ especially with high-dimensional covariates. In practice, we found this factorization often produced datasets that more closely resembled the source distribution, though not perfectly. 

\paragraph{Realcause} Realcause models the joint distribution of the source data using the same factorization as our modified Credence: $P(X, T, Y) = P(X) \cdot P(T \mid X) \cdot P(Y\mid T, X)$. It uses separate MLP or flow networks to model each marginal and conditional distribution. This structure preserves the empirical covariate distribution from the source data. Unlike Credence, Realcause adopts the TARNET~\cite{shalit2017estimatingindividualtreatmenteffect} architecture by learning two distinct models for the outcome: $P(Y \mid T = 1, X), P(Y \mid T = 0, X)$  which helps maintain treatment relevance in high-dimensional settings where the model might otherwise ignore $T$.

Realcause supports three user-defined {\knob}s: positivity (overlap), treatment effect heterogeneity, and the scale of the causal effect. These {\knob}s are applied as post-hoc transformations to the generated data, allowing a single trained model to be reused for multiple configurations by adjusting parameters that shift or scale the output. However, Realcause does not support the simulation of unobserved confounding.

\paragraph{FrugalFlows} FrugalFlows factorizes the joint distribution as the product of variationally independent distributions: $ P(T,X) \cdot P(Y \mid \text{do}(T)) \cdot \phi(Y,X)$. Here, $P(T, X)$ is the distribution of the `past', $P(Y \mid \text{do}(T))$ is the causal effect distribution and the $\phi(Y, X)$ encodes the dependence (or correlation) between outcomes and covariates. The model learns each component using normalizing flows and enforces their dependence using copulas. Unlike other generative methods, FrugalFlows generates only causal outcomes $Y$ and does not simulate counterfactuals.

FrugalFlows supports three {\knob}s: causal effect scale, unobserved confounding, and treatment effect heterogeneity. It incorporates these either through priors or post-hoc transformations, depending on the specific {\knob}. This approach offers stronger enforcement of constraints since they directly shape the data generation process. However, overly strict specification can lead to distortions that push the generated data away from the source distribution.

\section{Comparing generative methods across fixed point estimates of {\knob}s}
\label{app:comparing-gen-methods-expts}
This section expands on the results presented in Section~\ref{sec:comparing-gen-methods}, providing a detailed evaluation of estimator bias across datasets generated by different generative methods. We conduct this analysis across a range of source datasets, from simple synthetic data-generating processes to real-world observational data. We focus on the flexible {\knob} specification, where parameters are learned by the generative model due to the computational cost of training Credence and modified Credence for every {\knob} value. For each experiment, we also report the mean sliced-Wasserstein distance between generated and source datasets. 

\paragraph{Summary of key findings.} Our analysis reveals two consistent patterns. First, no single generative method performs best across all datasets---a method that achieves low distributional distance and estimator bias on one dataset may perform poorly on another. This underscores the importance of careful generative method selection for each source dataset. Second, when {\knob}s are incorrectly specified, the impact varies substantially across generative methods. Some methods maintain reasonable fidelity to the source distribution under misspecified constraints, while others diverge significantly. These findings motivate the need for a principled approach to both method and parameter selection, which we address with {\ourmethod}. The remainder of this section presents detailed results for each dataset. We first discuss synthetic datasets (Section~\ref{sec:comparing-gen-synthetic}), followed by real-world datasets (Section~\ref{sec:comparing-gen-real}).

\subsection{Synthetic datasets}
\label{sec:comparing-gen-synthetic}

\subsubsection{LinearParam DGP11}   

We design a synthetic dataset with three covariates $X_1..X_3$, a binary treatment $T$ and outcome $Y$ generated as follows.
\begin{equation}
    \begin{aligned}
        X_1 \sim \mathcal{N}(0, 1) \\
        X_2 \sim \text{Exponential}(\lambda = 0.5) \\
        X_3 \sim \mathcal{N}(1,1) \\
        T \sim \text{Binomial}\left(\text{Expit}\left(\frac{X_1 + X_2 + X_3}{3} + \mathcal{N}(0, 0.1)\right)\right) \\
        Y \sim \mathcal{N}(\mu = X_1 + X_2 + X_3 + 3T, 0.1)
    \end{aligned}
\end{equation}

We generate $9000$ samples, and remove rows exhibiting positivity violations, yielding approximately $8000$ samples. The ground-truth ATE for this dataset is $\tau^*=3.0$. We train all generative methods under three settings
\begin{enumerate}
    \item Flexible ATE: No constraints are imposed; the generative method learns the ATE from the source data. 
    \item True ATE: The ATE is constrained to the ground-truth value $\tau = 3.0$. 
    \item Incorrect ATE: The ATE is constrained to an incorrect value $\tau = 10.0$. 
\end{enumerate}

\paragraph{Distributional similarity.} Table~\ref{tab:swd-lp-dgp11} reports the mean sliced-Wasserstein distance between generated and source datasets for each method and setting. Distances increase substantially under the incorrect ATE constraint for all generative methods reflecting the tension between satisfying user-imposed constraints and maintaining fidelity to the source distribution.

\begin{table}[hbt]
\centering
\small
\begin{tabular}{@{}lcccc@{}}
\toprule
 & Credence & mod-Credence & Realcause & FrugalFlows \\ \midrule
Flexible ATE & $0.417\pm0.031$ & $0.462\pm0.029$ & $0.452\pm0.023$ & $0.414\pm0.030$ \\
True ATE & $0.624\pm0.032$ & $0.564\pm0.034$ & $0.454\pm0.026$ & $0.414\pm0.028$ \\
Incorrect ATE & $1.348\pm0.106$ & $1.386\pm0.105$ & $2.396\pm0.217$ & $2.048\pm0.182$ \\ 
\bottomrule
\end{tabular}
\caption{Mean sliced-Wasserstein distance ($\pm$ standard deviation) between generated and source datasets for LinearParam DGP11 across generative methods and ATE settings.}
\label{tab:swd-lp-dgp11}
\end{table}

\begin{figure}
    \centering
    \includegraphics[width=0.8\linewidth]{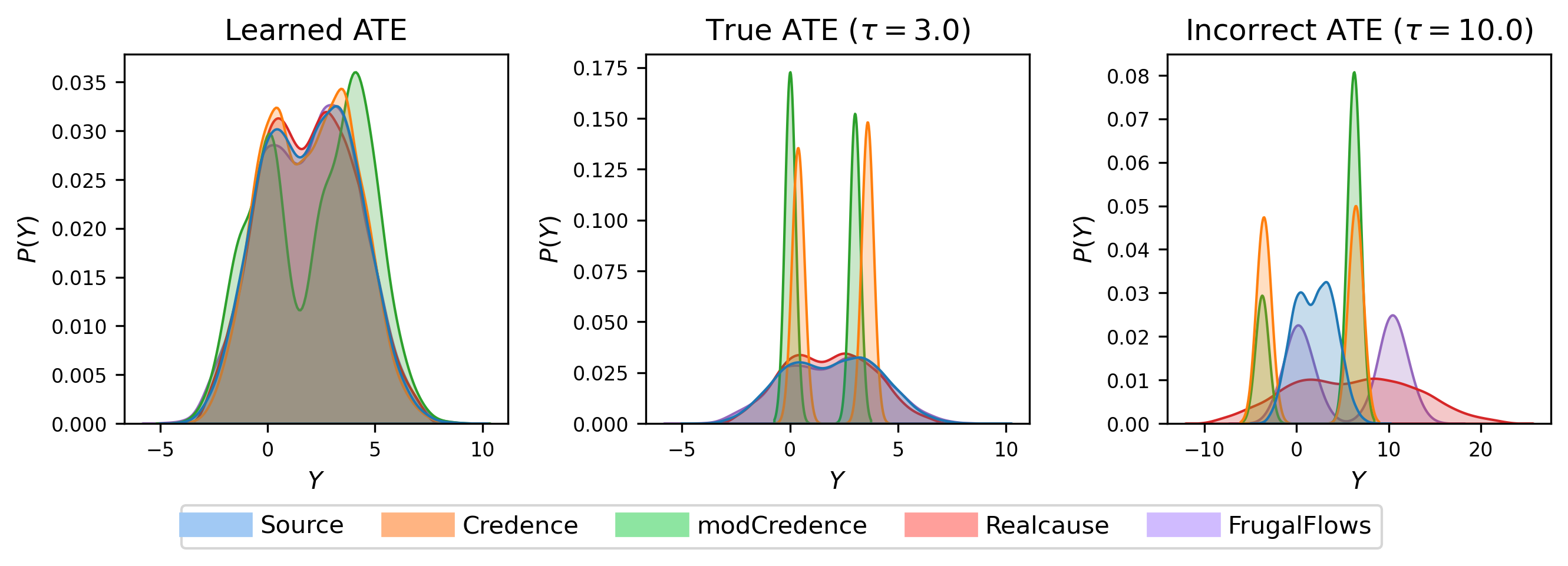}
    \caption{Marginal distribution of outcome $Y$ for LinearParam DGP11. }
    \label{fig:outcome-lp-dgp11}
\end{figure}

\paragraph{Marginal distributions.} Figure~\ref{fig:outcome-lp-dgp11} displays the marginal distribution of outcome $Y$ for a single generated dataset across all three settings. While most methods produce similar marginal distributions under the learned ATE setting, imposing constraints affects Credence and mod-Credence most dramatically. This difference arises from how constraints are enforced: Credence and mod-Credence incorporate constraints directly into the loss function during training, whereas Realcause and FrugalFlows apply post-hoc scaling to adjust the outcome distribution.

\begin{figure}[htb]
    \centering
    \includegraphics[width=0.3\linewidth]{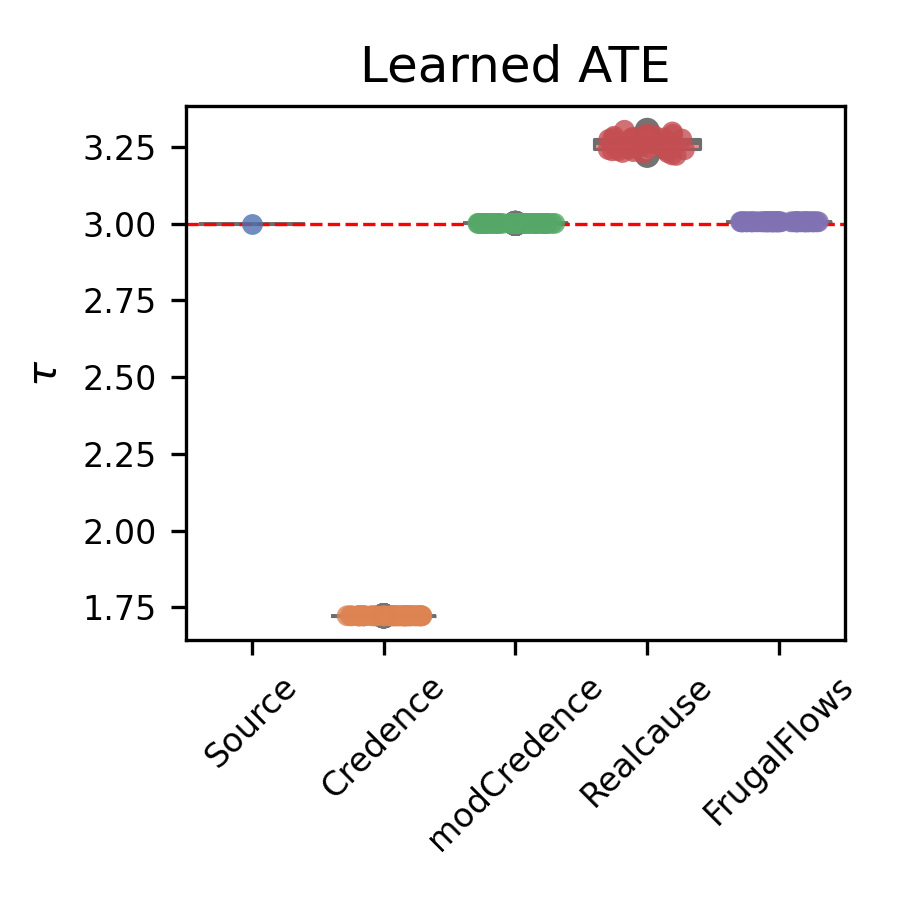}
    \includegraphics[width=0.3\linewidth]{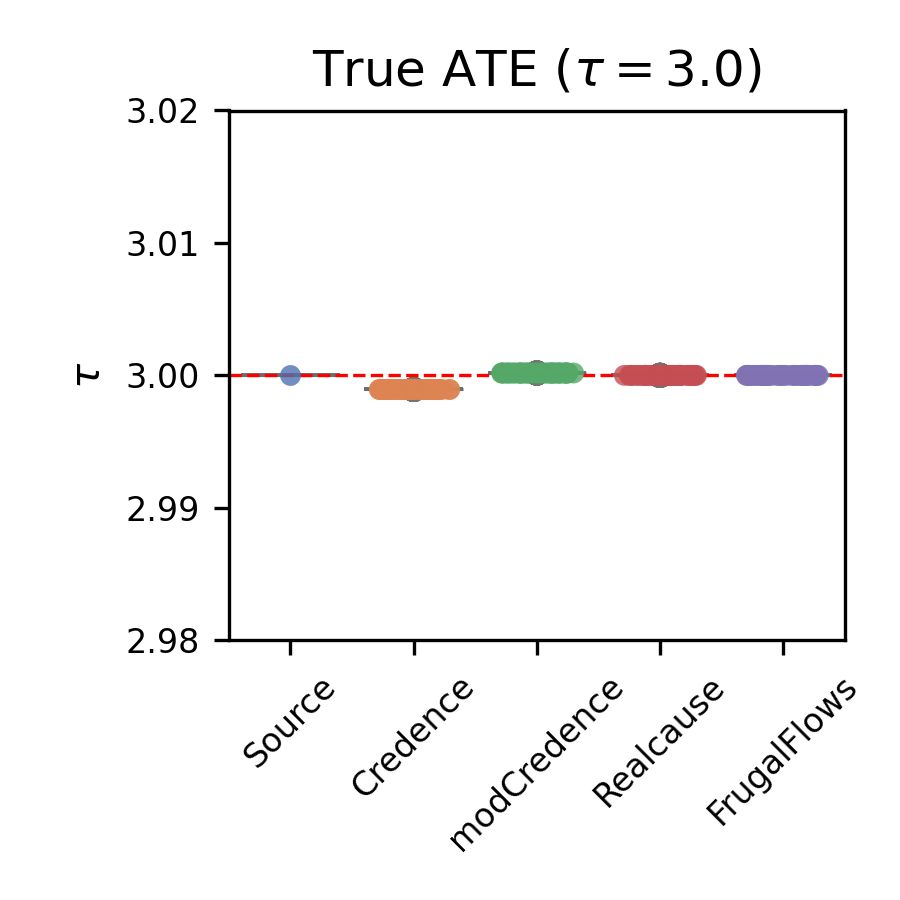}
    \includegraphics[width=0.3\linewidth]{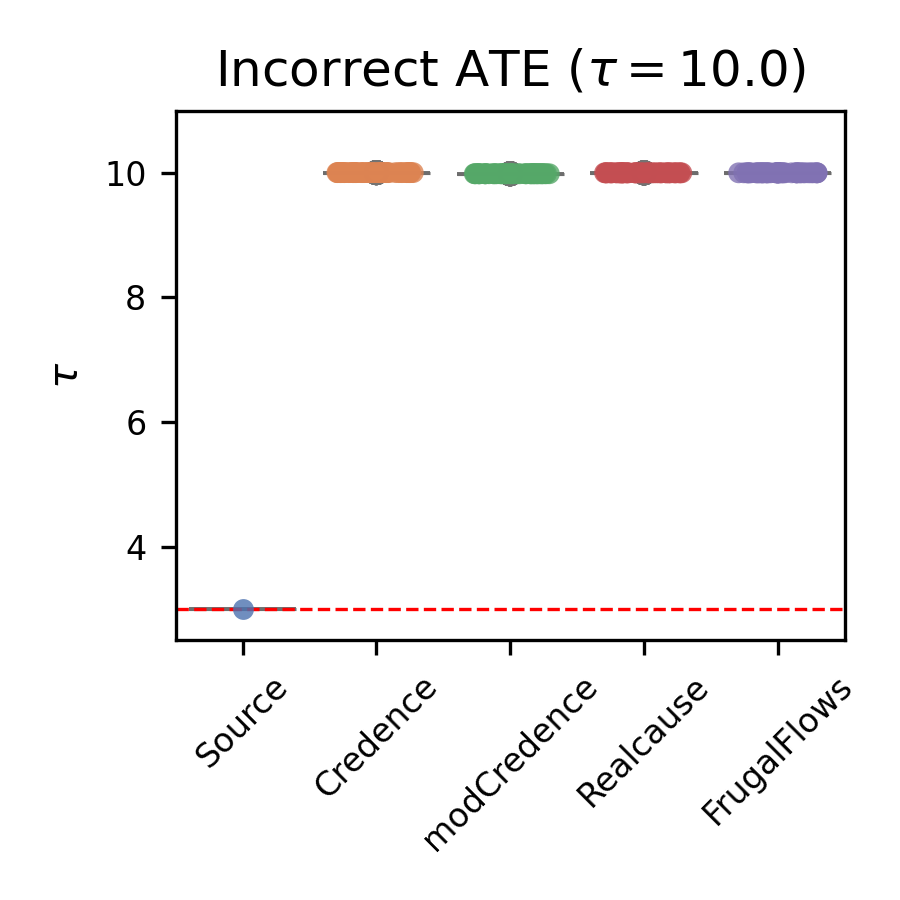}
    \caption{ATEs for the generated datasets across generative methods for the three different settings of the constraints.}
    \label{fig:ate-lp-dgp11}
\end{figure}

\paragraph{Learned ATE values.} Figure~\ref{fig:ate-lp-dgp11} compares the ATE of generated datasets across methods and settings. While all methods satisfy user-defined constraints when imposed, they learn different distributions under the flexible setting. Notably, Credence and Realcause produce different learned ATE values even for this simple, linear dataset.

\begin{figure}[htb]
    \centering
    \subfigure[Flexible ATE\label{fig:syn-linear-dgp1-flexible}]{
        \includegraphics[width=0.31\textwidth,trim=0 1.1cm 0 0,clip]{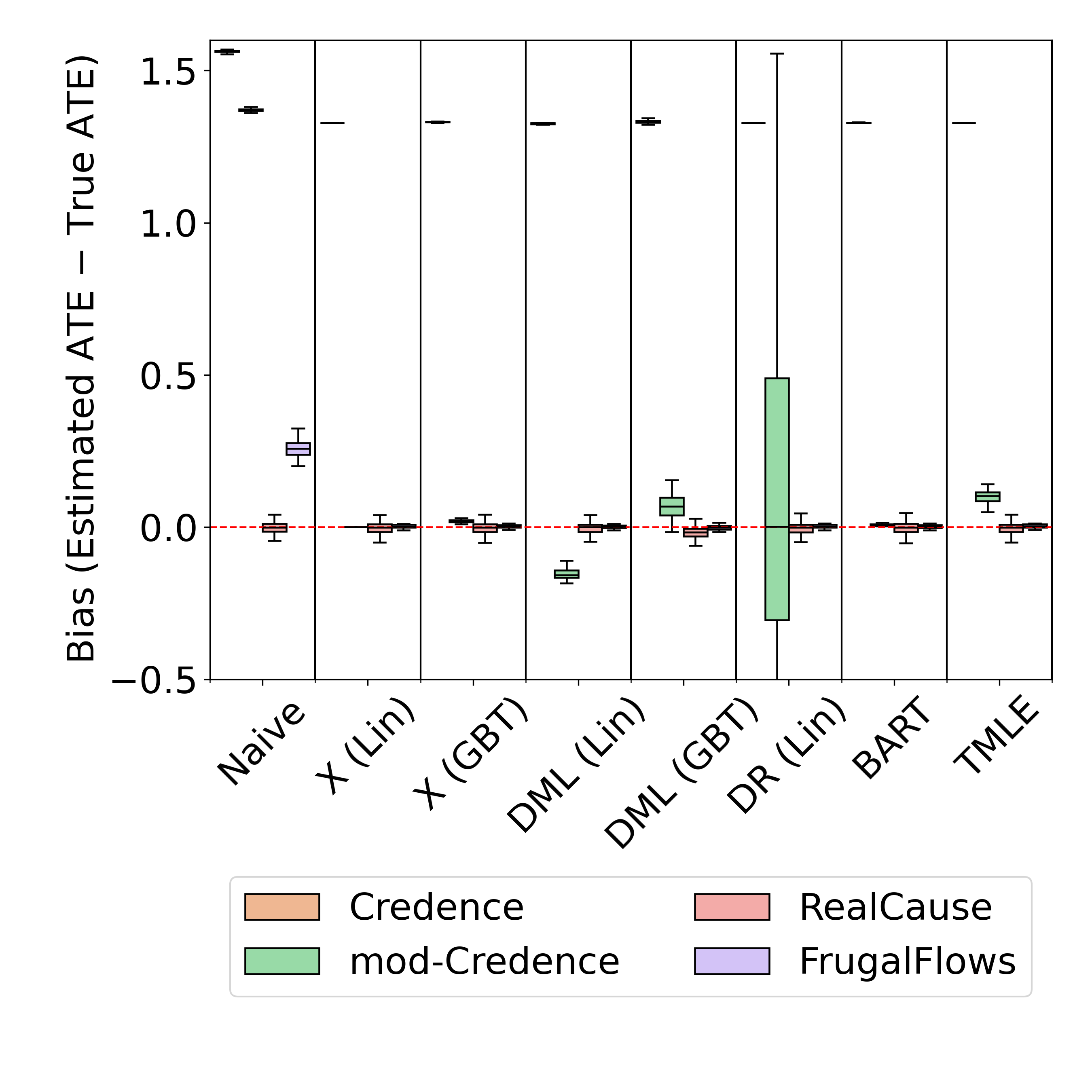}
    }
    \subfigure[True ATE\label{fig:syn-linear-dgp1-true}]{
        \includegraphics[width=0.31\textwidth,trim=0 1.1cm 0 0,clip]{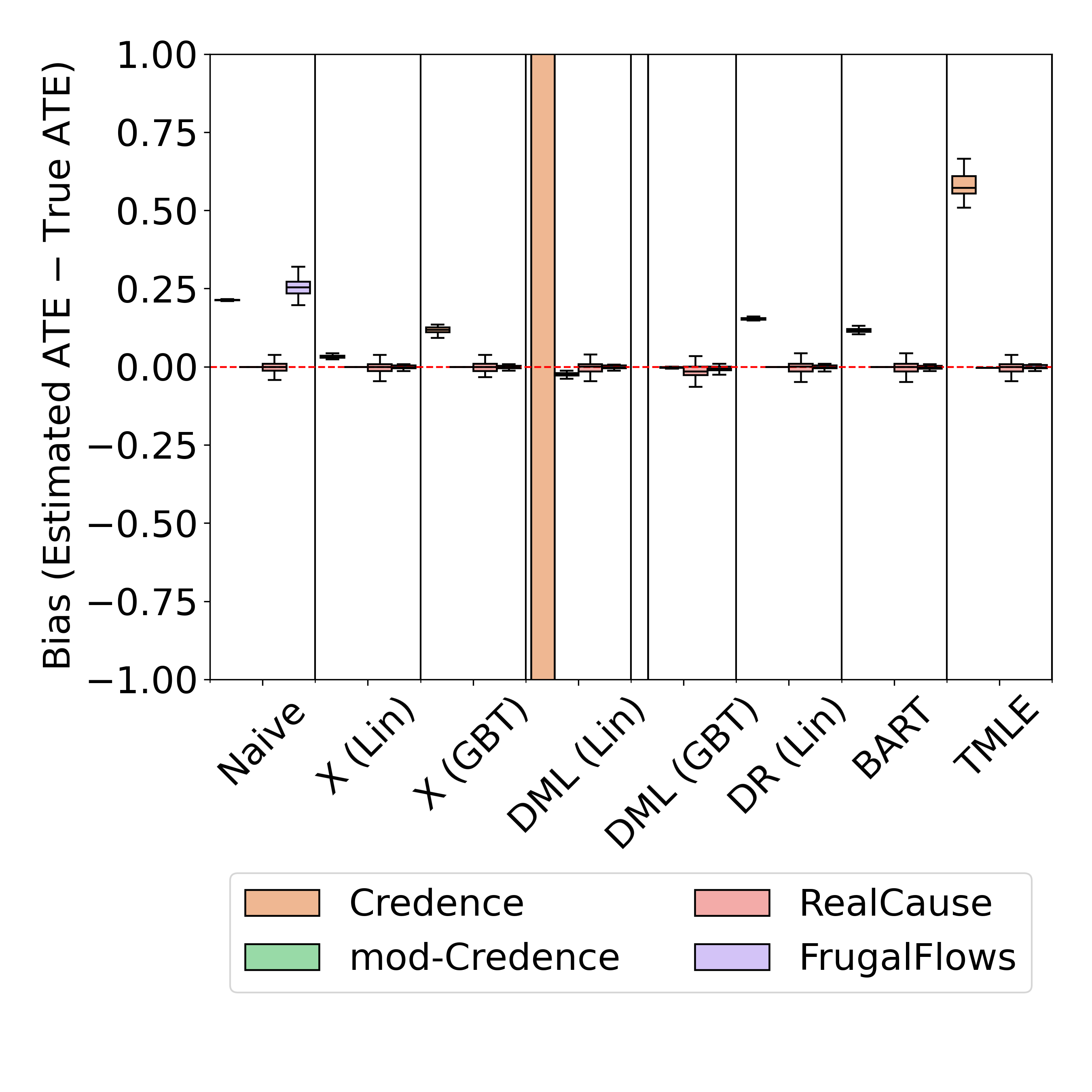}
    }
    \subfigure[Incorrect ATE\label{fig:syn-linear-dgp1-incorrect}]{
        \includegraphics[width=0.31\textwidth,trim=0 1.1cm 0 0,clip]{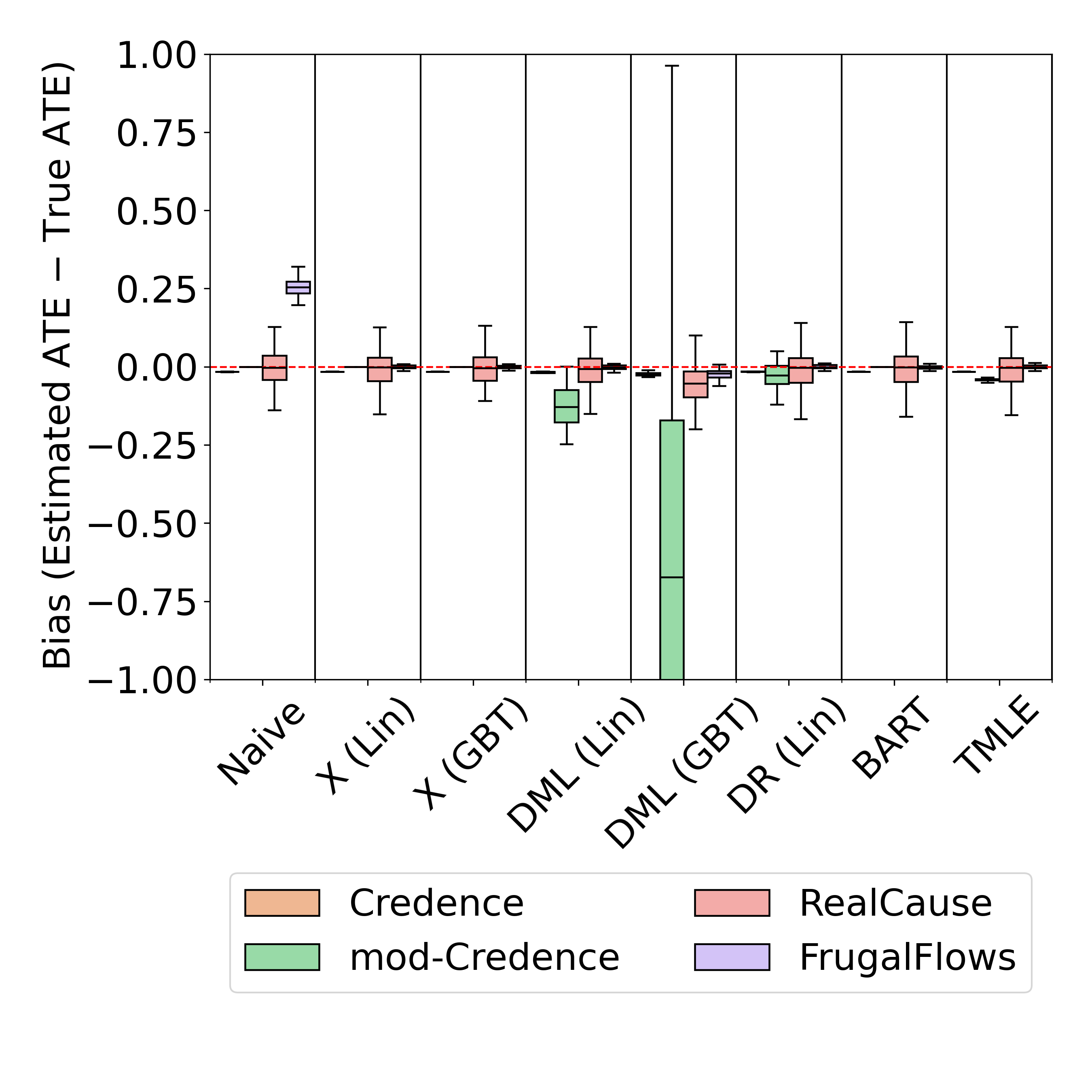}
    }
    \caption{Boxplots of the bias (estimated ATE $-$ true ATE) for a set of causal estimators for all generative methods across three different settings for the Synthetic DGP1 dataset.}
    \label{fig:syn-linear}
\end{figure}

\paragraph{Estimator bias.} Figure~\ref{fig:syn-linear} presents the bias (estimated ATE minus ground-truth ATE) for a set of causal estimators across 50 generated datasets per method. For most settings, estimator biases are comparable across methods. The exception is Credence under the learned ATE setting (Figure~\ref{fig:syn-linear-dgp1-flexible}), which exhibits notably higher bias for several estimators.

\paragraph{Key takeaway.} Even for a simple synthetic dataset with linear relationships and few covariates, generative methods produce substantially different results---particularly under the learned ATE setting where no constraints are imposed. This underscores that method selection matters even in idealized scenarios.

\subsubsection{Param DGP12}
We next examine generative methods on a dataset with a larger number of covariates and non-linear relationships between variables. We use a dataset derived from simulation SI 6 in~\citet{kunzel2019metalearners}, with the following data-generating process:

\begin{equation}
    \begin{aligned}
        X \sim \text{Uniform}([0, 1]^{N \times 20}) \\
        e(X) = \frac{1}{4}(1 + \text{Beta}(X_1, 2, 4)\\
        \mu_0(X) = 2 X_1 - 1 \\
        \mu_1(X) = \mu_0(X)
    \end{aligned}
    \label{eq:param-dgp12}
\end{equation}

This dataset contains 20 covariates $X_1\ldots X_{20}$, each drawn independently from a uniform distribution on $[0, 1]$. The propensity score $e(X)$ introduces non-linearity through a Beta distribution with shape parameters 2 and 4, evaluated at $X_1$. The potential outcomes $\mu_0(X)$ and $\mu_1(X)$ depend only on $X_1$, and since they are identical, the ground-truth ATE is $\tau^*=0.0$. We generate $N = 2000$ samples for this experiment. For this dataset, we train all generative methods under the flexible ATE setting only (without imposing constraints on the {\knob}s).

\begin{figure}[ht]
\centering
\begin{minipage}{0.55\textwidth}
    \centering
    \includegraphics[scale=0.33]{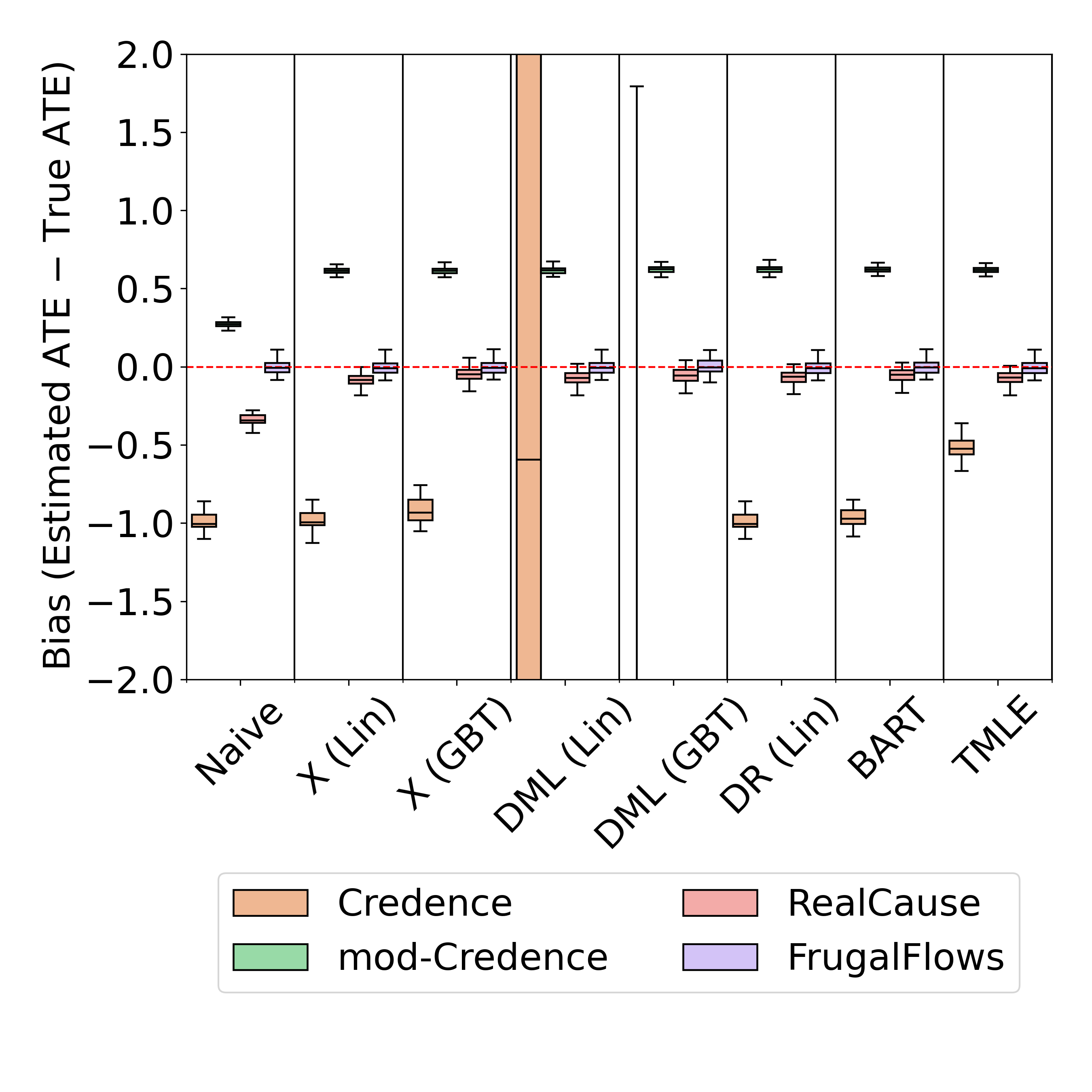}
\end{minipage}%
\begin{minipage}{0.45\textwidth}
    \centering
        \begin{tabular}{ll}
        \toprule
        Gen. Method & SWD \\
        \midrule
        Credence           & $1.027 \pm 0.026$ \\
        Modified Credence  & $0.784 \pm 0.031$ \\
        Realcause          & $0.788 \pm 0.031$ \\
        Frugal Flows        & $0.315 \pm 0.029$ \\
        \bottomrule
        \end{tabular}
\end{minipage}
\caption{Estimator bias and SWD for the flexible ATE setting across all methods: Param DGP12}
\label{fig:results-param-dgp12}
\end{figure}

\paragraph{Results} Figure~\ref{fig:results-param-dgp12} presents estimator bias across 50 generated datasets per method as well as the mean sliced-Wasserstein distances. With the exception of Credence, all methods yield comparable estimator performance, with biases centered near zero. Credence produces notably higher and more variable bias across most estimators. FrugalFlows achieves the lowest distance, indicating greater distributional fidelity, while Credence exhibits the highest distance. 

\paragraph{Key takeaway.} With higher-dimensional covariates and non-linear propensity scores, differences between generative methods become more apparent. Credence struggles to approximate the source distribution, resulting in higher distributional distance and greater estimator bias, while the other methods perform comparably.

\subsubsection{Param DGP13} 
We next consider a dataset with 20 covariates and no confounding relationship between treatment and covariates. We use simulation SI 2 from~\citet{kunzel2019metalearners}, with the following data-generating process:

\begin{equation}
    \begin{aligned}
        X \sim \mathcal{N}(0, \Sigma) \\
        e(X) = 0.5 \\
        \mu_1(X) = X^T\beta_1, \text{ with } \beta_1 \sim \text{Unif}([1, 30]^{20}) \\
        \mu_0(X) = X^T\beta_0, \text{ with } \beta_0 \sim \text{Unif}([1, 30]^{20}) 
    \end{aligned}
    \label{eq:param-dgp13}
\end{equation}

This dataset contains 20 covariates drawn from a multivariate normal distribution with covariance $\Sigma$. The propensity score $e(X) = 0.5$ is constant, implying random treatment assignment with no confounding. The potential outcomes $\mu_0(X)$ and $\mu_1(X)$ are linear functions of the covariates, with coefficients drawn independently from a uniform distribution on $[1, 30]$. We generate 2000 samples and train all generative methods under the learned ATE setting only.

\begin{figure}[ht]
\centering
\begin{minipage}{0.55\textwidth}
    \centering
    \includegraphics[scale=0.33]{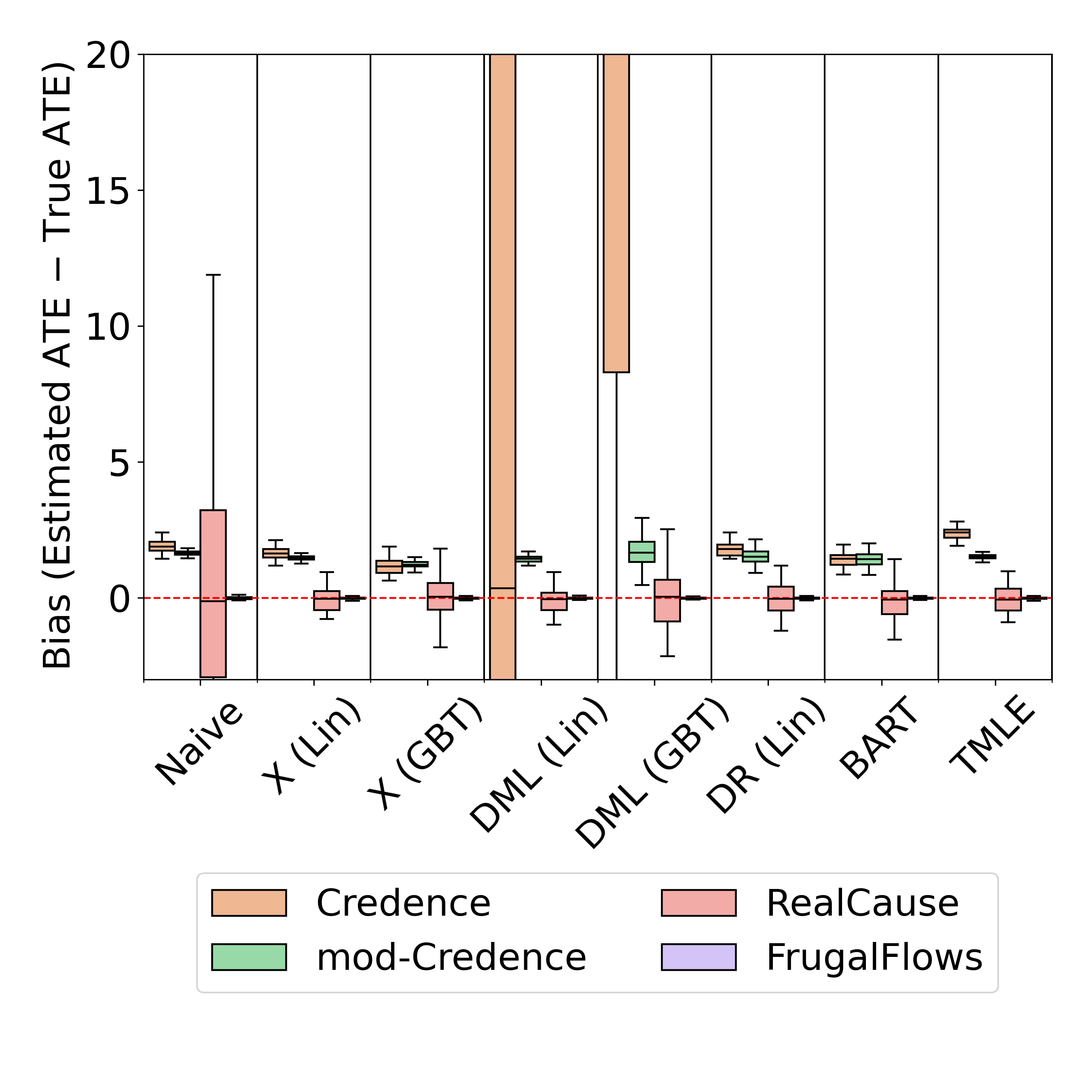}
\end{minipage}%
\begin{minipage}{0.45\textwidth}
    \centering
        \begin{tabular}{ll}
        \toprule
        Gen. Method & SWD \\
        \midrule
        Credence           & $1.002 \pm 0.003$ \\
        Modified Credence  & $26.436 \pm 2.263$ \\
        Realcause          & $20.825 \pm 1.869$ \\
        Frugal Flows        & $0.404 \pm 0.033$ \\
        \bottomrule
        \end{tabular}
\end{minipage}
\caption{Estimator bias and SWD for the flexible ATE setting across all methods: Param DGP13}
\label{fig:results-param-dgp13}
\end{figure}

\paragraph{Results.} Figure~\ref{fig:results-param-dgp13} presents estimator bias across 50 generated datasets per method as well as the mean sliced-Wasserstein distances. FrugalFlows achieves the lowest distance and exhibits the least bias across all estimators when compared to other generative methods.

\subsubsection{Param DGP14}
We next consider a lower-dimensional variant using simulation SI 4 from \citet{kunzel2019metalearners}. This dataset contains 5 covariates with the following data-generating process. As in Param DGP13, the propensity score is constant $e(X) = 0.5$ implying no confounding. The ground-truth ATE is $\tau^* = 0.0$. We generate 2,000 samples and train all generative methods under the learned ATE setting only. 

\begin{equation}
    \begin{aligned}
        X \sim \text{Uniform}([0, 1]^{N \times 5}) \\
        e(X) = 0.5\\
        \mu_0(X) = X^T \beta \text{ with } \beta \sim \text{Unif}([1, 30]^5) \\
        \mu_1(X) = \mu_0(X)
    \end{aligned}
    \label{eq:param-dgp14}
\end{equation}

\begin{figure}[ht]
\centering
\begin{minipage}{0.55\textwidth}
    \centering
    \includegraphics[scale=0.33]{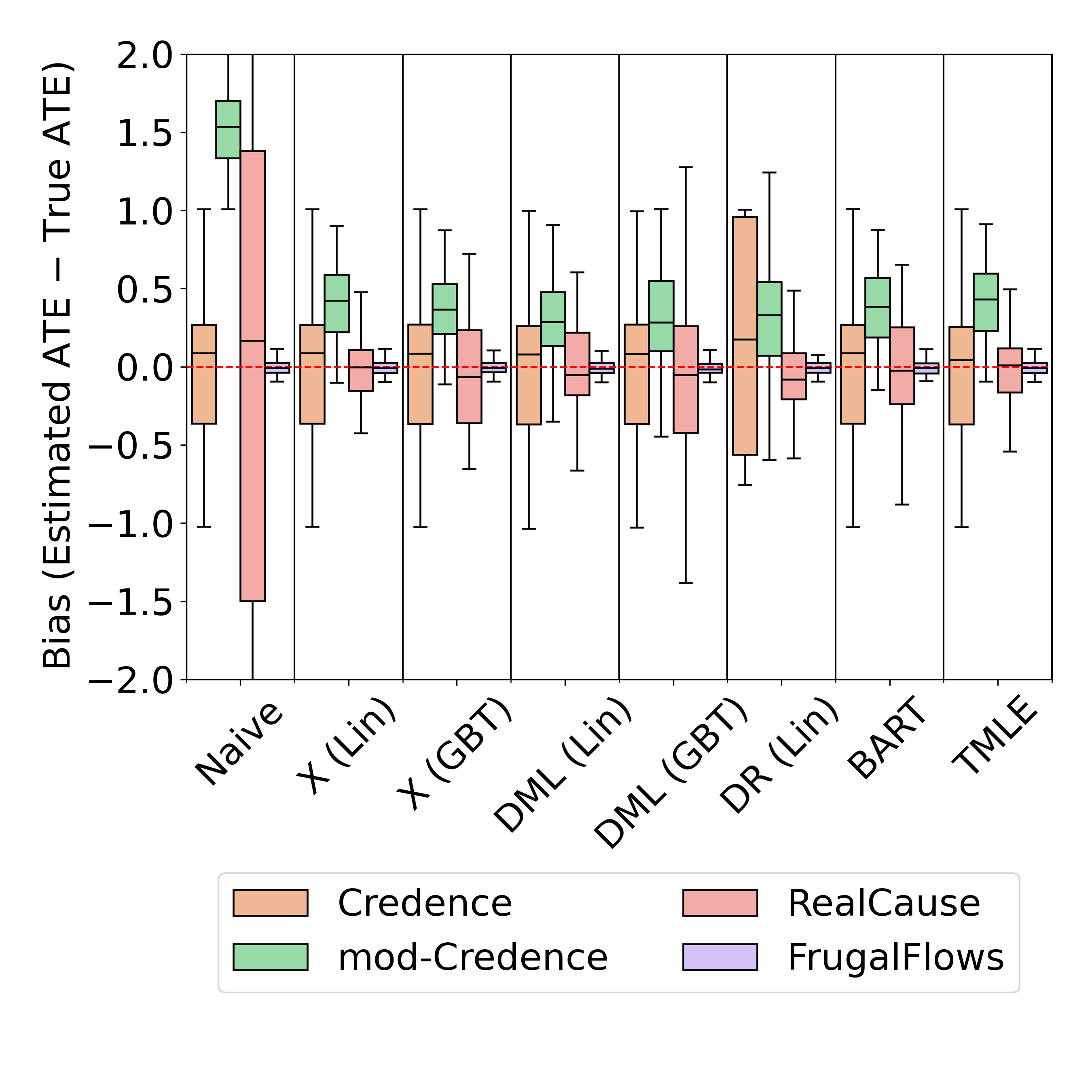}
\end{minipage}%
\begin{minipage}{0.45\textwidth}
    \centering
        \begin{tabular}{ll}
        \toprule
        Gen. method & SWD \\
        \midrule
        Credence           & $0.796 \pm 0.020$ \\
        Modified Credence  & $19.478 \pm 1.731$ \\
        Realcause          & $16.713 \pm 1.388$ \\
        Frugal Flows        & $0.389 \pm 0.033$ \\
        \bottomrule
        \end{tabular}
\end{minipage}
\caption{Estimator bias and SWD for the flexible ATE setting across all methods: Param DGP14}
\label{fig:results-param-dgp14}
\end{figure}

\paragraph{Results.} Figure~\ref{fig:results-param-dgp14} reports the mean sliced-Wasserstein distance between generated and source datasets. FrugalFlows achieves the lowest distance. In contrast, mod-Credence and Realcause exhibit substantially higher distances indicating poor distributional fidelity. These results are reflected by the estimator bias as well. 

\subsubsection{Param DGP15}

We next consider simulation SI 5 from \citet{kunzel2019metalearners}, which introduces a piecewise linear relationship between covariates and outcomes. The data-generating process is as follows:
\begin{equation}
    \begin{aligned}
        X \sim \text{Uniform}([0, 1]^{N \times 20}) \\
        e(X) = 0.5 \\
        \mu_0(X) &= 
            \begin{cases}
                X^T \beta_l & \text{ if } X_{20} < -0.4 \\
                X^T \beta_m & \text{ if } -0.4 \leq X_{20} \leq 0.4 \\
                X^T \beta_u & \text{ if } 0.4 < X_{20} 
            \end{cases}
        \\
        \mu_1(X) = \mu_0(X)
    \end{aligned}
\end{equation}
This dataset contains 20 covariates drawn from a uniform distribution on $[0,1]$. The propensity score is constant $e(X) = 0.5$, implying no confounding. The potential outcome $\mu_0(X)$ is a piecewise linear function of the covariates, with different coefficient vectors $\beta_l, \beta_m, \beta_u$ depending on the value of $X_{20}$. Since $\mu_1(X) = \mu_0(X)$ the ground-truth ATE is $\tau^*=0.0$. We generate 2000 samples and train all generative methods under the learned ATE setting only.

\begin{figure}[ht]
\centering
\begin{minipage}{0.55\textwidth}
    \centering
    \includegraphics[scale=0.33]{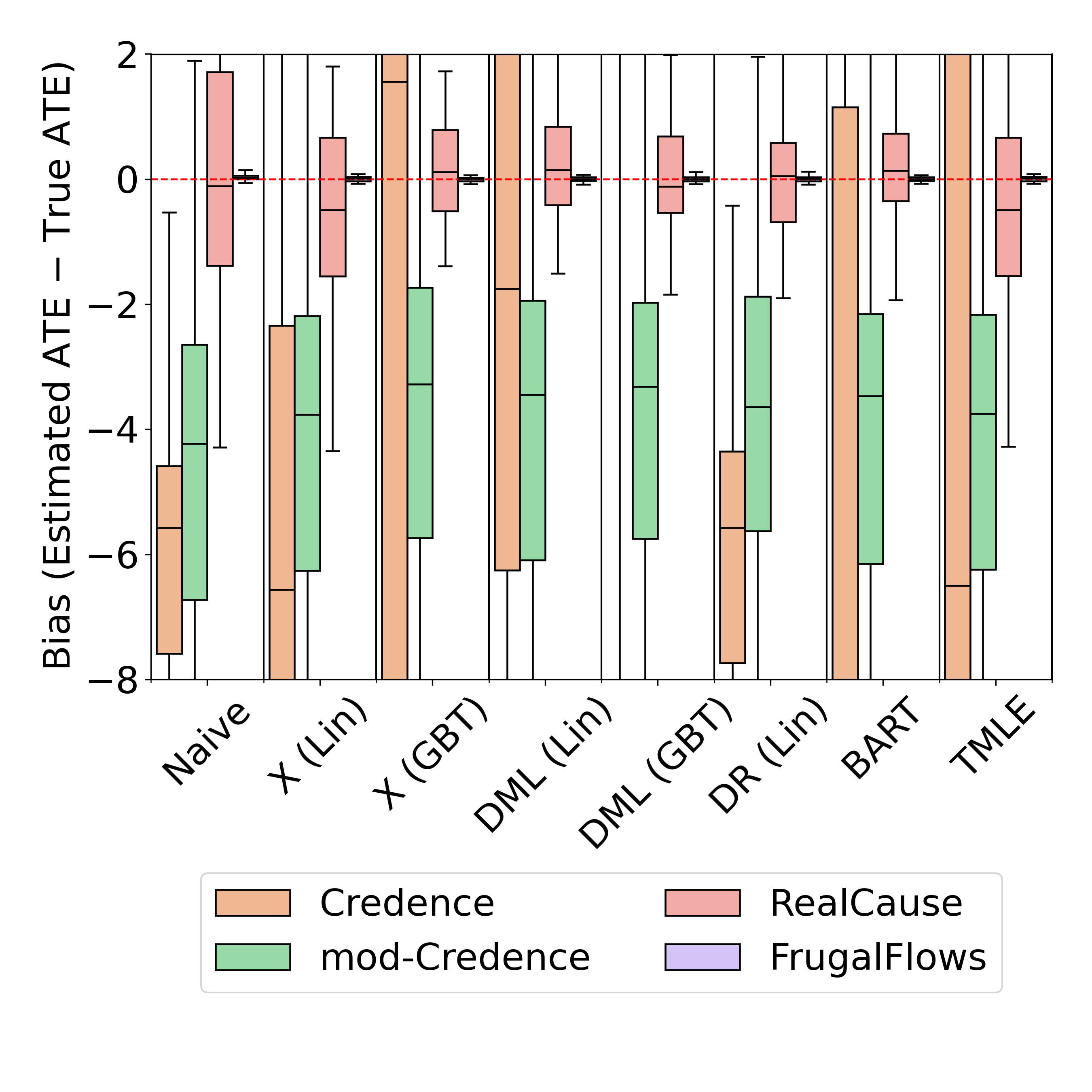}
\end{minipage}%
\begin{minipage}{0.45\textwidth}
    \centering
        \begin{tabular}{ll}
        \toprule
        Gen. method & SWD \\
        \midrule
        Credence           & $7.524 \pm 0.644$ \\
        Modified Credence  & $9.743 \pm 0.945$ \\
        Realcause          & $10.428 \pm 1.018$ \\
        Frugal Flows        & $0.220 \pm 0.018$ \\
        \bottomrule
        \end{tabular}
\end{minipage}
\caption{Estimator bias and SWD for the flexible ATE setting across all methods: Param DGP15}
\label{fig:results-param-dgp15}
\end{figure}

\paragraph{Results.} Figure~\ref{fig:results-param-dgp15} reports the mean sliced-Wasserstein distance between generated and source datasets. FrugalFlows achieves by far the lowest distance while Credence, mod-Credence, and Realcause all exhibit substantially higher distances. The piecewise linear structure appears to pose challenges for these methods. Consistent with its superior distributional fidelity, FrugalFlows produces estimator biases tightly centered around zero. The other methods exhibit higher and more variable bias, with Credence showing the largest spread across estimators.

\paragraph{Key takeaway.} The piecewise linear outcome model poses a significant challenge for most generative methods. FrugalFlows, which uses normalizing flows to flexibly model the data distribution, substantially outperforms the other methods in both distributional fidelity and estimator bias. This highlights the importance of selecting generative methods capable of capturing non-standard functional relationships in the data.

\subsection{Real-world observational datasets}
\label{sec:comparing-gen-real}

\subsubsection{Lalonde (RCT)}

We examine the experimental arm of the Lalonde dataset, which contains data from a randomized controlled trial evaluating the effect of a job training program on earnings. Due to randomization, this dataset exhibits no confounding, and the ground-truth ATE is directly estimable from the data. However, the sample size is relatively small, posing challenges for generative methods.

\paragraph{Distributional similarity.} Table~\ref{tab:swd-lalonde-rct} reports the mean sliced-Wasserstein distance between generated and source datasets across methods and `default settings' recommended by the generative method (such as rescaling values). We note that the SWD is high across all methods. We also note the limitations of using the SWD as a distance metric here---while it may be fast to compute, the distance metric is dependent on the scale of the data. We omit the SWD for Credence with the incorrect ATE as we found that the Credence simulator did not converge to an acceptable value. 
 
\begin{table}[htb]
\centering
\small
\begin{tabular}{@{}lcccc@{}}
\toprule
 & Credence & mod-Credence & Realcause & FrugalFlows \\ \midrule
Learned ATE & $5693\pm364$ & $5069\pm344$ & $5049\pm416$ & $7623\pm576$ \\
True ATE & $5525\pm457$ & $4998\pm336$& $5094\pm385$ & $7670\pm656$ \\
Incorrect ATE & $-$ & $5097\pm398$ & $5057\pm398$ & $7452\pm592$ \\ \bottomrule
\end{tabular}
\caption{SWD for (unscaled) Lalonde (RCT) across generative methods and three settings of the ATE parameter. }
\label{tab:swd-lalonde-rct}
\end{table}

\begin{figure}[htb]
    \centering
    \includegraphics[width=0.8\textwidth]{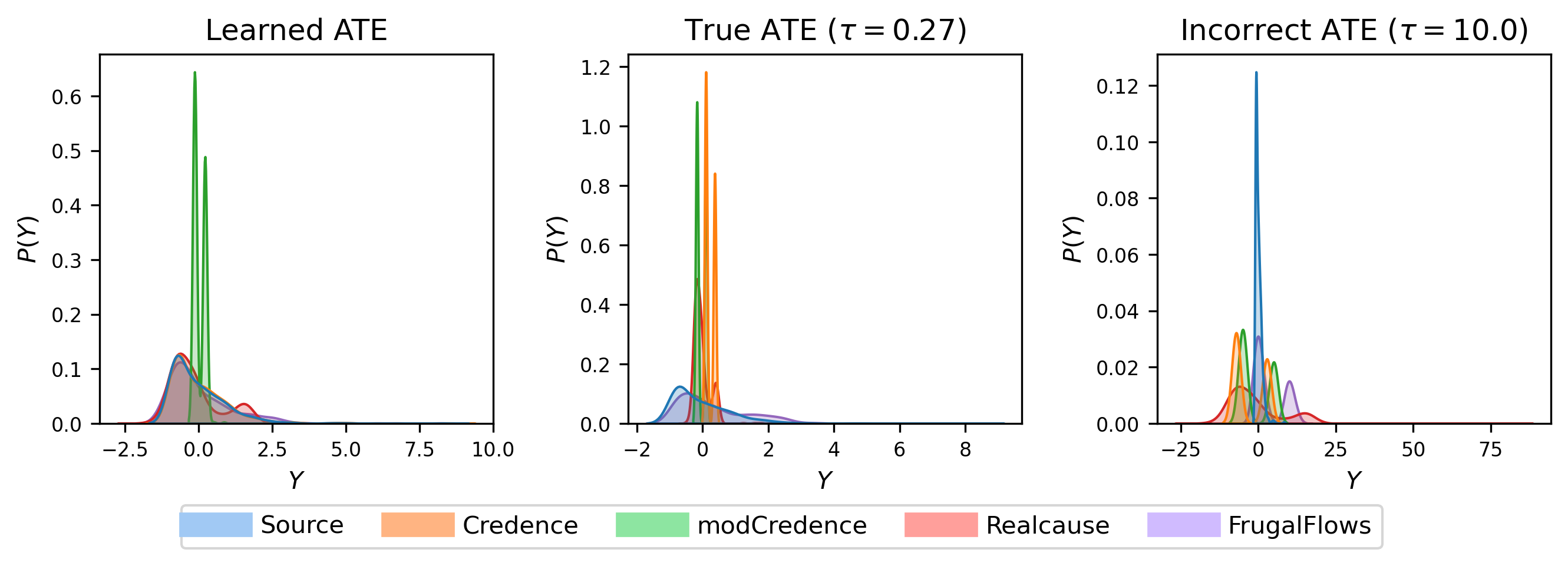}
    \caption{Marginal distribution of outcome $Y$ for the Lalonde (RCT) datasets.}
    \label{fig:outcome-lalonde-rct}
\end{figure}

\paragraph{Marginal distributions.} Figure~\ref{fig:outcome-lalonde-rct} displays the marginal distribution of outcome across settings and methods. All methods show visible differences from the source distribution, with variation across both methods and ATE settings. These discrepancies are consistent with the high SWD values observed in Table~\ref{tab:swd-lalonde-rct}.

\begin{figure}[htb]
    \centering
    \includegraphics[width=0.3\linewidth]{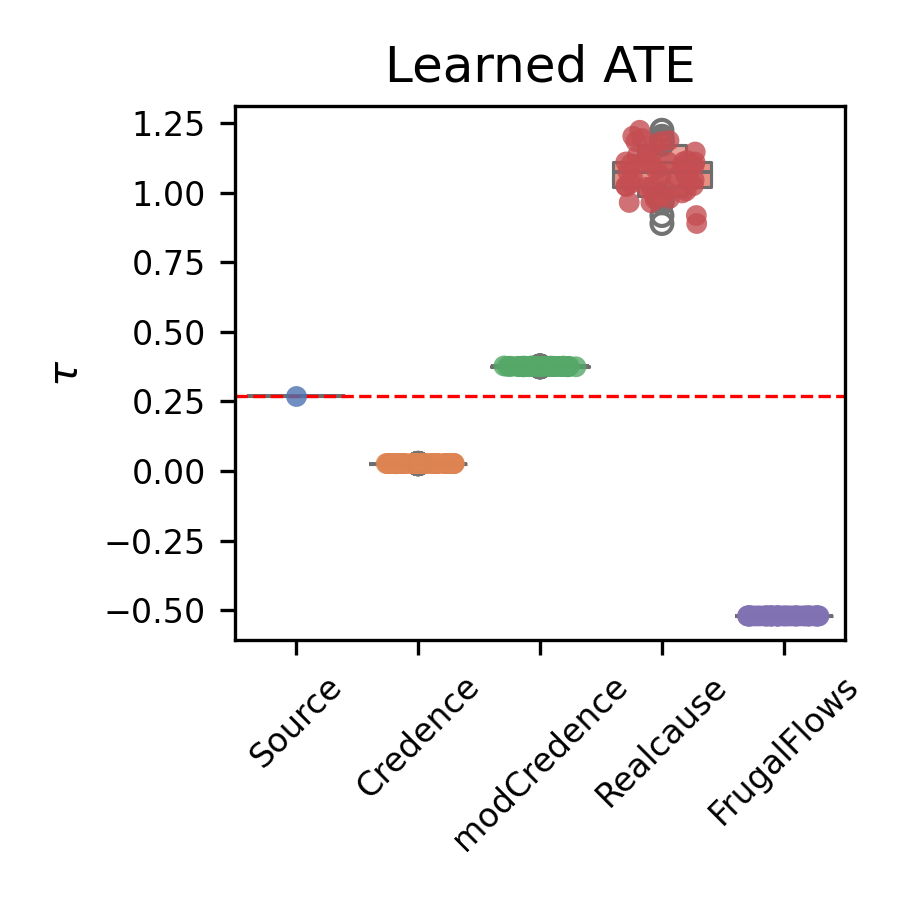}
    \includegraphics[width=0.3\linewidth]{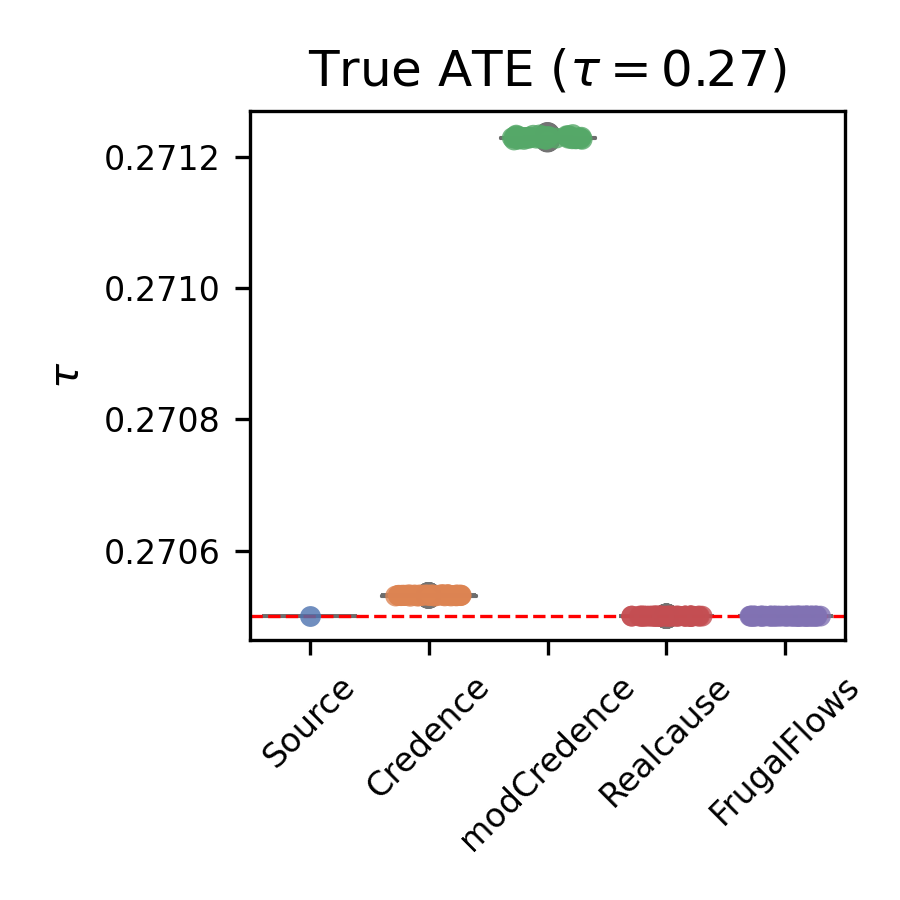}
    \includegraphics[width=0.3\linewidth]{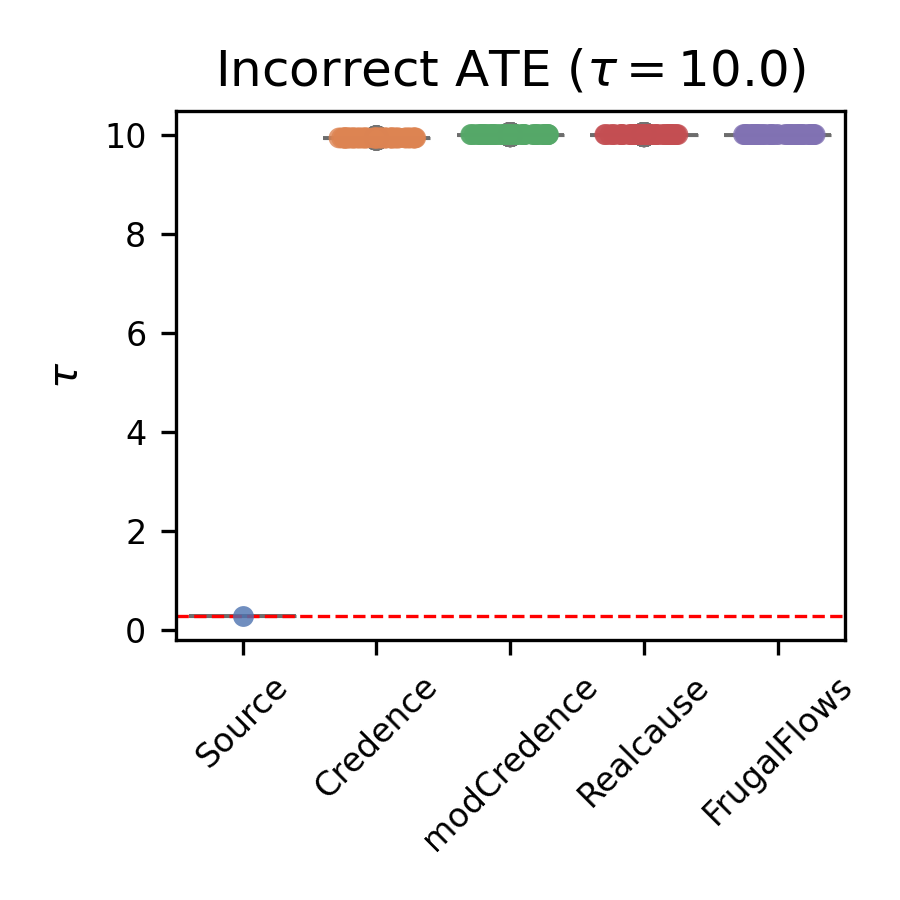}
    \caption{ATEs for the generated datasets across generative methods for the three different settings of the constraints.}
    \label{fig:ate-lalonde-rct}
\end{figure}

\paragraph{Learned ATE values.} Figure~\ref{fig:ate-lalonde-rct} compares the ATE of generated datasets across methods and settings. Under the learned ATE setting, methods converge to different values, illustrating how generative methods encode different assumptions about the treatment effect even when unconstrained. Under the true and incorrect ATE settings, all methods (except Realcause) approximately satisfy the imposed constraints.

\begin{figure}[htb]
    \centering
    \subfigure[Flexible ATE\label{fig:lalonde-exp-flexible}]{
        \includegraphics[width=0.31\textwidth,trim=0 1.1cm 0 0,clip]
        {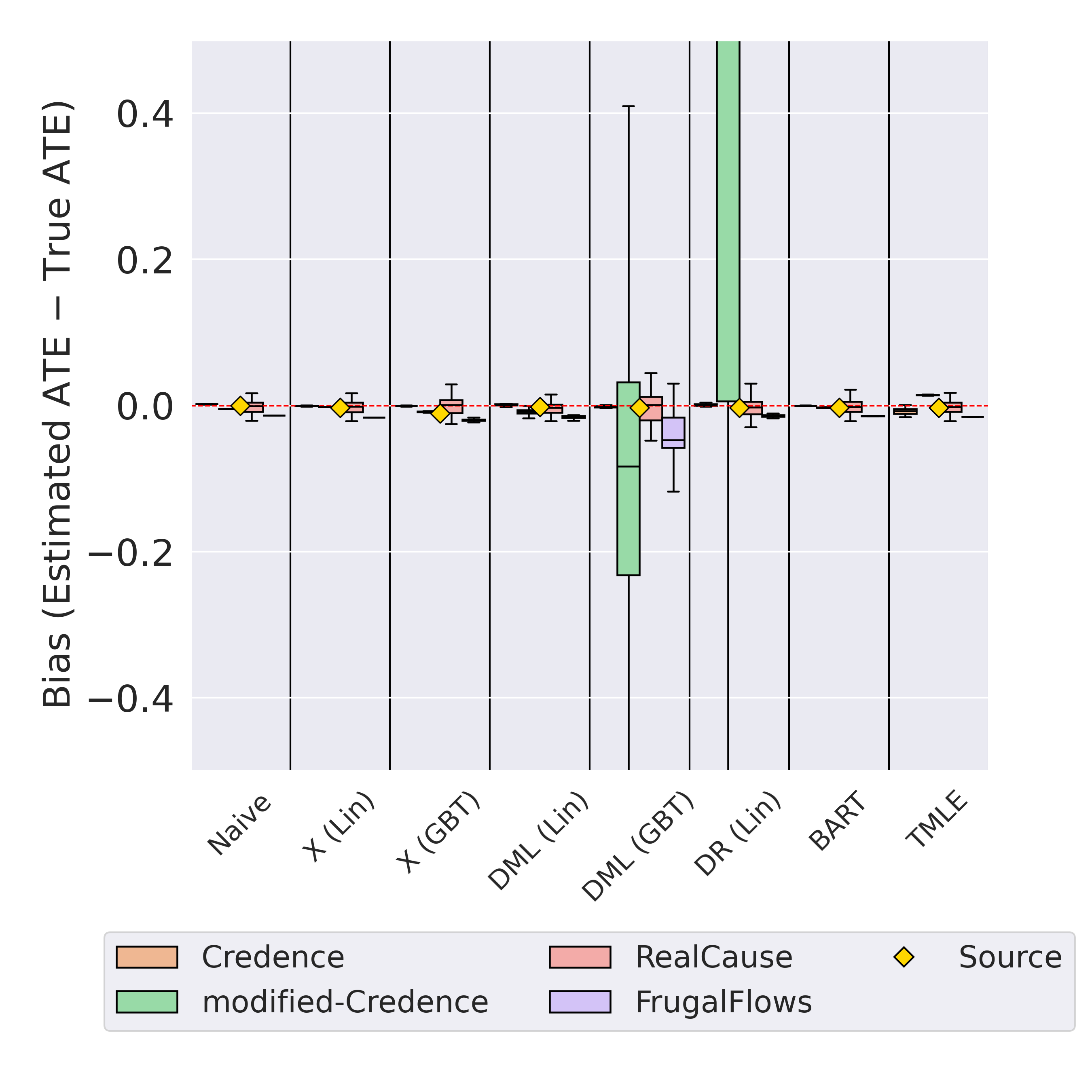}
    }
    \subfigure[True ATE\label{fig:lalonde-exp-true}]{
        \includegraphics[width=0.31\textwidth,trim=0 1.1cm 0 0,clip]
        {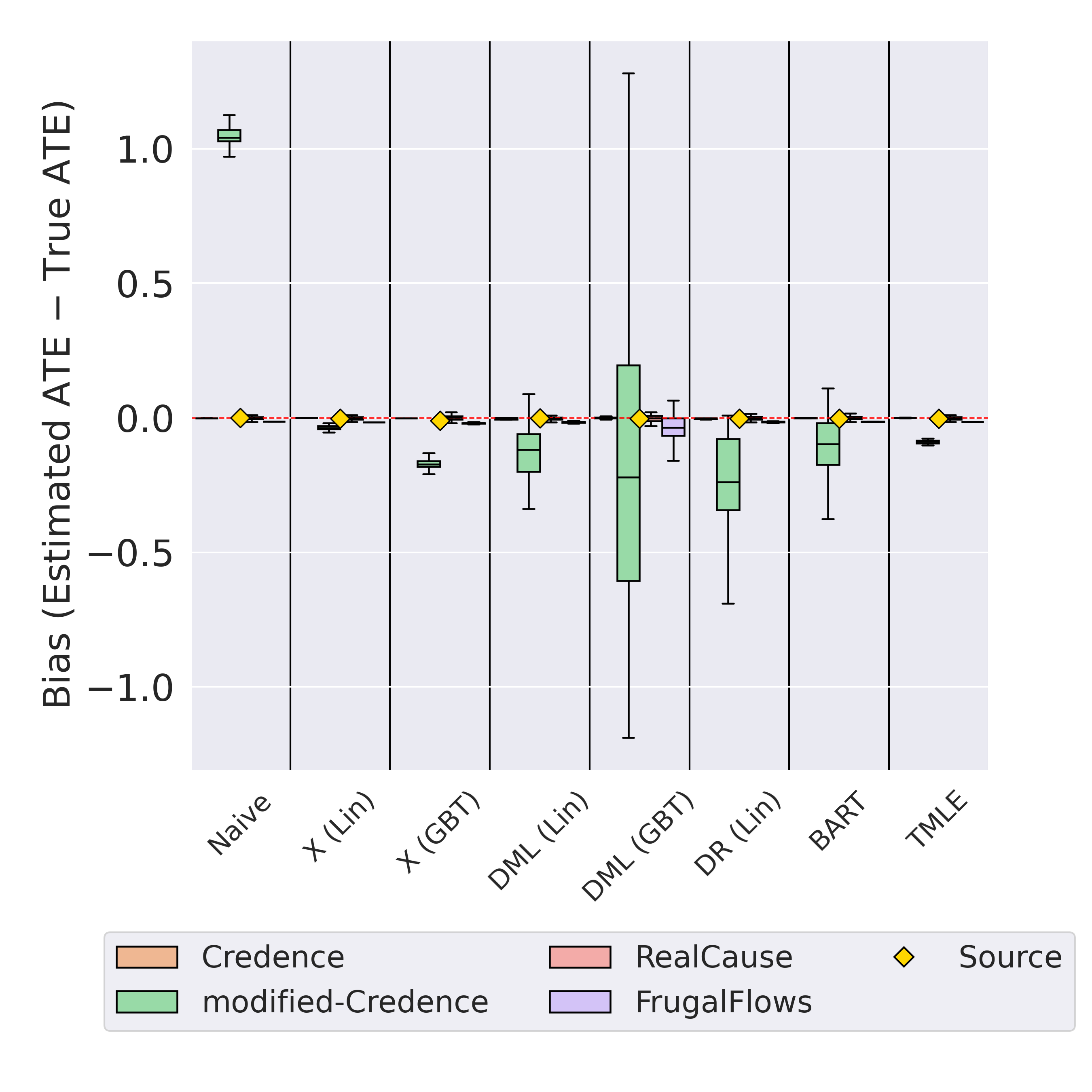}
    }
    \subfigure[Incorrect ATE\label{fig:lalonde-exp-incorrect}]{
        \includegraphics[width=0.31\textwidth,trim=0 1.1cm 0 0,clip]
        {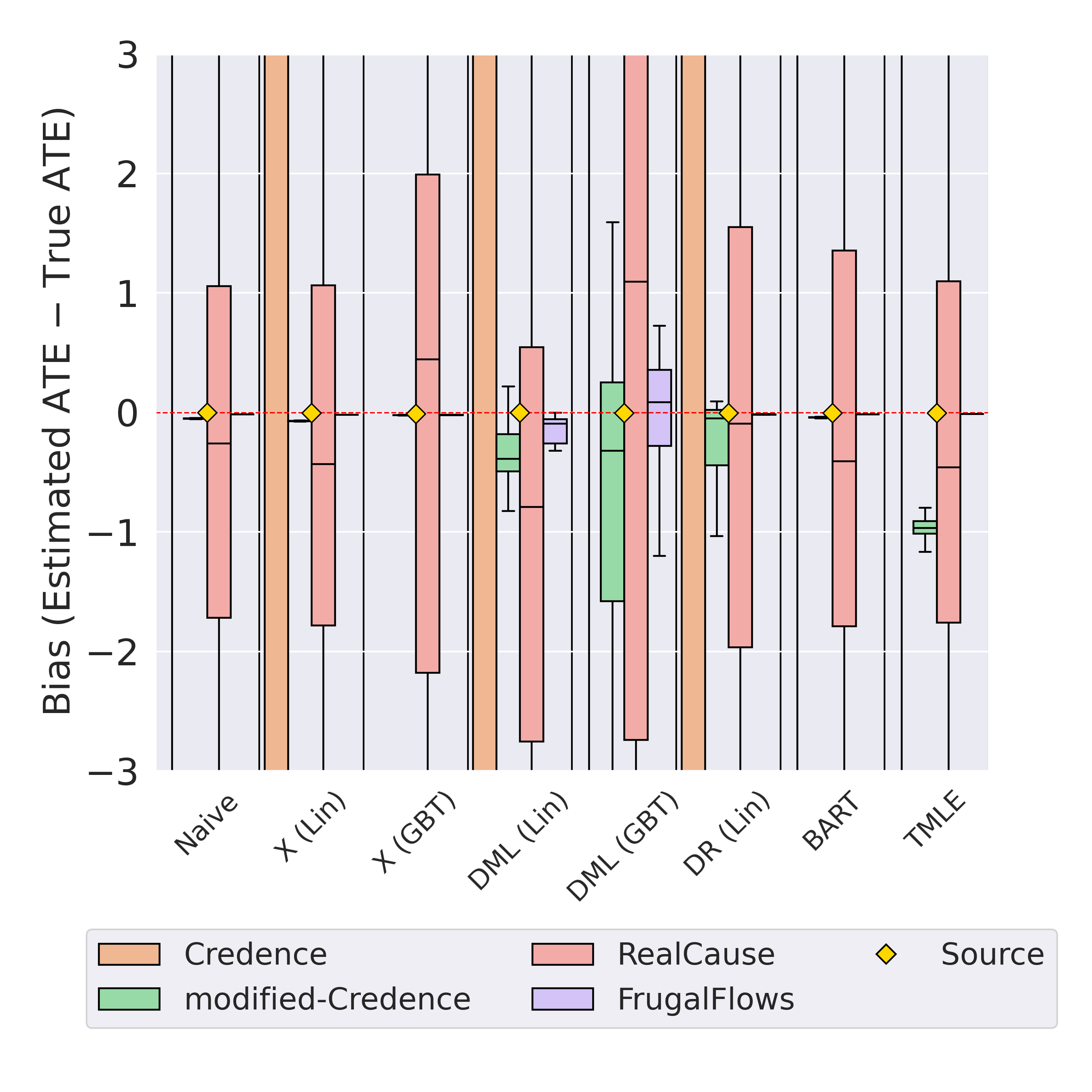}
    }
    \caption{Estimator bias across generative methods for Lalonde (RCT) under three ATE settings.}
    \label{fig:lalonde-rct}
\end{figure}

\paragraph{Estimator bias.} Figure~\ref{fig:lalonde-rct} presents estimator bias across 50 generated datasets per method. Despite there being no confounding bias in this dataset (drawn from an RCT), an incorrect {\knob} value results in biased conclusions for estimator performance. This variability likely reflects the small sample size of the source dataset, which limits the ability of generative methods to reliably approximate the underlying distribution.

\subsubsection{Lalonde}
\label{app:lalonde-obs-genmethods-full}

For the Lalonde (observational) dataset whose results were outlined in Section~\ref{sec:comparing-gen-methods}, we include additional details here. 

\begin{figure}[htb]
    \centering
    \includegraphics[width=0.8\textwidth]{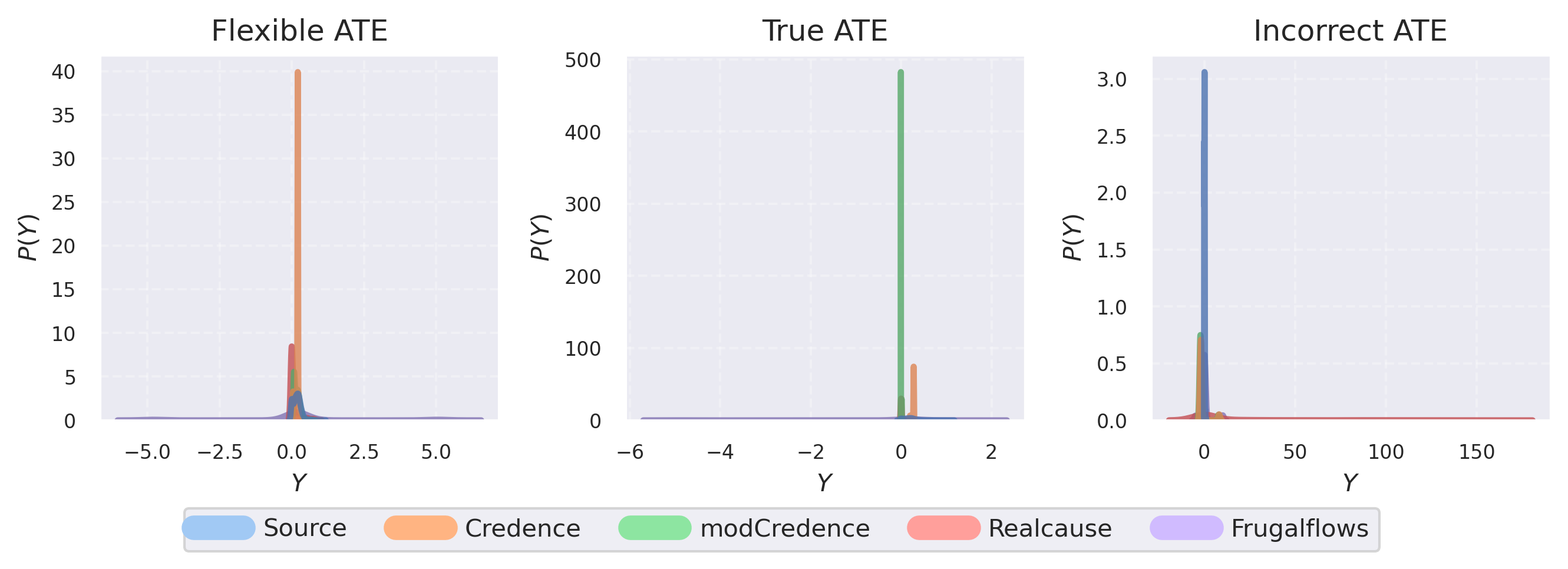}
    \caption{Marginal distribution of outcome $Y$ for the Lalonde datasets.}
    \label{fig:outcome-lalonde-obs}
\end{figure}

\paragraph{Marginal distributions.} Figure~\ref{fig:outcome-lalonde-obs} displays the marginal distribution of the outcome across settings and methods. We see that different methods learn different marginal distributions of the outcome.

\begin{figure}[htb]
    \centering
    \includegraphics[width=0.3\linewidth]{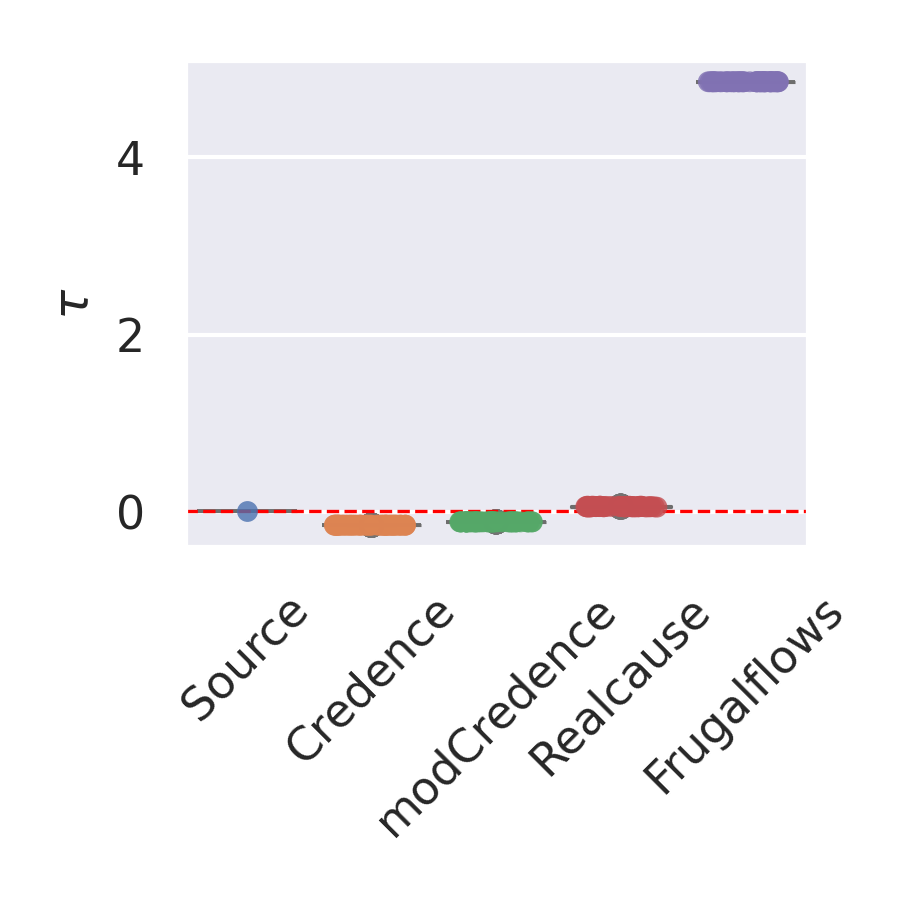}
    \includegraphics[width=0.3\linewidth]{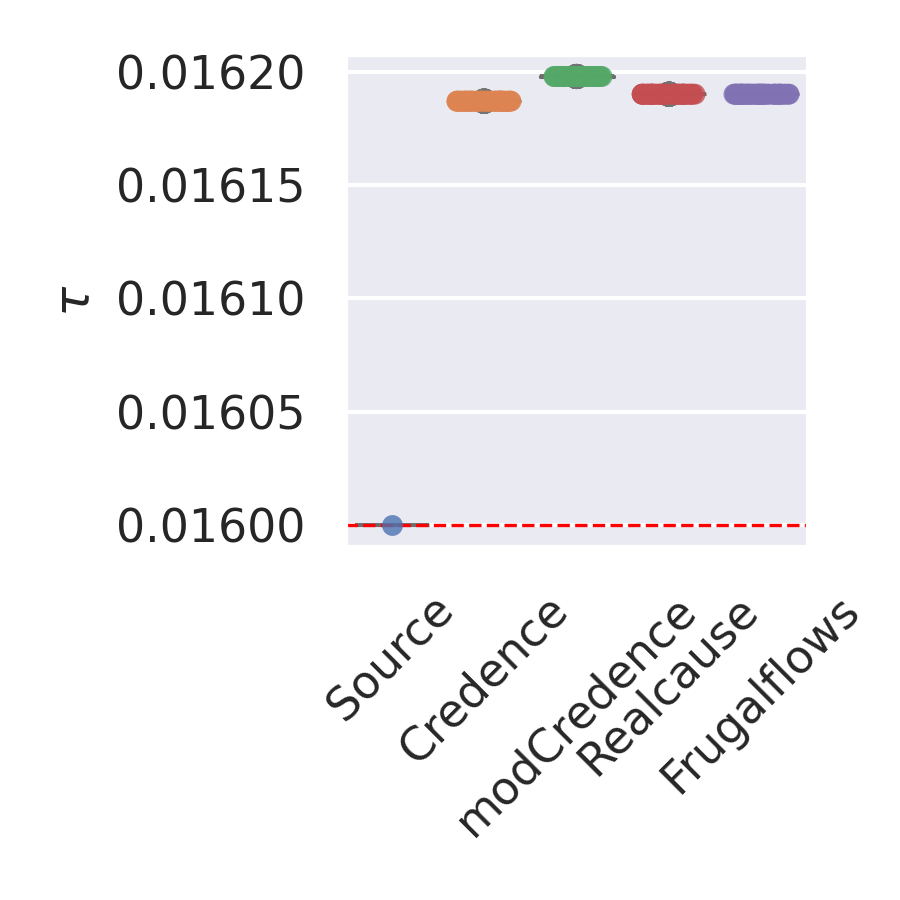}
    \includegraphics[width=0.3\linewidth]{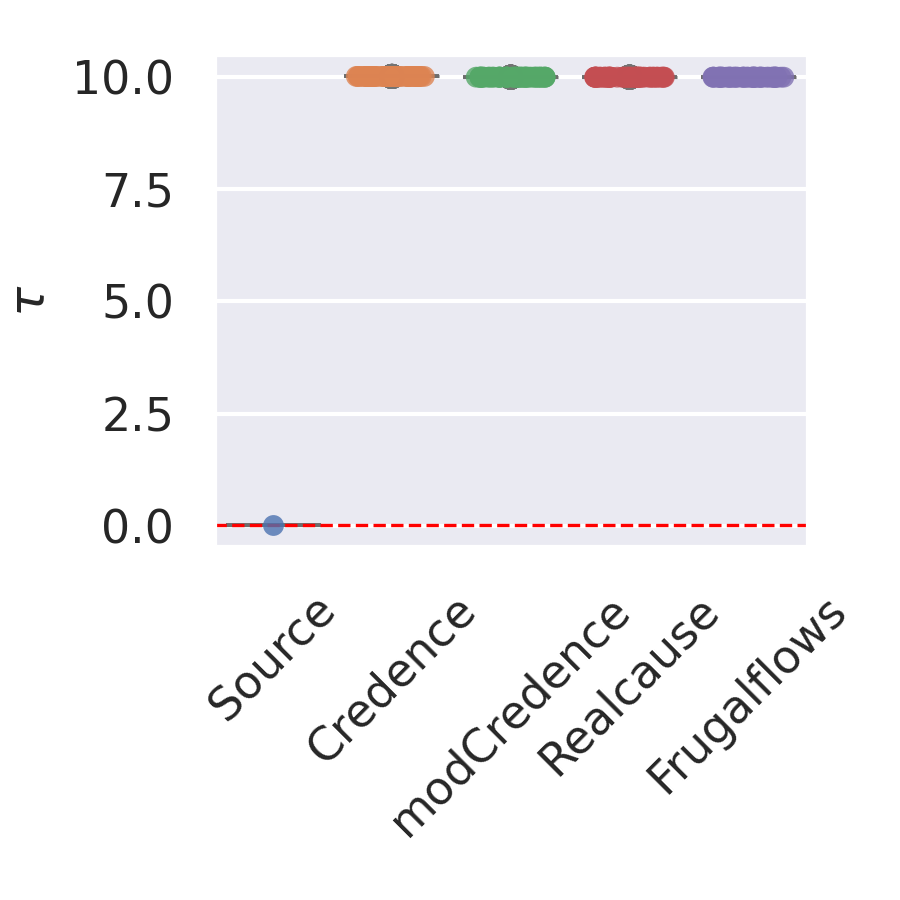}
    \caption{ATEs for the generated datasets across generative methods for the three different settings of the constraints.}
    \label{fig:ate-lalonde-obs}
\end{figure}

\paragraph{Learned ATE values.} Figure~\ref{fig:ate-lalonde-obs} compares the ATE of generated datasets across methods and settings. 

\begin{figure}[htb]
    \centering
    \subfigure[Flexible ATE\label{fig:lalonde-exp-flexible-full}]{
        \includegraphics[width=0.31\textwidth,trim=0 1.1cm 0 0,clip]
        {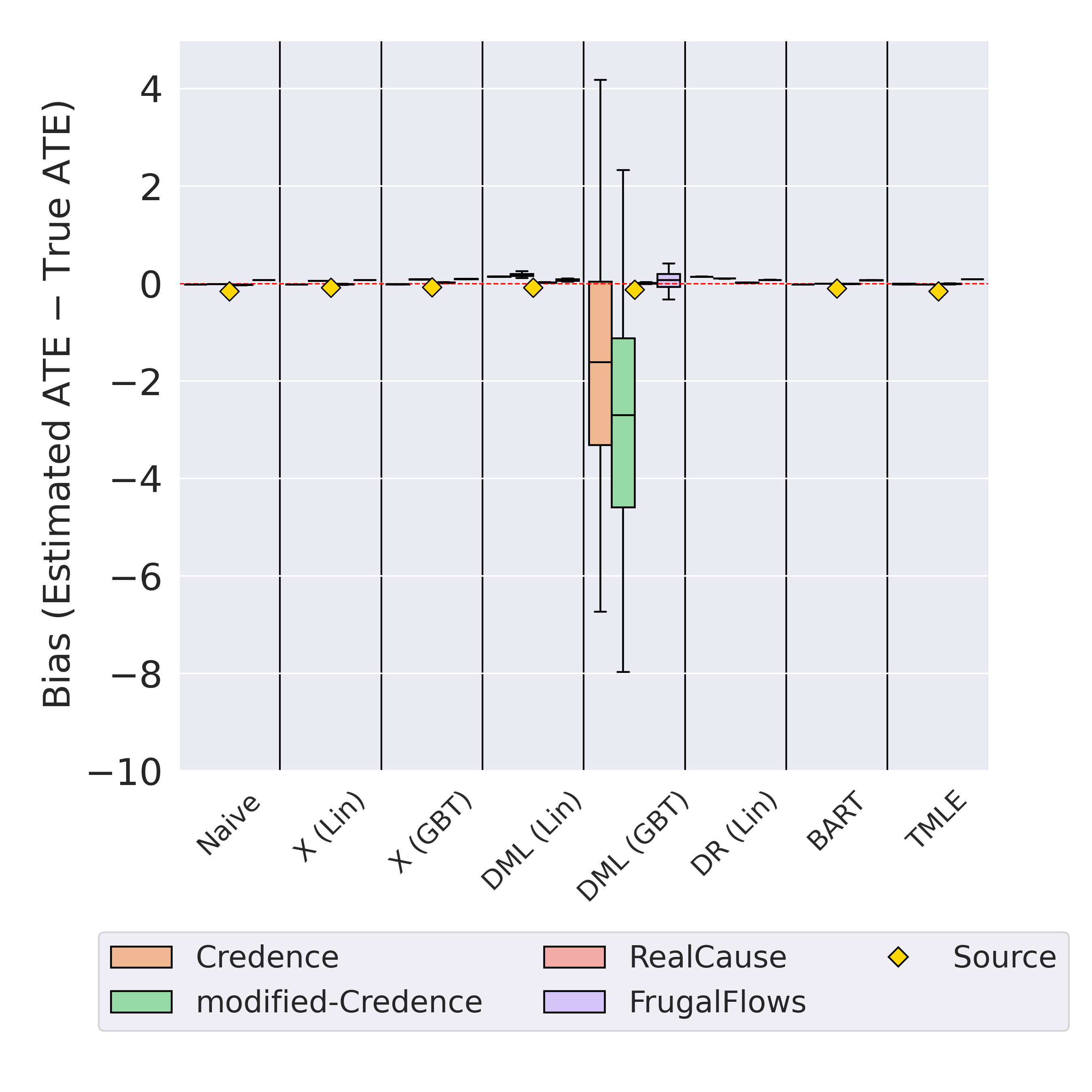}
    }
    \subfigure[True ATE\label{fig:lalonde-exp-true-full}]{
        \includegraphics[width=0.31\textwidth,trim=0 1.1cm 0 0,clip]
        {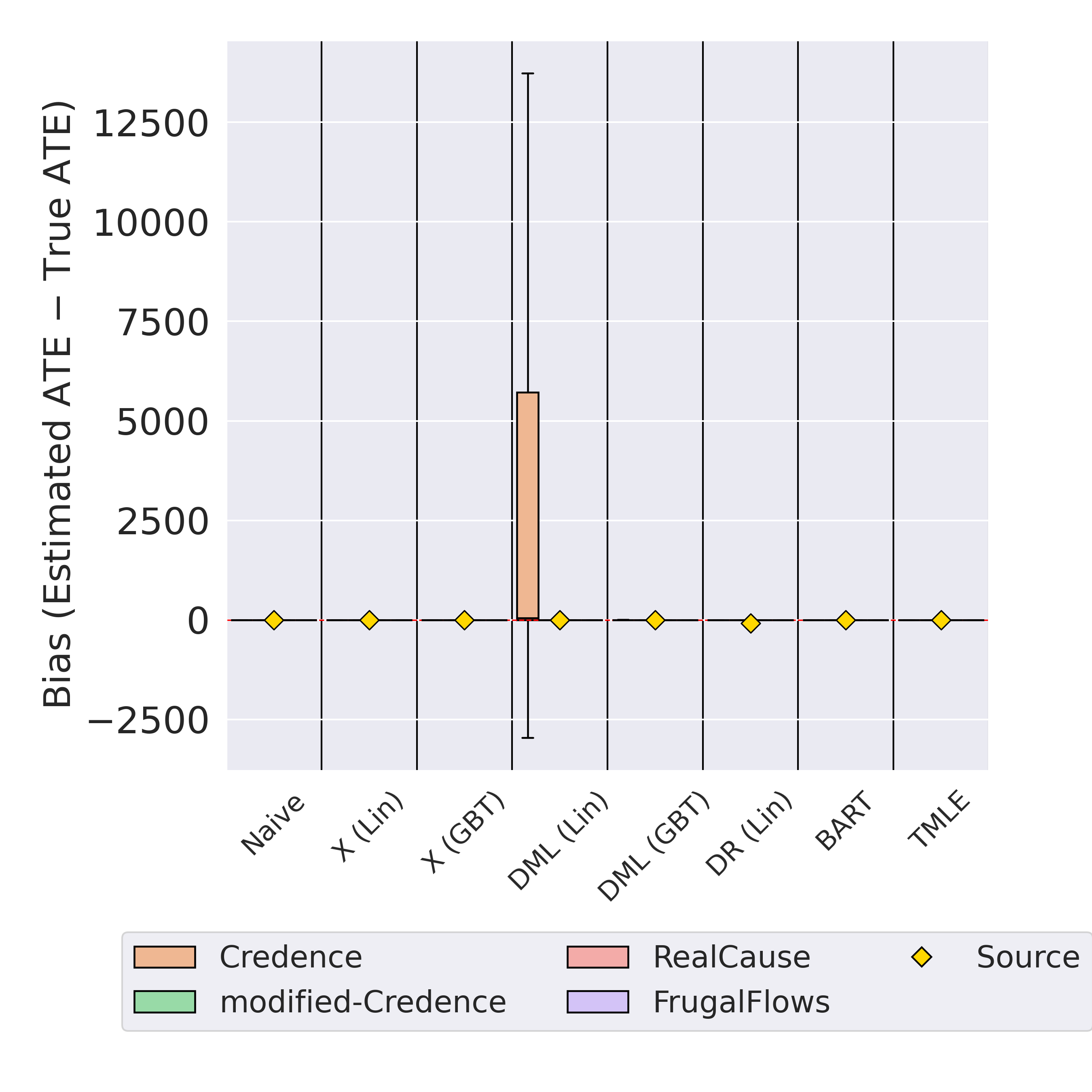}
    }
    \subfigure[Incorrect ATE\label{fig:lalonde-exp-incorrect-full}]{
        \includegraphics[width=0.31\textwidth,trim=0 1.1cm 0 0,clip]
        {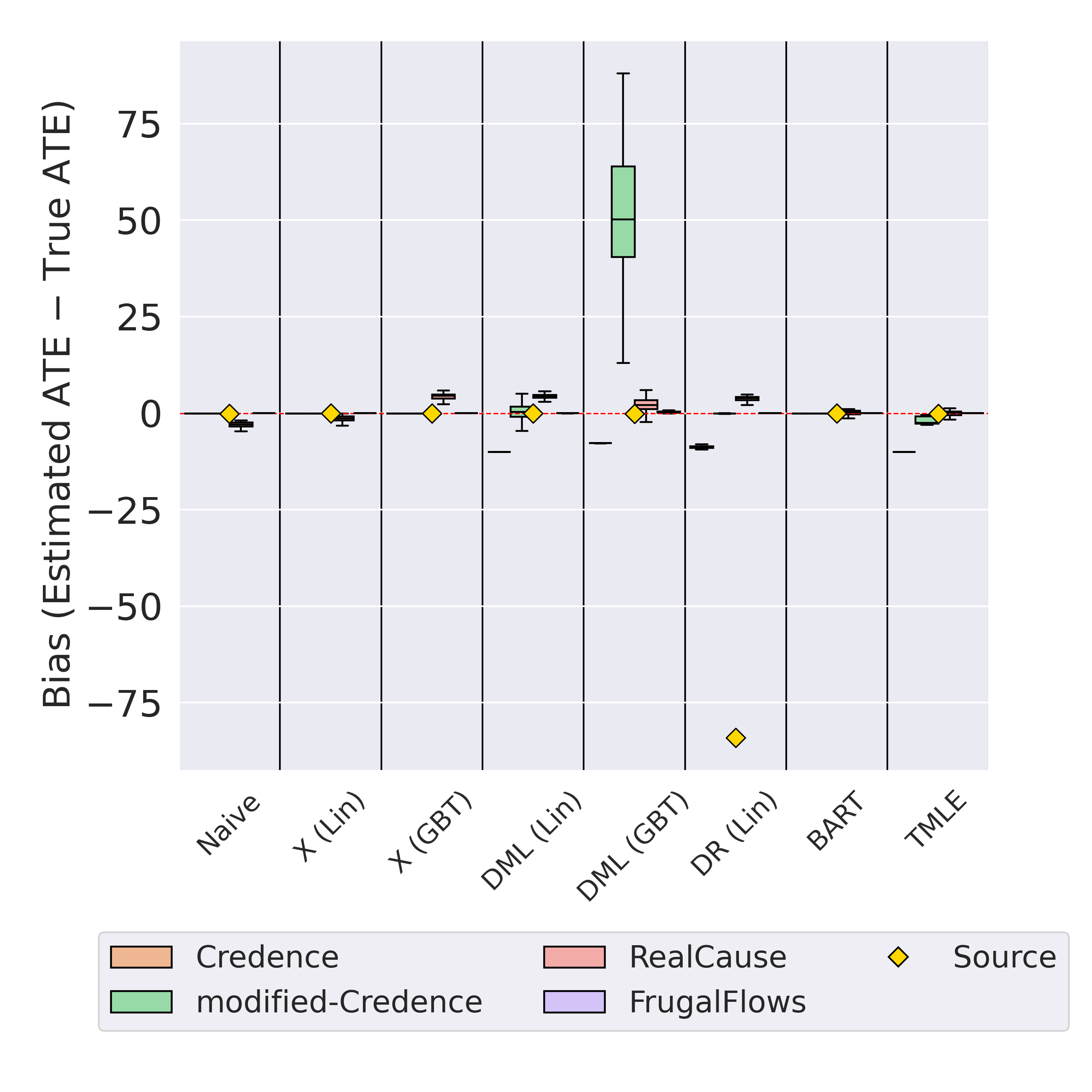}
    }
    \caption{Estimator bias across generative methods for Lalonde under three ATE settings.}
    \label{fig:lalonde-obs-expanded}
\end{figure}

\paragraph{Estimator bias.} Figure~\ref{fig:lalonde-obs} contained the estimator biases for the different settings and methods. We highlighted the differences between the generative methods by zooming in on the estimator bias values that showed differences between the generative methods. However, we notice that for some causal estimators, the variance of the estimator bias is very high. We present the full plot in Figure~\ref{fig:lalonde-obs-expanded}. 

\subsubsection{Postgres}

This is a dataset collected from a computational system, a Postgres database~\citep{gentzel2021and}. Given the nature of the experiment, it is possible to collect both potential outcomes for a given treatment with this dataset. We create an observational equivalent of this dataset by using subsampling to induce a bias between some covariates and the treatment. Since this is a large dataset, we limit the sample size to $3000$ samples. The covariates describe the configurations of the postgres database with the treatment being an indicator function of the index and the outcome being the runtime for that index. We choose to use this dataset in our experiments as the ground-truth ATE estimate is available for us to estimate the bias for each causal estimator. 

\paragraph{Distributional similarity} We denote the mean SWD for the different generative methods and settings of the ATE parameter in Table~\ref{tab:swd-postgres-linear}. We note that Realcause is most sensitive to misspecification of the {\knob} value. 

\begin{table}[htb]
\centering
\small
\begin{tabular}{@{}lcccc@{}}
\toprule
 & Credence & mod-Credence & Realcause & FrugalFlows \\ \midrule
Learned ATE & $0.181\pm 0.007$ & $0.071\pm0.004$ & $0.129\pm0.022$ & $0.228\pm0.014$ \\
True ATE & $0.474\pm0.020$ & $0.142\pm0.021$ & $1.733\pm0.236$ & $0.251\pm0.015$ \\
Incorrect ATE & $1.042\pm0.055$ & $0.472\pm0.065$ & $43.271\pm4.109$ & $0.93\pm0.09$ \\ \bottomrule
\end{tabular}
\caption{SWD for (unscaled) Postgres dataset across generative methods and three settings of the ATE parameter. }
\label{tab:swd-postgres-linear}
\end{table}

\begin{figure}[htb]
    \centering
    \subfigure[Flexible ATE\label{fig:postgres-flexible}]{
        \includegraphics[width=0.31\textwidth,trim=0 1.1cm 0 0,clip]
        {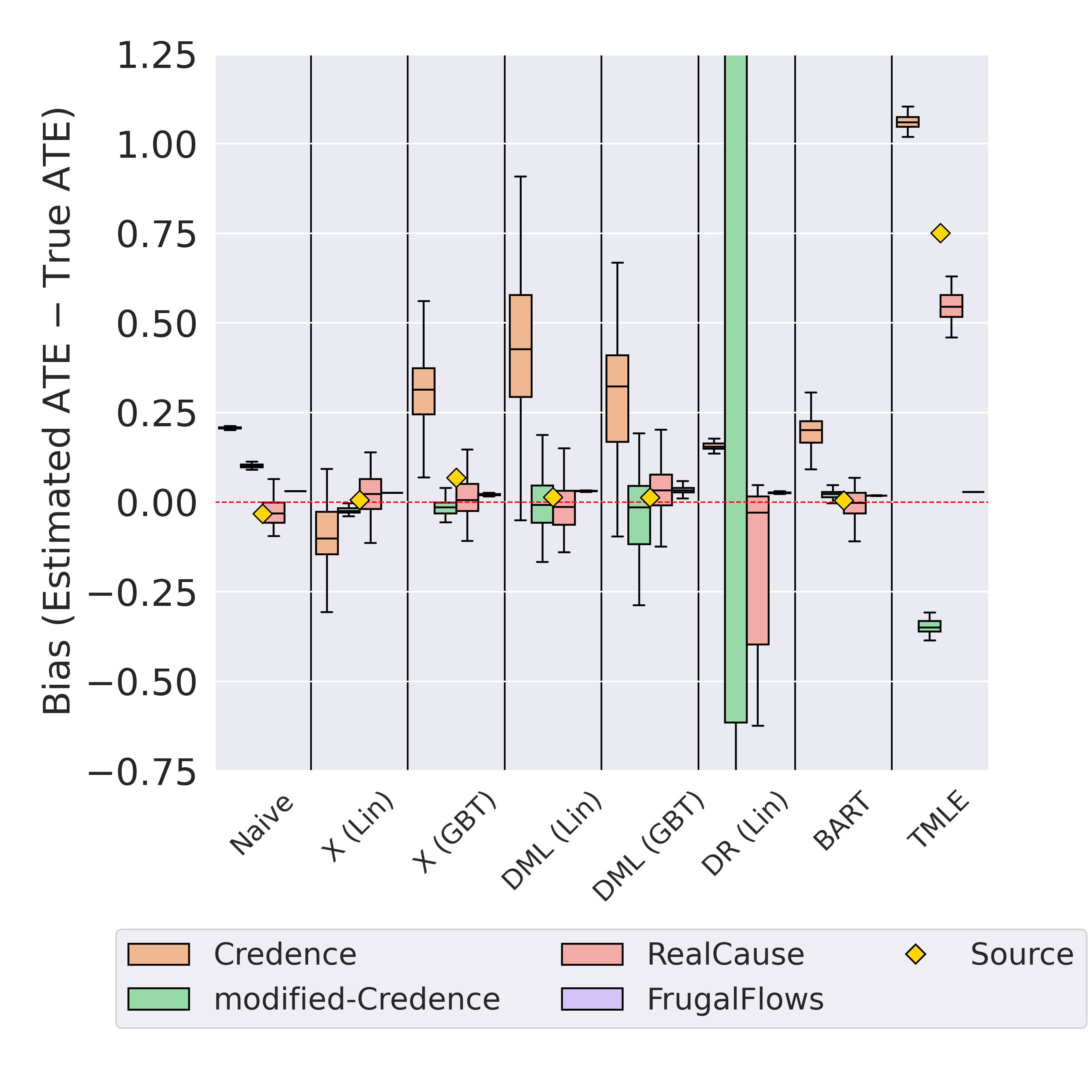}
    }
    \subfigure[True ATE\label{fig:postgres-true}]{
        \includegraphics[width=0.31\textwidth,trim=0 1.1cm 0 0,clip]
        {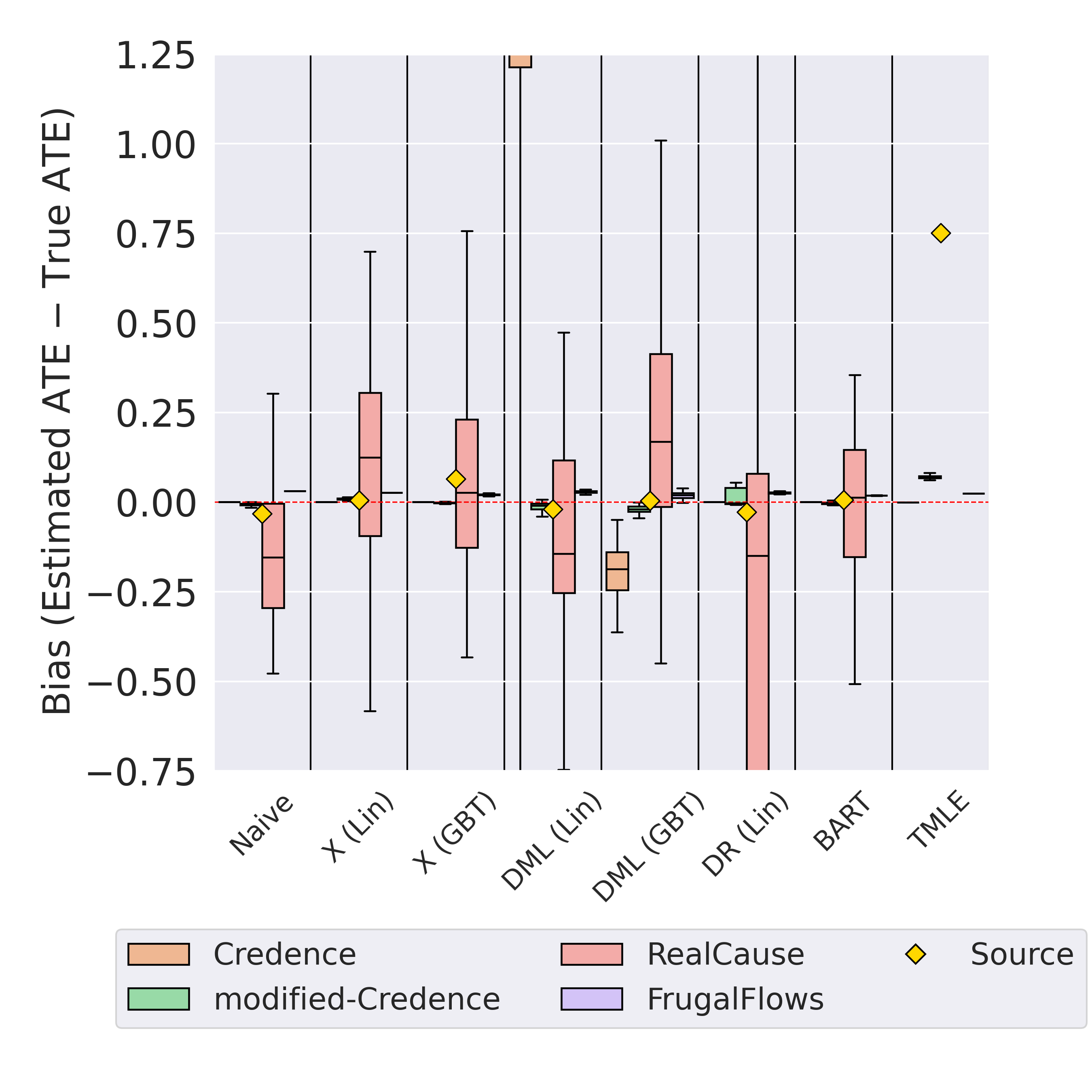}
    }
    \subfigure[Incorrect ATE\label{fig:postgres-incorrect}]{
        \includegraphics[width=0.31\textwidth,trim=0 1.1cm 0 0,clip]
        {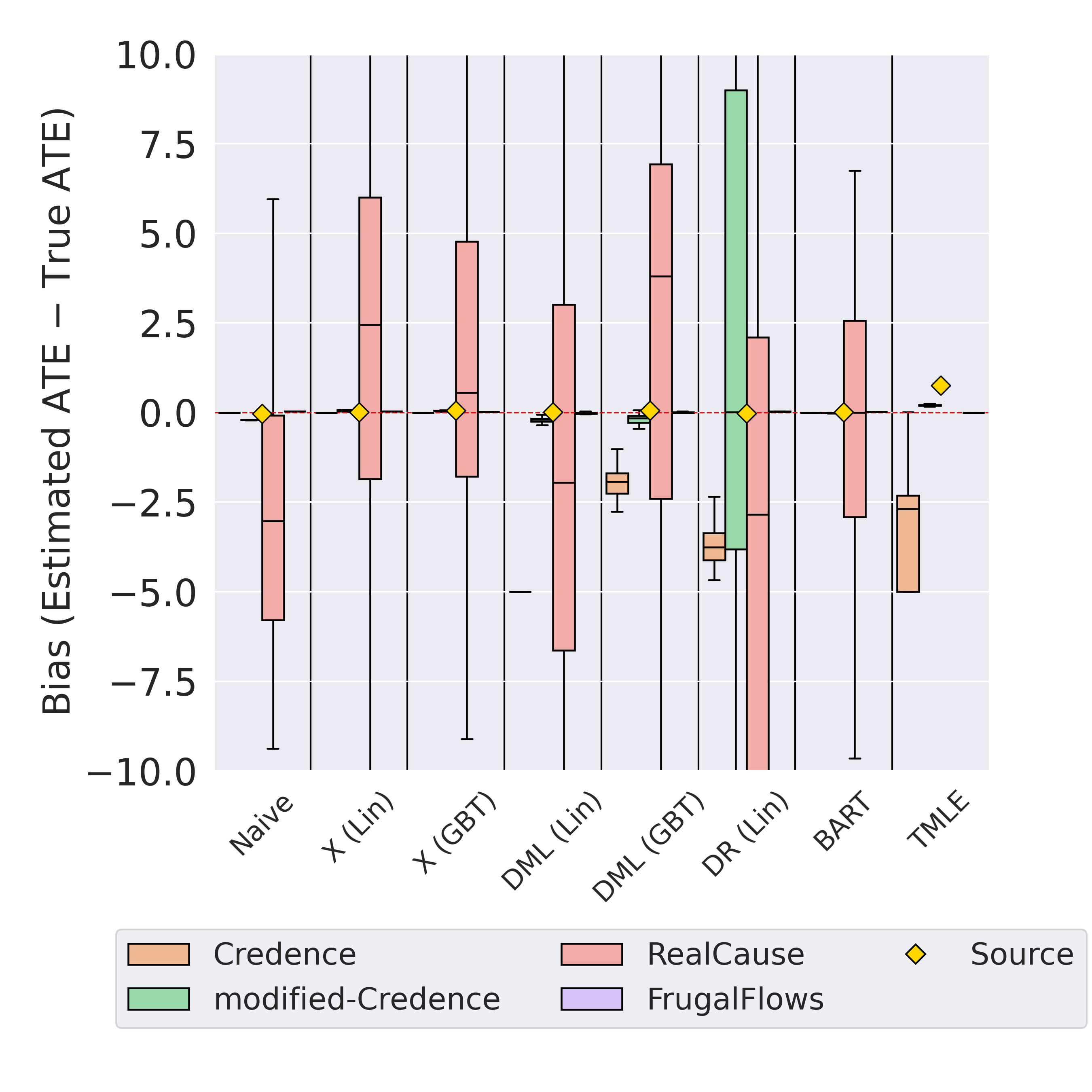}
    }
    \caption{Estimator bias across generative methods for the Postgres dataset under three ATE settings.}
    \label{fig:postgres-linear-bias}
\end{figure}

\paragraph{Estimator bias} We also include the plots of the estimator bias for 50 generated datasets for the different methods and settings of the ATE parameter discussed above~\ref{fig:postgres-linear-bias}. 

\begin{tcolorbox}[colback=gray!5!white, colframe=gray!75!black, title=\textbf{Key takeaway}]
    Across both synthetic and real-world datasets, we observe that different generative methods produce notably different synthetic datasets—even when configured with the same {\knob} constraints. This divergence persists across sample sizes and highlights the role of inductive biases in shaping the generated counterfactual outcomes. These findings support our theoretical proposition that fixed point estimates of {\knob} may be incompatible with the source data. Moreover, inconsistencies in the generated datasets lead to variability in causal estimator outputs, complicating sensitivity analyses and making estimator selection less reliable. These challenges motivate the need for a posterior-based approach like {\ourmethod}, which explicitly accounts for uncertainty in {\knob}s and improves alignment between generated and source data distributions.
\end{tcolorbox}

\section{SBICE: Algorithm and Experimental details}
\label{app:sbice-algorithm}
\subsection{SBICE: Algorithm}

We present the full algorithmic procedure to use {\ourmethod} to evaluate the sensitivity of a set of causal estimators in Algorithm~\ref{alg:sbice-algo}. We provide a pseudocode description of the SMC-ABC algorithm, and point to the original paper for details~\citep{toni2010simulation}.

\begin{algorithm}[t]
\caption{{\ourmethod}: Simulation-Based Inference for Causal Evaluation}
\label{alg:sbice-algo}
\begin{algorithmic}[1]
\REQUIRE Source dataset $D = \{X, T, Y\}$; Generative methods $\mathcal{G} = \{G_1, \ldots, G_K\}$; Method prior $P(G_k)$; Parameter priors $P(\boldsymbol{\theta}_k \mid G_k)$; Distance metric $\delta$; Tolerance schedule $\epsilon_1 > \cdots > \epsilon_T$; Number of particles $N$; Causal estimators $\{\mathcal{M}_1, \ldots, \mathcal{M}_J\}$
\ENSURE Posterior $P(G, \boldsymbol{\theta} \mid D)$; Posterior datasets $\{\hat{D}_{\text{post}}^{(i)}\}_{i=1}^N$; Causal estimates $\{\tau_{\mathcal{M}_j}^{(i)}\}_{i,j}$

\vspace{0.5em}
\STATE \textbf{// Step 1: Train simulators}
\FOR{$k = 1, \ldots, K$}
    \STATE Train generative method $G_k$ on source dataset $D$
\ENDFOR

\vspace{0.5em}
\STATE \textbf{// Step 2: SMC-ABC posterior inference}
\STATE Initialize $N$ particles: $(G^{(i)}, \boldsymbol{\theta}^{(i)}) \sim P(G_k) \, P(\boldsymbol{\theta}_k \mid G_k)$
\FOR{$t = 1, \ldots, T$}
    \FOR{$i = 1, \ldots, N$}
        \STATE Sample method $G^{(i)}$ from current particle weights
        \STATE Perturb parameters: $\boldsymbol{\theta}^{(i)} \sim K_t(\boldsymbol{\theta} \mid \boldsymbol{\theta}^{(i)}_{t-1})$
        \STATE Generate synthetic dataset: $\hat{D}^{(i)} \sim G^{(i)}(\boldsymbol{\theta}^{(i)})$
        \STATE Compute distance: $d^{(i)} = \delta(\hat{D}^{(i)}, D)$
    \ENDFOR
    \STATE Retain particles satisfying $d^{(i)} < \epsilon_t$
    \STATE Compute importance weights and resample
\ENDFOR

\vspace{0.5em}
\STATE \textbf{// Step 3: Generate posterior datasets and compute causal estimates}
\FOR{$i = 1, \ldots, N$}
    \STATE Sample $(G^{(i)}, \boldsymbol{\theta}^{(i)}) \sim P(G, \boldsymbol{\theta} \mid D)$
    \STATE Generate posterior dataset: $\hat{D}_{\text{post}}^{(i)} \sim G^{(i)}(\boldsymbol{\theta}^{(i)})$
    \FOR{$j = 1, \ldots, J$}
        \STATE Compute causal estimate: $\tau_{\mathcal{M}_j}^{(i)} = \mathcal{M}_j(\hat{D}_{\text{post}}^{(i)})$
    \ENDFOR
\ENDFOR

\vspace{0.5em}
\RETURN $P(G, \boldsymbol{\theta} \mid D)$, $\{\hat{D}_{\text{post}}^{(i)}\}_{i=1}^N$, $\{\tau_{\mathcal{M}_j}^{(i)}\}_{i,j}$
\end{algorithmic}
\end{algorithm}

\subsection{Computational Complexity}
\label{app:computational-complexity-sbice}

Our choice of SMC-ABC is motivated by its simplicity and suitability for the low-dimensional parameter spaces considered in current causal simulators (typically fewer than three DGP parameters). Alternative neural SBI methods (e.g., NLE/NPE) require training additional neural density estimators using simulator-generated data, which can compound model misspecification and be sensitive to limited real-world sample sizes. We include an analysis of the computational complexity of SMC-ABC here.

Let $p$ be the number of parameters that are computed here. Since each parameter is one-dimensional, we use $p$ to also denote the effective dimensionality of the parameters. The complexity of running {\ourmethod} depends on the costs associated with each phase of the algorithm. We compute the costs for each phase separately and then describe the total cost.

\begin{enumerate}
    \item Simulator inference cost: Let $C_{\text{sim}}$ be the cost associated with sampling from the simulator. This would vary depending on the simulator in question, but we show an example for the NSF (normalizing spline flows) generative model. The cost of inference from an NSF is given by $C_{\text{sim}} = O(F \times p \times S)$ where $F$ is the number of flow layers and $S$ is the number of splines. In almost all cases, we restricted the number of flow layers and splines to $\sim 10$.
    \item Distance computation cost: The cost of computing the sliced-Wasserstein distance for the tabular dataset that we have typically is linear in the number of projections used. We represent this cost as a fixed value $D$.
    \item SMC-ABC updates: The naive algorithm used to update the weights in SMC-ABC is $O(N^2)$, where $N$ is the number of particles. By using adaptive algorithms, this can be computed in $O(N)$ time. Note that $N \propto \epsilon^{-p}$ for a tolerance of $\epsilon$, which means that the acceptance probability (which determines the number of particles) decreases exponentially with the dimensionality of the parameters used.
\end{enumerate}

Using $T$ to denote the tolerance levels (or number of iterations), we have the cost of {\ourmethod} as $O(T\epsilon^{-2p} + C_{\text{sim}}T\epsilon^{-p})$. In our experiments, we found that for $p \leq 3$ using $T\leq 15, \epsilon \geq 0.001$ produced posterior datasets with minimal error and with computation times averaging 3 hours when running experiments parallelized across $\sim 10$ samplers on CPUs with 16G RAM.

\subsection{Experimental Details}

We implement SMC-ABC using the Python library \texttt{pyABC}~\citep{schaelte2022pyabc}. The hyperparameter and algorithmic choices required to run SMC-ABC are outlined in Table~\ref{tab:pyabc-details}. All of our experiments were run on a HPC cluster with distributed samplers to make sampling of datasets from the simulator more efficient. 

\begin{table}[ht]
    \centering
    \caption{Hyperparameters and Implementation choices for SMC-ABC. \\}
    
    \label{tab:pyabc-details}
    \begin{tabular}{lc} 
    \toprule
        Hyperparameter & Range of values \\ \midrule
        Maximum number of iterations & $[12, 15]$ \\
        Minimum epsilon threshold & $[0.001, 0.005]$ \\
        Distance metric & Sliced-Wasserstein (L2 norm; $100$ projections) \\
    \bottomrule
    \end{tabular}
\end{table}

We used the following set of causal estimators in our experiments. 

\begin{enumerate}
    \item X-Learner~\citep{kunzel2019metalearners}: A meta-learner algorithm to estimate the average treatment effect using two different underlying function: Linear regression (referred to as X (Lin)) and Gradient Boosted Trees (referred to as X (GBT)). We used the implementation in the \texttt{EconML}~\citep{econml} package.
    \item Double machine learning (DML)~\citep{chernozhukov21w}: An algorithm that constructs a de-biased estimator of the causal parameter by using two models to estimate the residual errors. We used the implementation in the \texttt{EconML} python package, and used two learners: Linear regression (DML (Lin)) and Gradient Boosted Trees (DML (GBT)).
    \item Doubly robust (DR)~\citep{dudik2014dr}: This estimator combines two models: one for outcome regression and another for the treatment (propensity score) to estimate the causal effect. The advantage of using this estimator is that the effect is unbiased if either model is correctly specified. We use the implementation in the \texttt{EconML} package and use a linear model for both the treatment and outcome functions (DR (Lin)). 
    \item Causal BART~\citep{hill2011}: This estimator leverages Bayesian Additive Regression Trees (BART) to estimate the causal effect. We use its R implementation~\citep{dbarts2025dorie} in our experiments.
    \item Targeted Maximum Likelihood Estimator (TMLE)~\citep{van2006targeted}: We use the implementation of TMLE as described in the \texttt{zEpid} package. 
\end{enumerate}

\section{Evaluating {\ourmethod} on synthetic datasets generated using parametric data-generating processes}
\label{app:sbice-parametric-dgps}


We evaluate {\ourmethod} on a suite of synthetic datasets generated from simple, parametric data-generating processes. These experiments assess when the inferred posterior yields more informative datasets than those generated by sampling from the prior. We systematically vary characteristics of the data-generating process and simulator by introducing model misspecification, using non-identifiable and partially identifiable {\knob}s. We label the synthetic datasets generated in this manner as LinearParam DGP($x$), with $x$ being the experiment identifier. 

For each experiment, we specify the data-generating process, simulator design, set of {\knob}s, and user-defined prior. To evaluate {\ourmethod}, we compute the bias squared error (BSE) of several causal estimators applied to these datasets, with lower values for the posterior indicating better fidelity to the source distribution. 

\subsection{LinearParam Sim1: Linear, parametric data-generating process and simulator}
\label{app:lp-dgp1}

In this setting, we consider a linear, parametric data generating process (DGP) involving a single unobserved confounder $Z$, an observed confounder $X$, a binary treatment $T$, and a continuous outcome $Y$. 

\paragraph{LinearParam DGP1}
\begin{equation}
\begin{aligned}
    Z \sim \mathcal{N}(0, 1) \\
    X \sim \mathcal{N}(0, 1) \\
    T \sim \text{Binomial}(\rho Z + \beta X + \mathcal{N}(0, 0.1)) \\
    Y = \rho Z + \beta X + \tau T + \mathcal{N}(0, 0.1)
\end{aligned}
\label{dgp:parametric-linear}
\end{equation}

We assume that the simulator is correctly specified, i.e., it closely matches the data generating process.

\paragraph{LinearParam Sim1}
\begin{equation}
\begin{aligned}
    Z \sim \mathcal{N}(0, 1) \\
    X = X \\
    T \sim \text{Binomial}(\rho Z + \beta X +  \mathcal{N}(0, 0.1)) \\
    Y = \rho Z + \beta X + \tau T + \mathcal{N}(0, 0.1) 
\end{aligned}
\label{sim:parametric-linear}
\end{equation}

This setting has three {\knob}s: $\rho$ (unobserved confounding), $\beta$ (observed confounding), and $\tau$ (causal effect). We assume that they are independent and therefore use uniform, non-overlapping priors for them. The true parameter values used in the DGP are $\rho = 1.0$, $\beta = -1.5$, and $\tau = 1.5$. 

\paragraph{Prior}
\begin{equation}
    \begin{aligned}
        \text{Prior}(\rho) \sim U[0.0, 2.0] \\
        \text{Prior}(\beta) \sim U[-2.0, 1.0] \\
        \text{Prior}(\tau) \sim U[0.0, 2.0] \\
    \end{aligned}
\label{prior:parametric-linear}
\end{equation} 

We generate three sets of 50 datasets each:
\begin{enumerate}
    \item \textbf{Source datasets} from the true DGP: LinearParam DGP1,
    \item \textbf{Posterior datasets} using parameter samples from the posterior distribution, and
    \item \textbf{Prior datasets} using parameter samples from the prior.
\end{enumerate}

\paragraph{Evaluation}

\begin{figure}[ht]
\centering
\begin{minipage}{0.49\textwidth}
    \centering
    \includegraphics[scale=0.4]{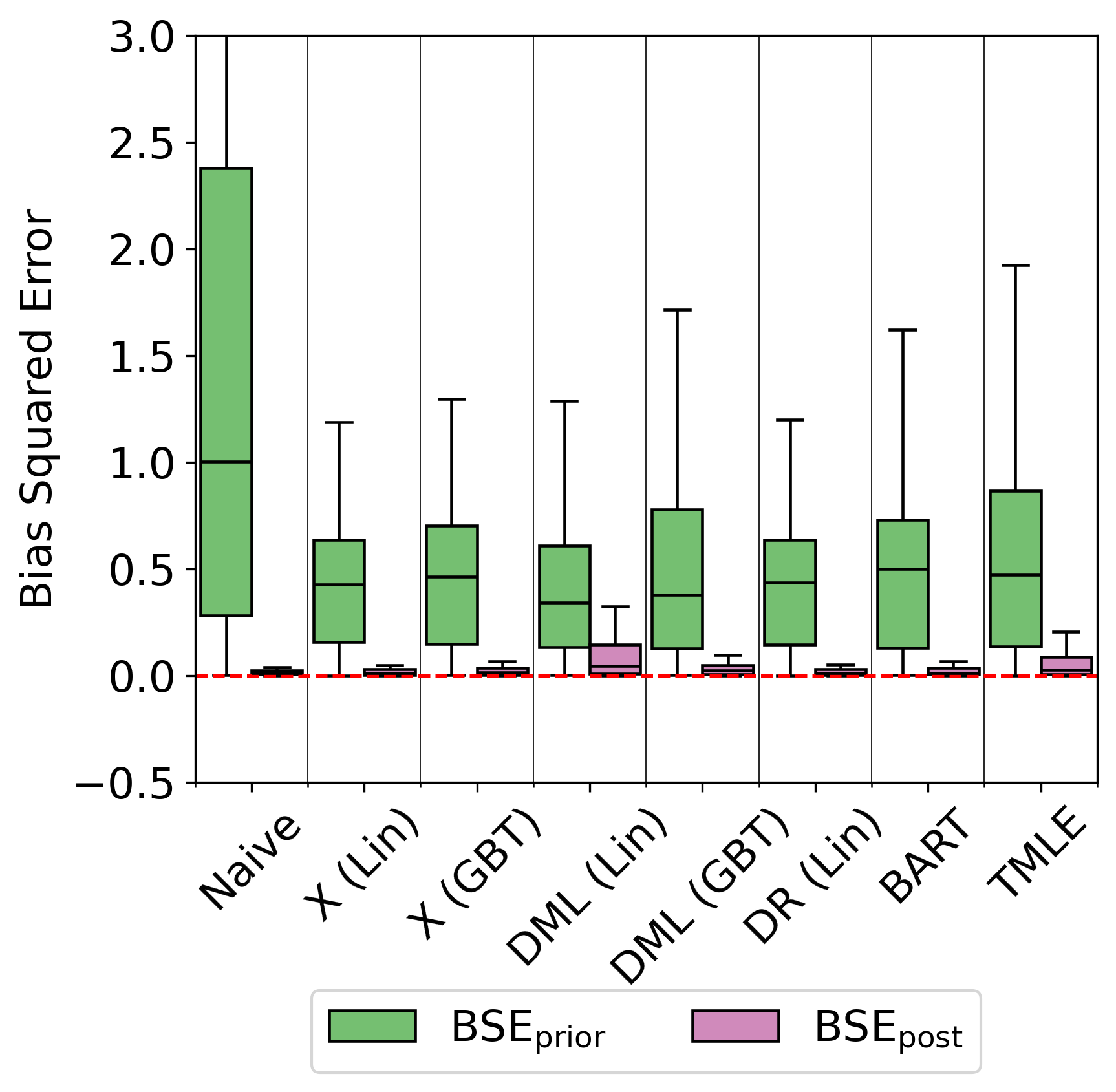}
\end{minipage}%
\begin{minipage}{0.47\textwidth}
    \centering
        \begin{tabular}{lcc}
        \toprule
        & Prior & Posterior \\ \midrule
        Diff. of Means & 1.382 & 0.018 \\
        X (Lin) & 0.575 & 0.015 \\
        X (GBT) & 0.587 & 0.020 \\
        DML (Lin) & 0.551 & 0.083 \\
        DML (GBT) & 0.564 & 0.028 \\
        DR (Lin) & 8e5 & 0.016 \\
        BART & 0.590 & 0.018 \\
        TMLE & 0.624 & 0.065\\
        \bottomrule
        \end{tabular}
\end{minipage}
\caption{BSE for DGP: LinearParam DGP1 and Simulator: LinearParam Sim1.}
\label{fig:results-lp-dgp1}
\end{figure}

\begin{figure}[ht]
    \centering
    \includegraphics[width=0.75\linewidth]{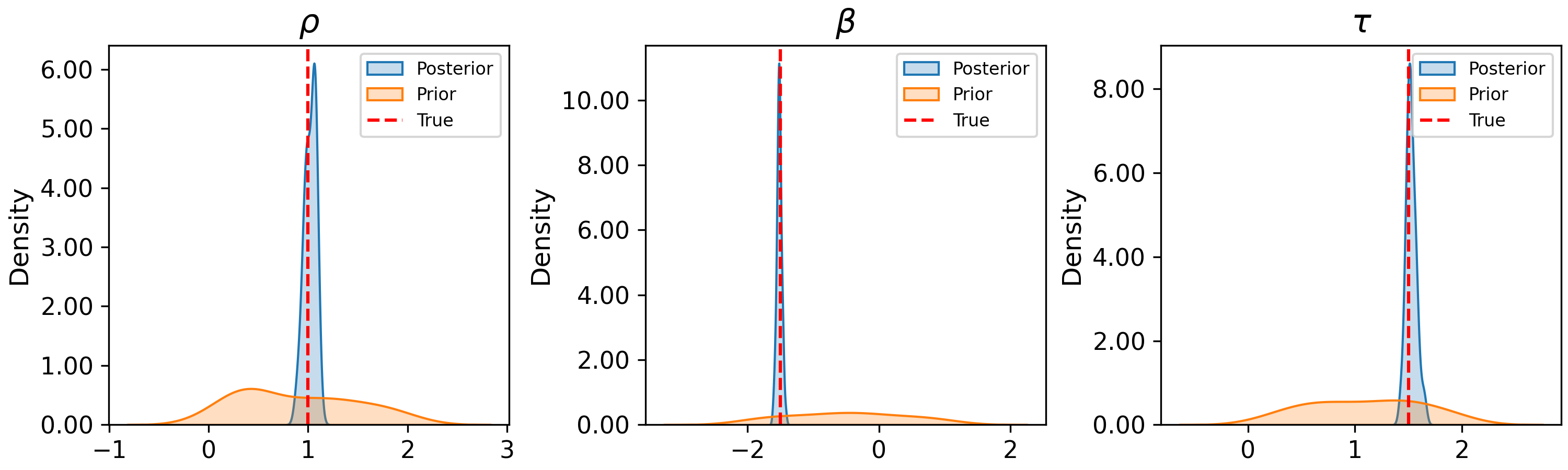}
    \caption{Posterior distribution for DGP: LinearParam DGP1, Simulator: LinearParam Sim1}
    \label{fig:post-lp-dgp1}
\end{figure}

We report the mean BSE for the posterior and prior datasets and also plot the BSE of the causal estimators for the posterior, prior and source datasets in Figure~\ref{fig:results-lp-dgp1}. In addition, we plot the posterior distribution for each of the {\knob}s in Figure~\ref{fig:post-lp-dgp1}. For this setting, we find that {\ourmethod} improves posterior estimates and that posterior datasets are informative of the performance of causal estimators.  

\subsection{Misspecified simulators}
\label{app:model-misspecification}

However, in real-world applications, the true data generating process (DGP) is typically unknown, and the simulator specified by the researcher may be misspecified. To understand the effects of such misspecification in parameterized linear models, we investigate how different types of model mismatch influence the estimated posterior distribution and subsequent performance of causal estimators. We consider several categories of misspecification, including variations in the form of unobserved confounding, increased noise levels, linear approximations to nonlinear functions, and the inclusion of interaction terms. Each variant of the DGP and its corresponding simulator is described below.

\paragraph{LinearParam Sim2: Noisy simulator}

In this setting, we examine the effect of additive, Gaussian noise to the outcome model within the simulator. We use the same DGP as before: LinearParam DGP1, but use the simulator LinearParam Sim2 described below. The true parameter values are $\rho = 1.0$, $\beta = -1.5$, and $\tau = 1.5$.

\paragraph{LinearParam Sim2}
\begin{equation}
    \begin{aligned}
        Z \sim \mathcal{N}(0, 1) \\
        X = X \\
        T \sim \text{Binomial}(\rho  Z + \beta  X +  \mathcal{N}(0, 0.1)) \\
        Y = \rho  Z + \beta  X  + \tau  T + \mathcal{N}(0, \textbf{1.0}) 
    \end{aligned}
\end{equation}

\paragraph{Prior}

\begin{equation}
\label{eq:prior-2a}
    \begin{aligned}
        \text{Prior}(\rho) \sim U[-5.0, 5.0] \\
        \text{Prior}(\beta) \sim U[-5.0, 5.0] \\
        \text{Prior}(\tau) \sim U[-5.0, 5.0] 
    \end{aligned}
\end{equation}

\begin{figure}[h]
\centering
\begin{minipage}{0.49\textwidth}
    \centering
    \includegraphics[scale=0.4]{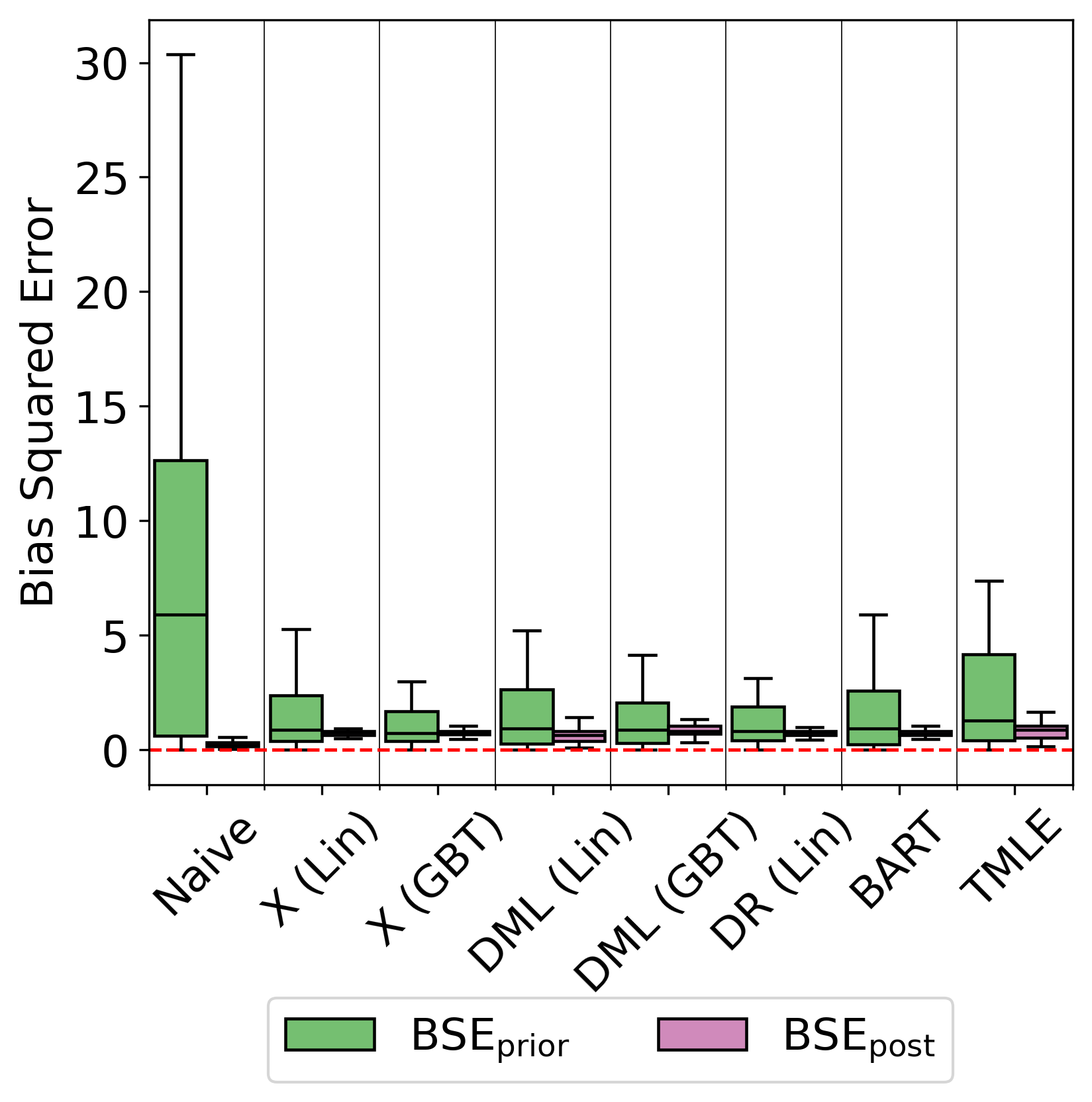}
\end{minipage}%
\begin{minipage}{0.47\textwidth}
    \centering
        \begin{tabular}{lcc}
        \toprule
         & Prior & Posterior \\ \midrule
        Diff. of Means & 8.964 & 0.237 \\
        X (Lin) & 1.568 & 0.720 \\
        X (GBT) & 1.324 & 0.740 \\
        DML (Lin) & 1.815 & 0.660 \\
        DML (GBT) & 1.847 & 0.847 \\
        DR (Lin)  & 1.520 & 0.729 \\
        BART & 1.699 & 0.728 \\
        TMLE & 5.069 & 0.843\\
        \bottomrule
        \end{tabular}
\end{minipage}
\caption{BSE for DGP: LinearParam DGP1 and Simulator: LinearParam Sim2.}
\label{fig:results-lp-dgp2}
\end{figure}

\begin{figure}[h]
    \centering
    \includegraphics[width=0.75\linewidth]{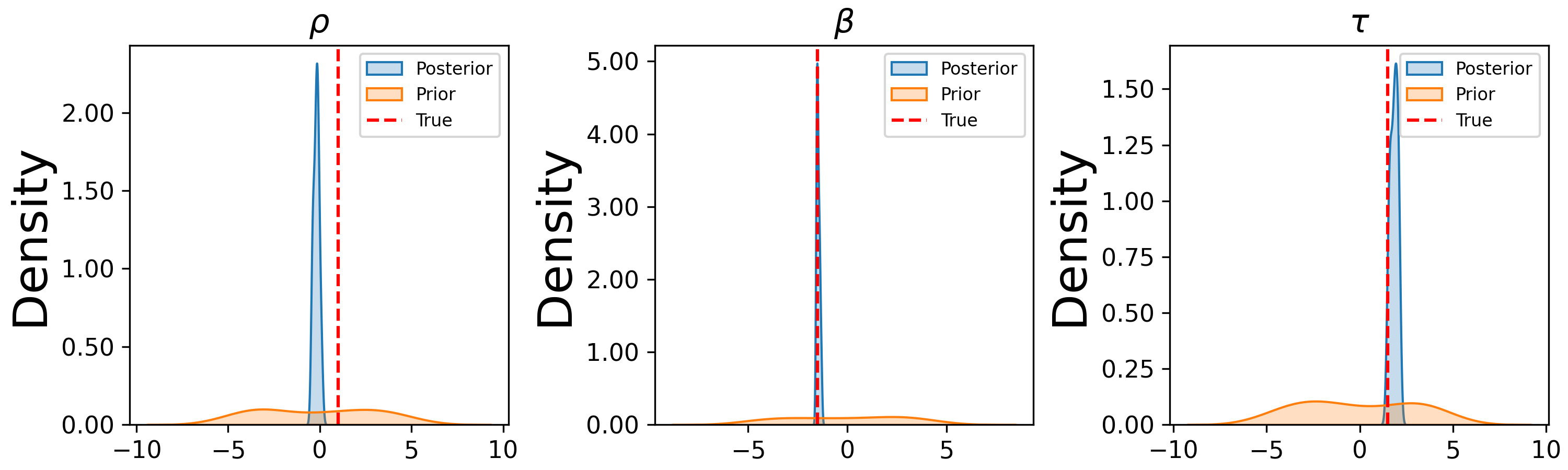}
    \caption{Posterior distribution for DGP: LinearParam DGP1, Simulator: LinearParam Sim2}
    \label{fig:post-lp-dgp2}
\end{figure}

\paragraph{Evaluation} We present the results of our evaluation for this setting in Figures~\ref{fig:results-lp-dgp2} and \ref{fig:post-lp-dgp2}. We find that despite the noisy simulator, the BSE of the posterior estimates is lower compared to the prior, highlighting the robustness of {\ourmethod} to moderate misspecification. 

\paragraph{LinearParam Sim3: Misspecified unobserved confounder ($Z$)}

In this setting, we investigate the effect of misspecifying the distribution of the unobserved confounder $Z$. We use LinearParam DGP1 as the data-generating process but misspecify the distribution of the unobserved confounder $Z$. The true parameters are $\rho = 1.0$, $\beta = -1.5$, and $\tau = 1.5$. We use the same priors as in Linear Sim2 (Equation~\ref{eq:prior-2a}). The unobserved confounder $Z$ is modeled using an exponential distribution rather than the true Gaussian distribution. 

\paragraph{LinearParam Sim3}
\begin{equation}
    \begin{aligned}
        Z \sim \textbf{Exponential}(0.5) \\
        X = X \\
        T \sim \text{Binomial}(\rho  Z + \beta  X +  \mathcal{N}(0, 0.1)) \\
        Y = \rho  Z + \beta X + \tau  T + \mathcal{N}(0, 0.1) 
    \end{aligned}
\end{equation}

\begin{figure}[h]
\centering
\begin{minipage}{0.49\textwidth}
    \centering
    \includegraphics[scale=0.4]{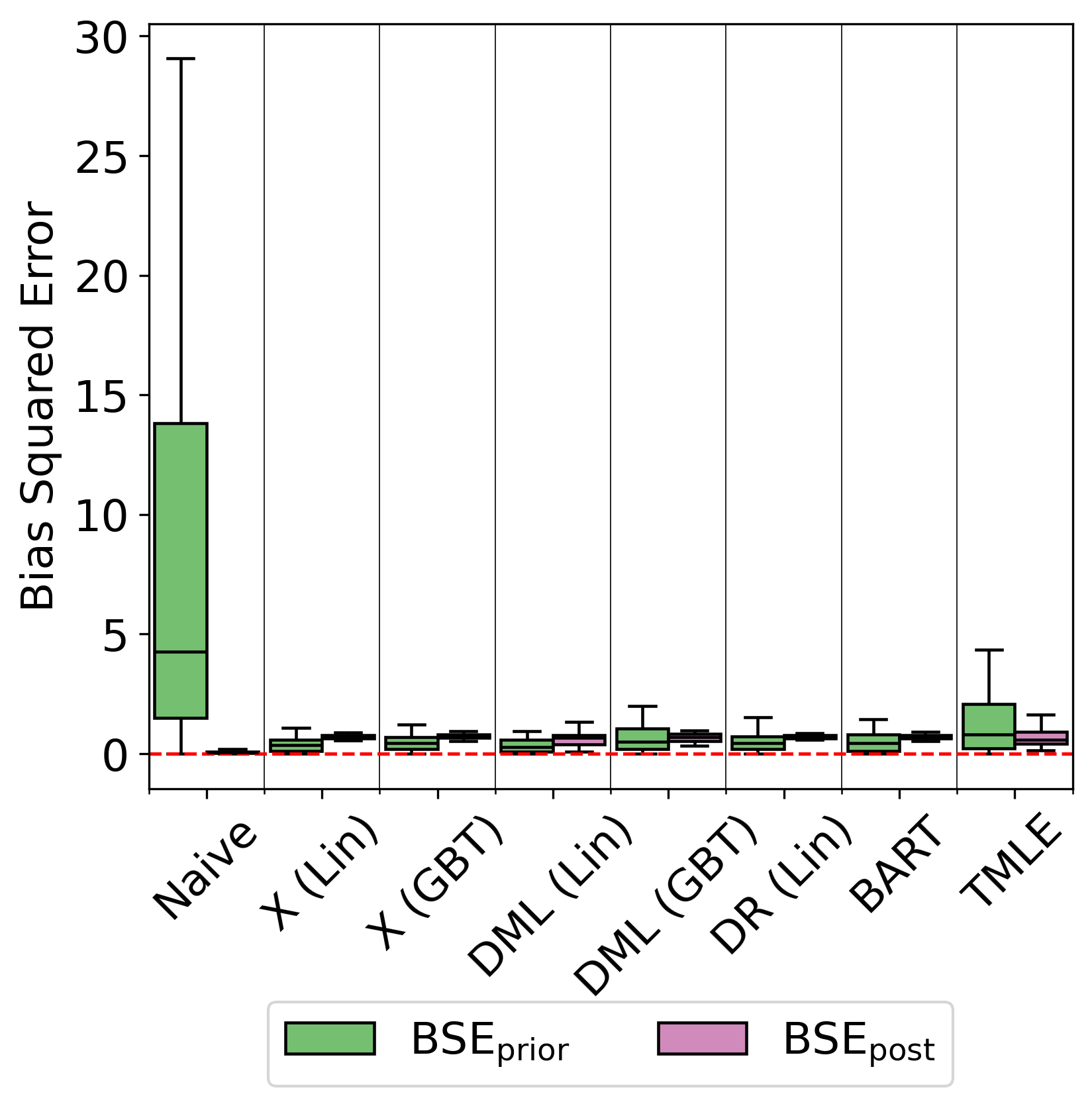}
\end{minipage}%
\begin{minipage}{0.47\textwidth}
    \centering
        \begin{tabular}{lcc}
        \toprule
         & Prior & Posterior \\ \midrule
        Diff. of Means & 8.984 & 0.072 \\
        X (Lin) & 0.491 & 0.709 \\
        X (GBT) & 0.614 & 0.728 \\
        DML (Lin) & 0.474 & 0.642 \\
        DML (GBT) & 1.709 & 0.691 \\
        DR (Lin)  & 30.894 & 0.715 \\
        BART & 0.805 & 0.722 \\
        TMLE & 2.310 & 0.693\\
        \bottomrule
        \end{tabular}
\end{minipage}
\caption{BSE for DGP: LinearParam DGP1 and Simulator: LinearParam Sim3.}
\label{fig:results-lp-dgp3}
\end{figure}

\begin{figure}[h]
    \centering
    \includegraphics[width=0.75\linewidth]{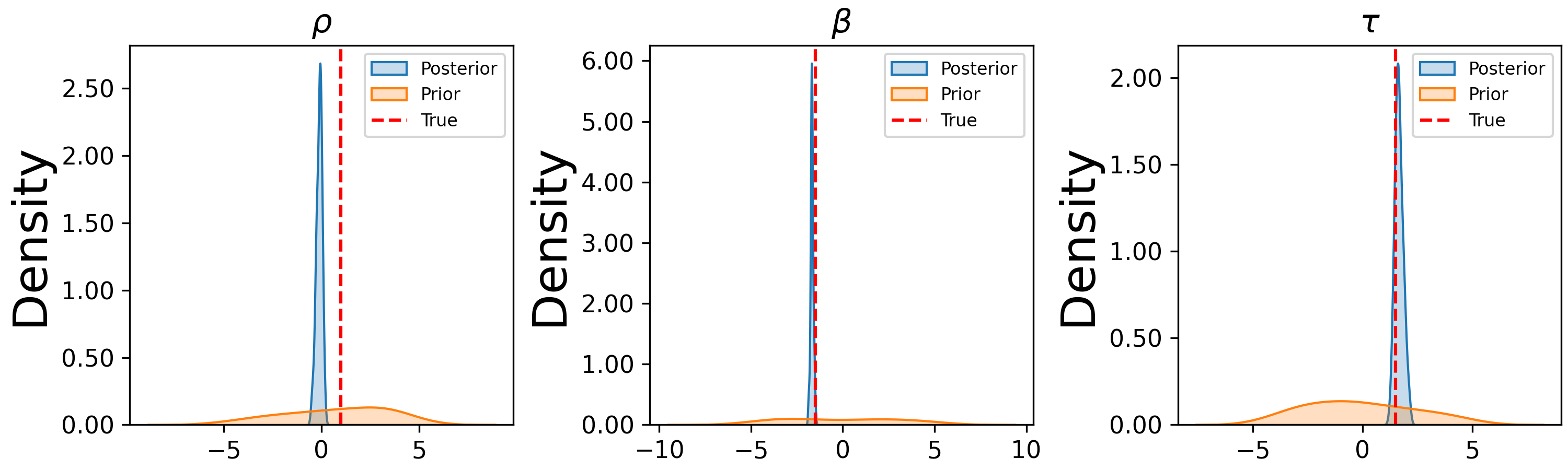}
    \caption{Posterior distribution for DGP: LinearParam DGP1, Simulator: LinearParam Sim3}
    \label{fig:post-lp-dgp3}
\end{figure}

\paragraph{Evaluation} This experiment evaluates the robustness of {\ourmethod} when the unobserved confounder deviates substantially from the assumed distribution. Notably, in this setting, the posterior distribution over $\rho$ concentrates near zero (see Figure~\ref{fig:post-lp-dgp3}), effectively down-weighting the influence of the misspecified confounder in order to better match the source data distribution. Figure~\ref{fig:results-lp-dgp3} illustrates the estimator bias and the posterior/prior distributions over parameters. The mean BSE for the posterior estimates remain low, as the variance of the posterior distribution is low. 

\paragraph{LinearParam Sim4: Misspecified simulator with additional interaction terms}

In this setting, we explore the impact of misspecifying the functional form of the simulator by introducing an interaction term between the observed and unobserved confounders, $X$ and $Z$. The true parameter values and prior distributions are identical to those used with LinearParam Sim2 (Equation~\ref{eq:prior-2a}). The simulator is defined as follows. 

\paragraph{LinearParam Sim4}
\begin{equation}
    \begin{aligned}
        Z \sim \mathcal{N}(0, 1) \\
        X = X \\
        T \sim \text{Binomial}(\rho  Z +\beta X + X \cdot Z + \mathcal{N}(0, 0.1)) \\
        Y = \rho  Z +\beta X + \tau T + X \cdot Z + \mathcal{N}(0, 0.1) 
    \end{aligned}
\end{equation}

\begin{figure}[h]
\centering
\begin{minipage}{0.49\textwidth}
    \centering
    \includegraphics[scale=0.4]{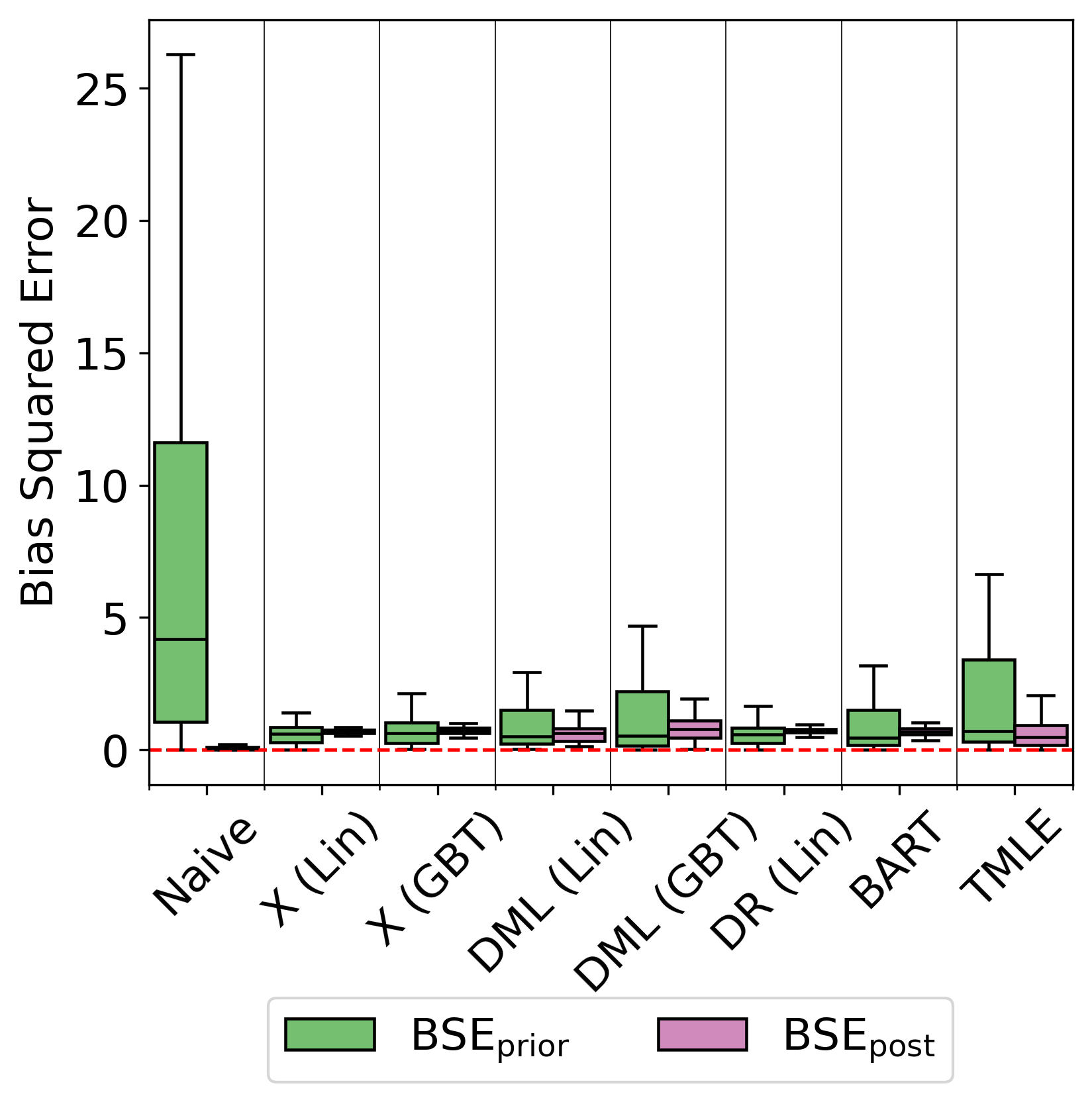}
\end{minipage}%
\begin{minipage}{0.47\textwidth}
    \centering
        \begin{tabular}{lcc}
        \toprule
         & Prior & Posterior \\ \midrule
        Diff. of Means & 8.375 & 0.079 \\
        X (Lin) & 1.212 & 0.700 \\
        X (GBT) & 1.142 & 0.724 \\
        DML (Lin) & 1.313 & 0.633 \\
        DML (GBT) & 2.407 & 0.858 \\
        DR (Lin)  & 1.148 & 0.731 \\
        BART & 1.222 & 0.680 \\
        TMLE & 2.362 & 0.688\\
        \bottomrule
        \end{tabular}
\end{minipage}
\caption{BSE for DGP: LinearParam DGP1 and Simulator: LinearParam Sim4.}
\label{fig:results-lp-dgp4}
\end{figure}

\begin{figure}[h]
    \centering
    \includegraphics[width=0.75\linewidth]{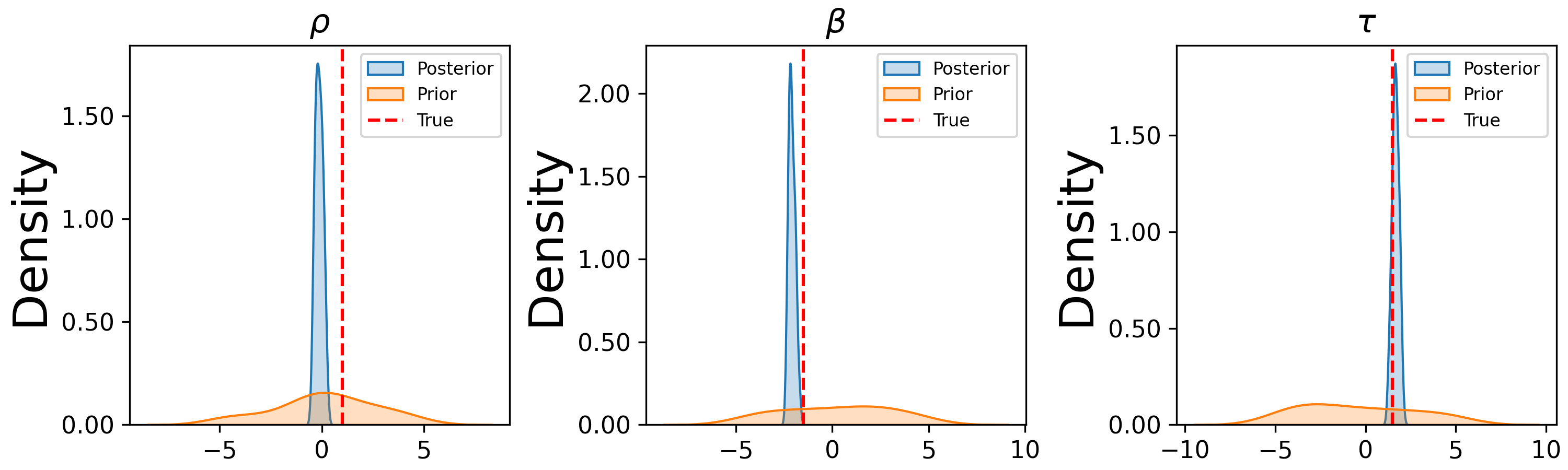}
    \caption{Posterior distribution for DGP: LinearParam DGP1, Simulator: LinearParam Sim4}
    \label{fig:post-lp-dgp4}
\end{figure}

\paragraph{Evaluation} The interaction term $X \cdot Z$, which is absent in the true DGP (LinearParam DGP1), introduces a form of structural misspecification in both the treatment and outcome models of the simulator. Figure~\ref{fig:results-lp-dgp4} displays the results. Despite the added complexity, {\ourmethod} is able to adjust for the mismatch to some extent by adapting the posterior over $\boldsymbol{\theta}$, though some deviation in estimator bias for the posterior datasets remains due to the unmodeled interaction. We include the posterior distribution in Figure~\ref{fig:post-lp-dgp4}.

\paragraph{LinearParam Sim5: Misspecified simulator with linear outcome function}

In this setting, we explore a structural mismatch where the true outcome function is nonlinear—specifically, a polynomial function of the covariates and treatment—while the simulator assumes a linear relationship. This setup reflects a common form of model misspecification in applied settings, where complex interactions or nonlinearities in the data are simplified in simulation. The true DGP is defined as follows.

\paragraph{LinearParam DGP5}
\begin{equation}
    \begin{aligned}
        Z \sim \mathcal{N}(0, 1) \\
        X \sim \mathcal{N}(0, 1) \\
        T \sim \text{Binomial}(\rho Z + \beta X + \mathcal{N}(0, 0.1)) \\
        Y = \rho (Z^2 + Z \cdot X) + \beta (X^2 - X \cdot T) + \tau  T + \mathcal{N}(0, 0.1) 
    \end{aligned}
\end{equation}

The simulator, by contrast, is the same as LinearParam Sim1 (Equation~\ref{sim:parametric-linear}), and does not capture the true polynomial structure of the outcome. The true parameter values and uniform priors remain the same as in LinearParam Sim2 (Equation~\ref{eq:prior-2a}). This experiment allows us to evaluate how well {\ourmethod} handles misspecification arising from oversimplified functional forms in the simulator.

\begin{figure}[h]
\centering
\begin{minipage}{0.49\textwidth}
    \centering
    \includegraphics[scale=0.4]{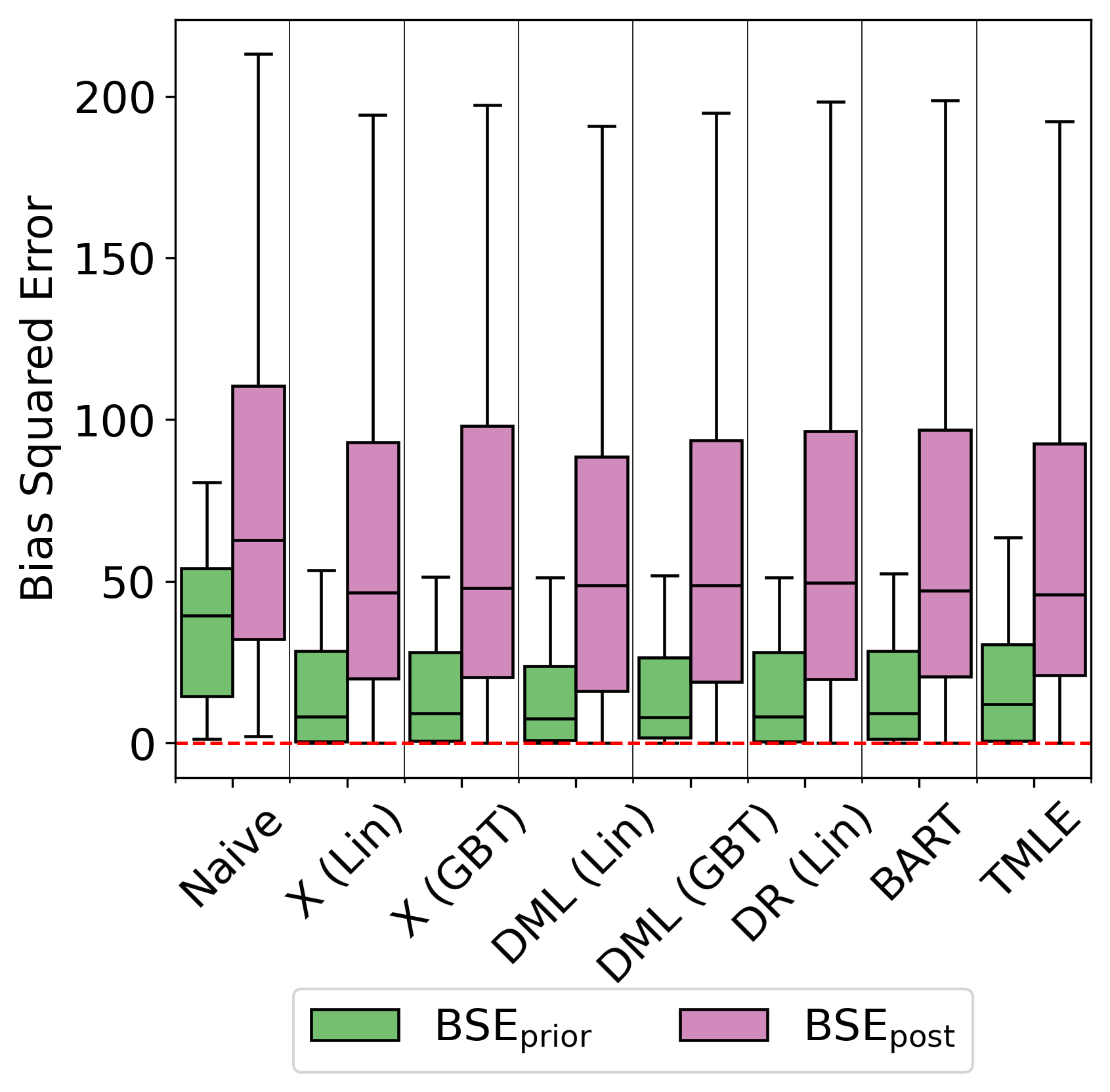}
\end{minipage}%
\begin{minipage}{0.47\textwidth}
    \centering
        \begin{tabular}{lcc}
        \toprule
        & Prior & Posterior \\ \midrule
        Diff. of Means & 37.990 & 77.401 \\
        X (Lin) & 15.124 & 62.779 \\
        X (GBT) & 15.722 & 63.065 \\
        DML (Lin) & 13.822 & 62.020 \\
        DML (GBT) & 16.163 & 63.475 \\
        DR (Lin)  & 8e6 & 62.929 \\
        BART & 16.416 & 63.297 \\
        TMLE & 18.377 & 62.499\\
        \bottomrule
        \end{tabular}
\end{minipage}
\caption{BSE for DGP: LinearParam DGP5 and Simulator: LinearParam Sim1.}
\label{fig:results-lp-dgp5}
\end{figure}

\begin{figure}[h]
    \centering
    \includegraphics[width=0.75\linewidth]{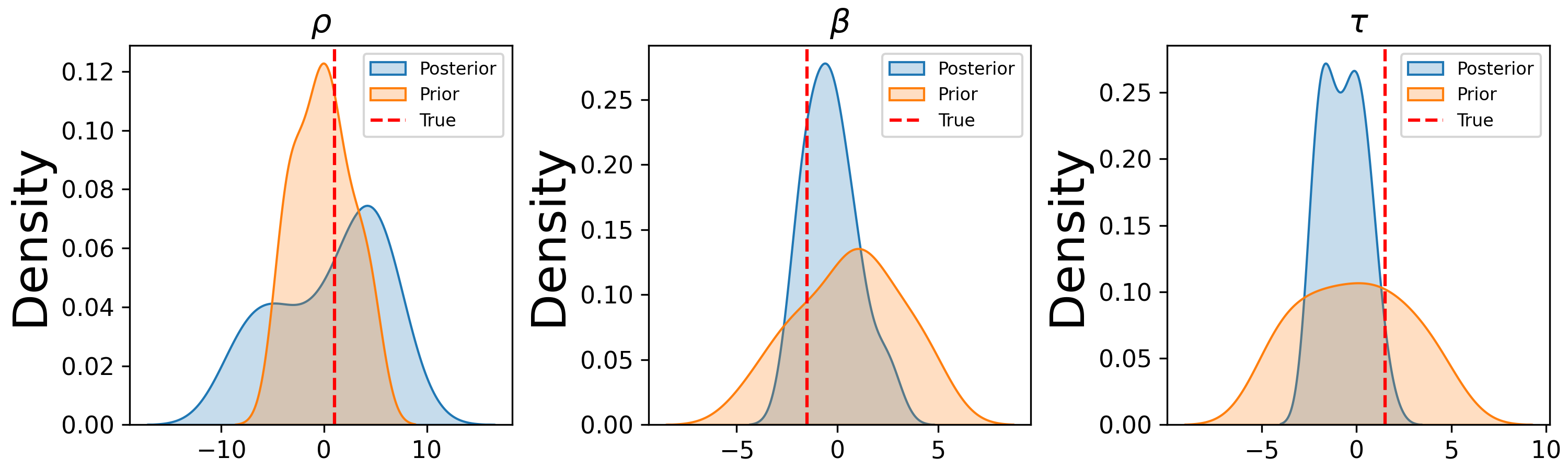}
    \caption{Posterior distribution for DGP: LinearParam DGP5, Simulator: LinearParam Sim1}
    \label{fig:post-lp-dgp5}
\end{figure}

\paragraph{Evaluation} The results are shown in Figures~\ref{fig:results-lp-dgp5} and \ref{fig:post-lp-dgp5}, which compares estimator bias and parameter distributions for datasets generated from the posterior and prior. In this setting, we find that both the posterior and prior estimates are different compared to the source distribution (indicated by the high BSE) and high variance across all estimates. We also find that {\ourmethod} is unable to recover the true parameters from the data.

\textbf{Key takeaway:} With misspecified simulators, the causal estimates on posterior datasets do not align with the source data—-sometimes performing worse than the prior—-depending on the nature of the misspecification. This highlights the importance of using flexible simulators that can accurately capture the source data distribution and independently vary the {\knob}s during simulation. Such flexibility is crucial for ensuring meaningful inference under misspecification.

\subsection{Non-identifiable parameters}
We analyze the cases where the observed data distribution provides no information about the values of {\knob}s.

\paragraph{LinearParam Sim6: Non-identifiable parameters from observed data} In this setting, we constrain the sum of two parameters: $\rho, \tau$. This leads to a fundamental non-identifiability in the model. In this formulation, $T$ is deterministically derived from $Z$ and a uniform random noise term. As a result, the joint distribution of $(T, Y)$ remains invariant for any pair of values $(\rho, \tau)$ such that their sum $\rho + \tau = c$ is constant. While individual values of $\rho$ and $\tau$ are not identifiable from the data, their sum is identifiable. Consequently, any prior over $\rho$ and $\tau$ that maintains a constant sum will yield a posterior distribution that is also uniform across that constraint. The data generating process is described below.

\paragraph{LinearParam DGP6}

\begin{equation}
\label{eq:dgp-6}
    \begin{aligned}
        Z \sim \text{Binomial}(0.5) \\
        X \sim \mathcal{N}(0,1) \\
        T = \mathbbm{1}(Z + U[0, 0.5)) \\
        Y = \rho Z + \beta X + \tau T + U[0, 0.5)
    \end{aligned}
\end{equation}

We assume that the simulator is correctly specified and matches the structure of the true DGP. The true parameters are $\rho = 2.0$, $\beta = 0.5$, and $\tau = 2.0$. The simulator is specified as:

\paragraph{LinearParam Sim6}
\begin{equation}
    \begin{aligned}
        Z \sim \text{Binomial}(0.5) \\
        X = X \\
        T = \mathbbm{1}(Z + U[0, 0.5)) \\
        Y = \rho Z + \beta X + \tau T + U[0, 0.5)
    \end{aligned}
\end{equation}

We use uniform, independent priors over all three parameters: 

\paragraph{Prior}
\begin{equation}
\label{eq:prior-6}
    \begin{aligned}
        \text{Prior}(\rho) \sim U[0.0, 10.0] \\
        \text{Prior}(\beta) \sim U[0.0, 10.0] \\
        \text{Prior}(\tau) \sim U[0.0, 10.0] \\
    \end{aligned}
\end{equation}

\begin{figure}[h]
\centering
\begin{minipage}{0.49\textwidth}
    \centering
    \includegraphics[scale=0.4]{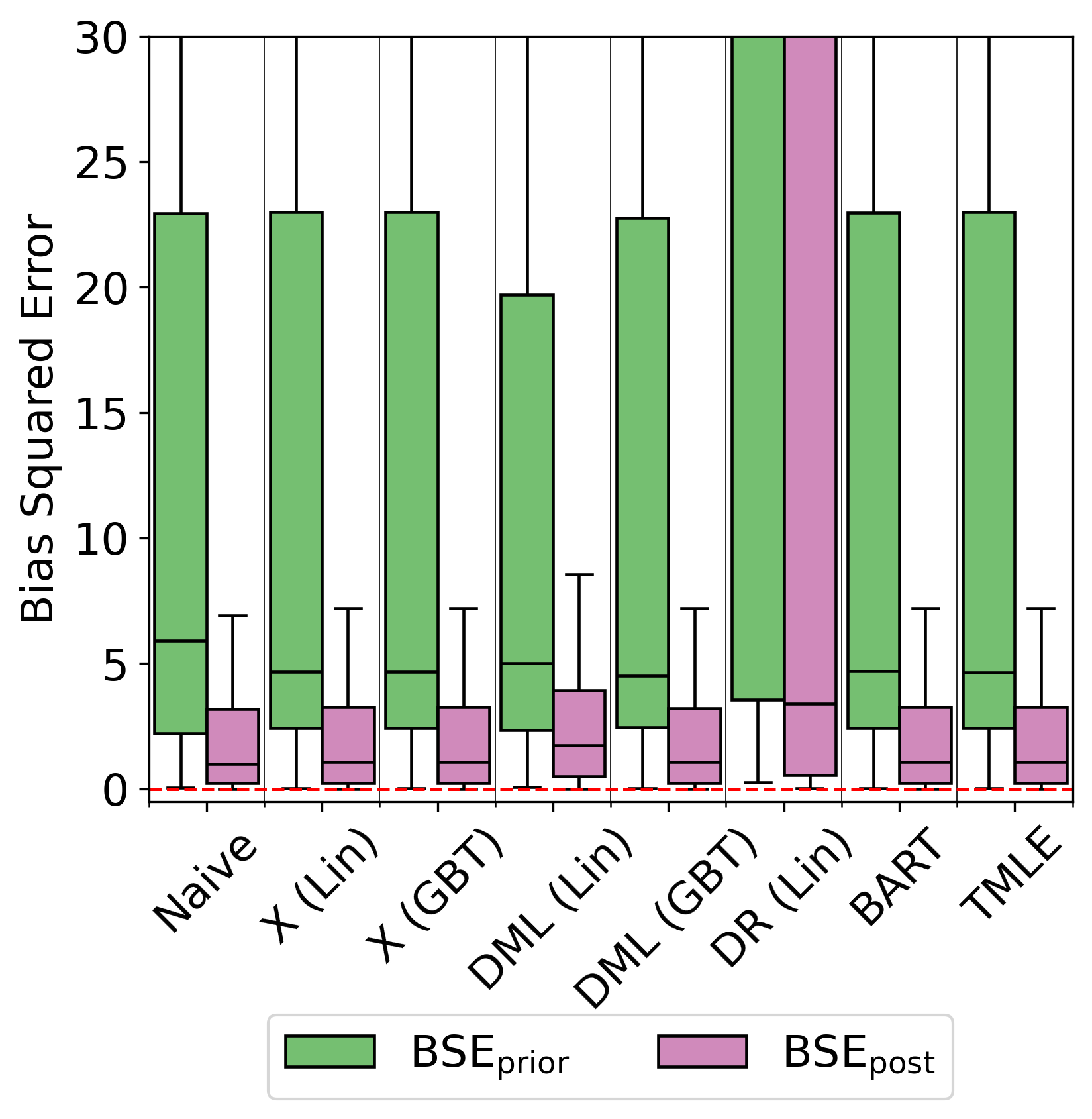}
\end{minipage}%
\begin{minipage}{0.47\textwidth}
    \centering
        \begin{tabular}{lcc}
        \toprule
        & Prior & Posterior \\ \midrule
        Diff. of Means & 15.952 & 2.040 \\
        X (Lin) & 16.078 & 2.054 \\
        X (GBT) & 16.081 & 2.055 \\
        DML (Lin) & 14.988 & 2.723 \\
        DML (GBT) & 15.894 & 2.048 \\
        DR (Lin)  & 1e6 & 1e6 \\
        BART & 16.085 & 2.054 \\
        TMLE & 16.081 & 2.054\\
        \bottomrule
        \end{tabular}
\end{minipage}
\caption{BSE for DGP: LinearParam DGP6 and Simulator: LinearParam Sim6.}
\label{fig:results-lp-dgp6}
\end{figure}

\begin{figure}[h]
    \centering
    \includegraphics[width=0.75\linewidth]{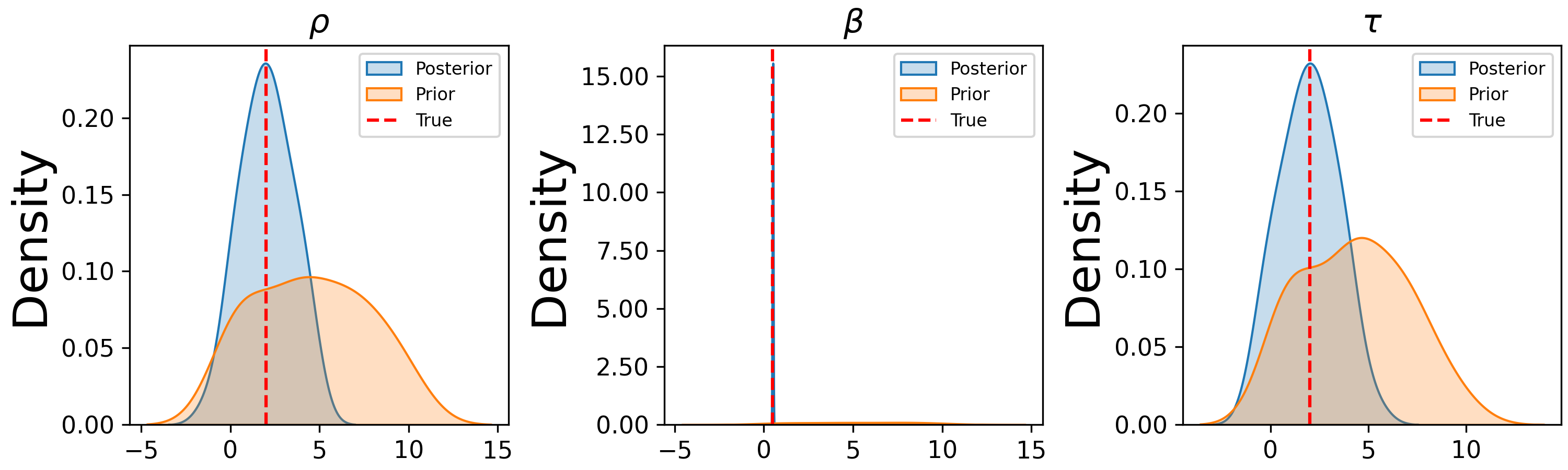}
    \caption{Posterior distribution for DGP: LinearParam DGP6, Simulator: LinearParam Sim6}
    \label{fig:post-lp-dgp6}
\end{figure}

\paragraph{Evaluation} This setup allows us to analyze the implications of parameter non-identifiability on the posterior distributions and causal estimator performance. Figure~\ref{fig:results-lp-dgp6} displays the BSE for posterior and prior datasets. Despite the non-identifiability, the posterior estimates are similar to the source estimates as we still have information on $\beta$ learned from the source data. The posterior also converges to a distribution of values such that the sum of $\rho + \tau$ is a constant. This distribution is displayed in Figure~\ref{fig:post-lp-dgp6}.

\paragraph{LinearParam Sim7: Using joint priors with constrained sum} We revisit the non-identifiability scenario from LinearParam Sim6 with a key modification: instead of using independent priors over $\rho$ and $\tau$, we impose a joint prior distribution that explicitly enforces the constraint $\rho + \tau = c$, where $c$ is a known, user-specified constant. This reflects a setting in which the researcher has domain knowledge about the functional relationship between parameters, even if the individual values remain unknown.

Our hypothesis is that under this constraint, the posterior and prior datasets may exhibit similar behavior, as both are restricted to lie on the same manifold in parameter space. Consequently, any observed differences can be primarily attributed to better posterior estimation of the identifiable parameter $\beta$. The true parameter values used to generated the observed data are $\rho = 1.0, \beta = 0.3, \tau = 2.0$. The joint prior is defined as follows. 

\paragraph{Prior}
\begin{equation}
\label{eq:joint-prior}
    \begin{aligned}
        \text{Prior}(\rho) \sim U[-5.0, 5.0] \\
        \text{Prior}(\beta) \sim  U[0.0, 5.0] \\
        \text{Prior}(\tau) \sim  U[-20.0, 20.0] \\
        \text{Constraint: } \rho + \tau = c, c = 3.0
    \end{aligned}
\end{equation}

\begin{figure}[h]
\centering
\begin{minipage}{0.49\textwidth}
    \centering
    \includegraphics[scale=0.4]{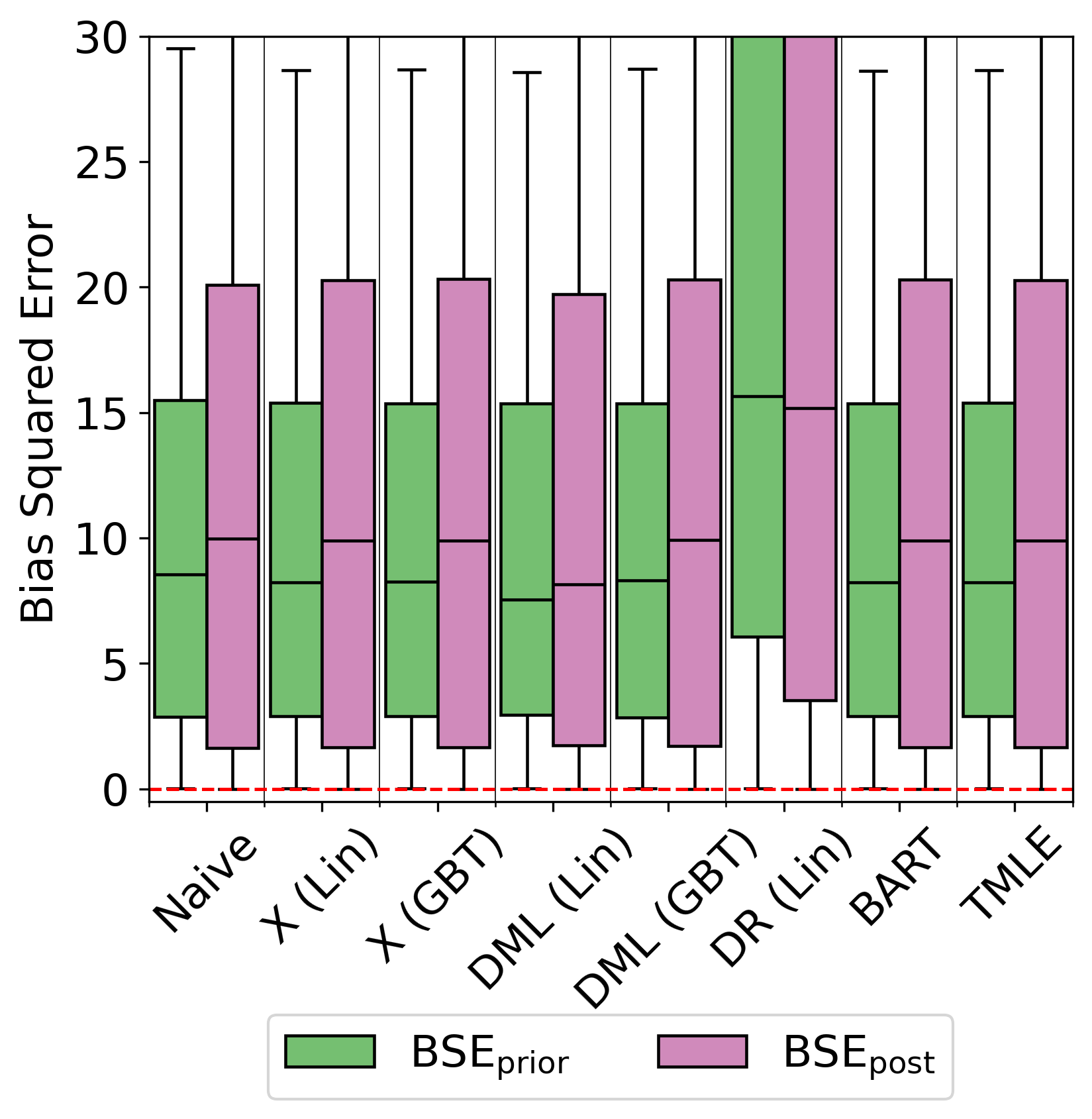}
\end{minipage}%
\begin{minipage}{0.47\textwidth}
    \centering
        \begin{tabular}{lcc}
        \toprule
        & Prior & Posterior \\ \midrule
        Diff. of Means & 10.737 & 13.110 \\
        X (Lin) & 10.860 & 13.120 \\
        X (GBT) & 10.864 & 13.121 \\
        DML (Lin) & 11.230 & 13.011 \\
        DML (GBT) & 10.856 & 13.145 \\
        DR (Lin)  & 1e6 & 1e6 \\
        BART & 10.856 & 13.122 \\
        TMLE & 10.860 & 13.120\\
        \bottomrule
        \end{tabular}
\end{minipage}
\caption{BSE for DGP: LinearParam DGP6 and Simulator: LinearParam Sim7 (which is similar to Sim6 with updated prior distribution).}
\label{fig:results-lp-dgp7}
\end{figure}

\begin{figure}[h]
    \centering
    \includegraphics[width=0.75\linewidth]{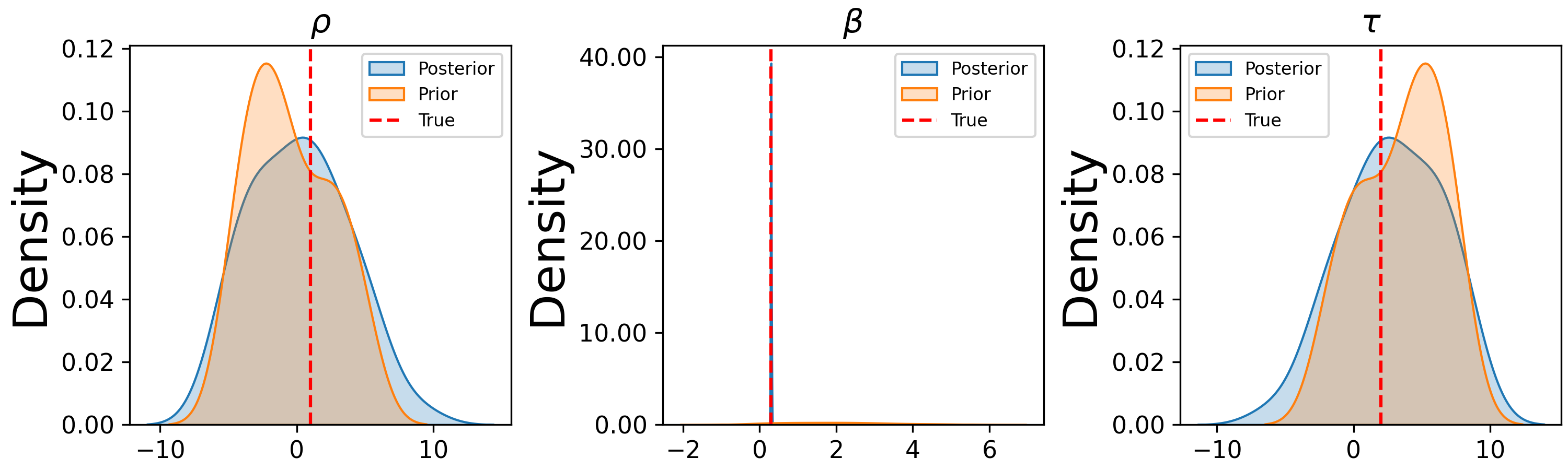}
    \caption{Posterior distribution for DGP: LinearParam DGP6, Simulator: LinearParam Sim7 (similar to LinearParam Sim6 with joint priors).}
    \label{fig:post-lp-dgp7}
\end{figure}

Formally, the joint prior distribution over $(\rho, \tau)$ is defined as $P(\rho, \tau) = P(\rho) \cdot P(\tau) \text{ iff } \rho + \tau = c$ else $P(\rho, \tau) = 0$. During simulation, we sample values for $\rho$ from its marginal prior and determine $\tau$ accordingly to satisfy the constraint. The mean BSE in this setting is visualized in Figure~\ref{fig:results-lp-dgp7} and~\ref{fig:post-lp-dgp7}. 

Contrary to our initial hypothesis, imposing a joint prior that enforces a known relationship between parameters does not lead to a more informative posterior. Instead, we find that the constraint restricts the flexibility of the posterior distribution, limiting its ability to align with the source data. Both the posterior and prior datasets exhibit high BSE across all estimators. These results suggest that, in scenarios with non-identifiable parameters, incorporating partial knowledge through hard constraints on the {\knob}s may hinder rather than help. It may be preferable to allow the inference process to explore a broader parameter space without such restrictions.

\paragraph{LinearParam Sim8: Partial identifiability of parameters} Earlier, we examined a case where the parameters $\rho$ and $\tau$ were non-identifiable due to a symmetry in the data generating process (DGP): the treatment assignment depended solely on the unobserved confounder $Z$, and only the sum $\rho + \tau$ could be inferred. To make the setting more realistic and address the identifiability issue, we now introduce a modified DGP where treatment $T$ depends not only on $Z$ but also on the observed covariate $X$. This modification helps break the symmetry and allows for partial identifiability of the individual parameters. The DGP is given by: 

\paragraph{LinearParam DGP8}
\begin{equation}
    \begin{aligned}
        Z \sim \text{Binomial}(0.5) \\
        X \sim \mathcal{N}(0, 1) \\
        T = \mathbbm{1}(Z + 0.4 X + U[0.0, 0.5)) \\
        Y = \rho Z + \beta X + \tau T + U[0, 0.5)
    \end{aligned}
\end{equation}

We use the following true parameter values: $\rho = 1.0, \beta = 0.3, \tau = 2.0$. The priors are uniform, independent random variables:

\paragraph{Prior}
\begin{equation}
    \begin{aligned}
        \text{Prior}(\rho) \sim U[0.0, 10.0] \\
        \text{Prior}(\beta) \sim U[0.0, 10.0] \\
        \text{Prior}(\tau) \sim U[0.0, 10.0] \\
    \end{aligned}
\end{equation}

\begin{figure}[h]
\centering
\begin{minipage}{0.49\textwidth}
    \centering
    \includegraphics[scale=0.4]{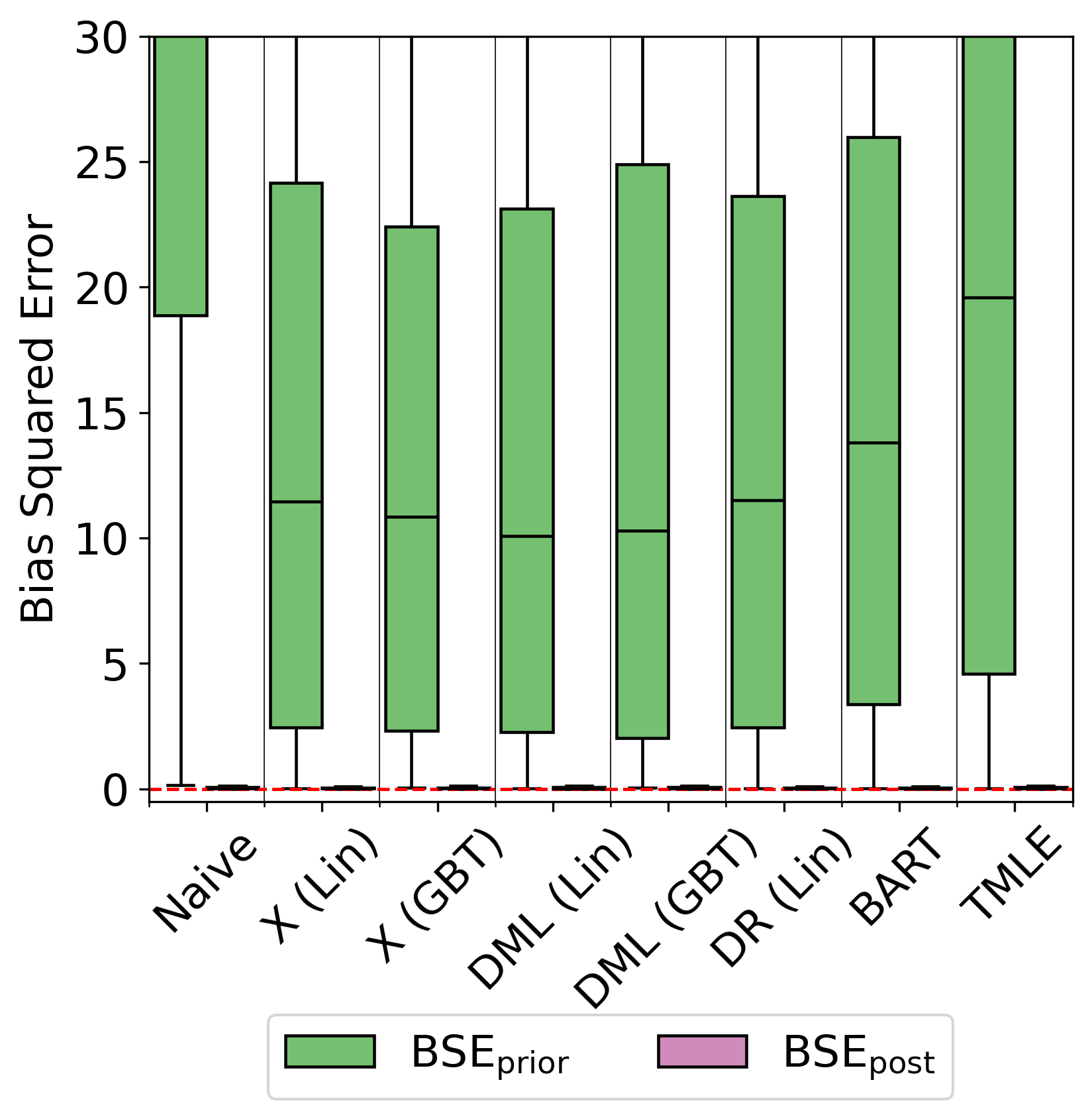}
\end{minipage}%
\begin{minipage}{0.47\textwidth}
    \centering
        \begin{tabular}{lcc}
        \toprule
        & Prior & Posterior \\ \midrule
        Diff. of Means & 58.401 & 0.044 \\
        X (Lin) & 14.985 & 0.034 \\
        X (GBT) & 13.974 & 0.034 \\
        DML (Lin) & 14.053 & 0.034 \\
        DML (GBT) & 13.784 & 0.046 \\
        DR (Lin)  & 14.914 & 0.033 \\
        BART & 16.374 & 0.034 \\
        TMLE & 24.021 & 0.043\\
        \bottomrule
        \end{tabular}
\end{minipage}
\caption{BSE for DGP: LinearParam DGP8 and Simulator: LinearParam Sim8 (which is similar to Sim6).}
\label{fig:results-lp-dgp8}
\end{figure}

\begin{figure}[h]
    \centering
    \includegraphics[width=0.75\linewidth]{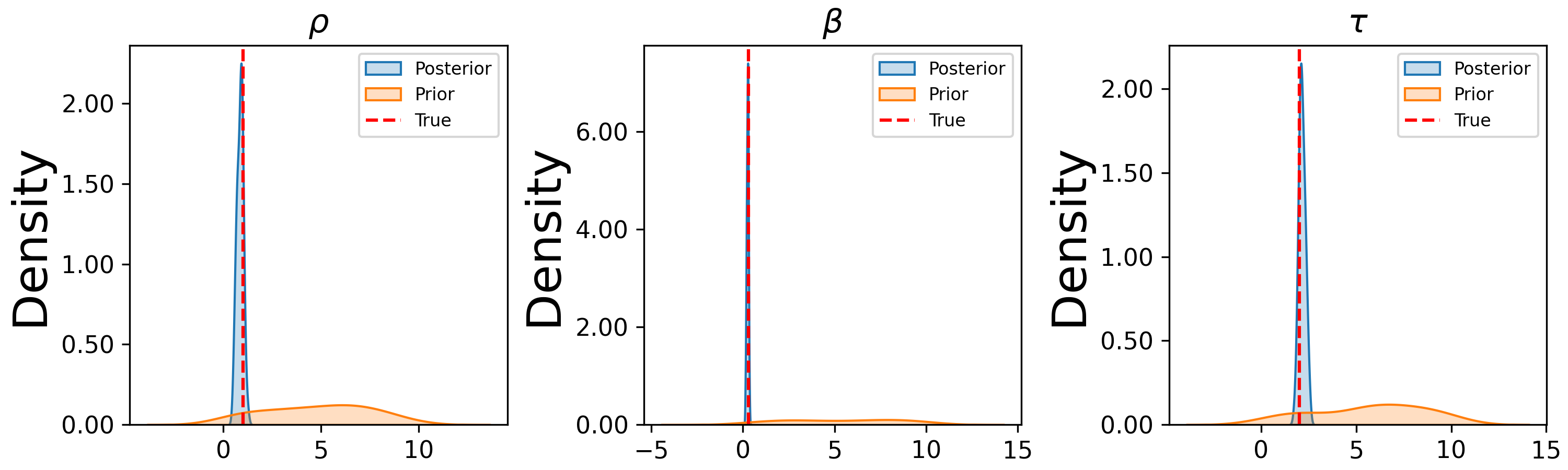}
    \caption{Posterior distribution for DGP: LinearParam DGP8, Simulator: LinearParam Sim8 (similar to LinearParam Sim6).}
    \label{fig:post-lp-dgp8}
\end{figure}

\paragraph{Evaluation} The inclusion of $X$ as a predictor of $T$ introduces additional variation that helps disambiguate the effects of $\rho$ and $\tau$, making it easier to learn about these parameters from the observed data. The results of this experiment are shown in Figure~\ref{fig:results-lp-dgp8} and ~\ref{fig:post-lp-dgp8}.

By modifying the treatment assignment mechanism to include an observed covariate, we ensure that the dataset is informative for the values of the {\knob}s, and learn a posterior over $\rho$ and $\tau$. As a result, we observe lower BSE for the posterior datasets. These results demonstrate that even small changes in the DGP that provide additional structure can lead to significantly more informative posteriors--highlighting the importance of leveraging partial identifiability when complete identification is not possible.

\subsection{Misspecified prior distributions}
In many real-world applications, domain knowledge about the underlying data-generating mechanisms may be limited or imprecise. As a result, prior distributions over simulator parameters can often be either too narrow or entirely misaligned with the true values. In this section, we explore such scenarios to better understand the consequences of using misspecified priors in causal estimation.

\paragraph{LinearParam Sim9: Narrow priors} We begin with a scenario in which the prior distributions are relatively narrow but still include the true parameter values. These priors represent a situation where the researcher believes they have strong—but not necessarily perfect—knowledge about the plausible range of parameter values. We reuse the data-generating process from earlier, LinearParam DGP6, with a correctly specified simulator:

\paragraph{LinearParam Sim9}
\begin{equation}
    \begin{aligned}
        Z \sim \text{Binomial}(0.5) \\
        X \sim \mathcal{N}(0,1) \\
        T = \mathbbm{1}(Z + U[0, 0.5)) \\
        Y = \rho Z + \beta X + \tau T + U[0, 0.5)
    \end{aligned}
\end{equation}

The true parameter values are $\rho = 1.0, \beta = 0.3, \tau =1.0$. We assume independent uniform priors over a narrow range: 

\paragraph{Prior}
\begin{equation}
    \begin{aligned}
        \text{Prior}(\rho) \sim U[0.0, 2.0] \\
        \text{Prior}(\beta) \sim U[0.0, 2.0] \\
        \text{Prior}(\tau) \sim U[0.0, 2.0] \\
    \end{aligned}
\end{equation}

\begin{figure}[h]
\centering
\begin{minipage}{0.49\textwidth}
    \centering
    \includegraphics[scale=0.4]{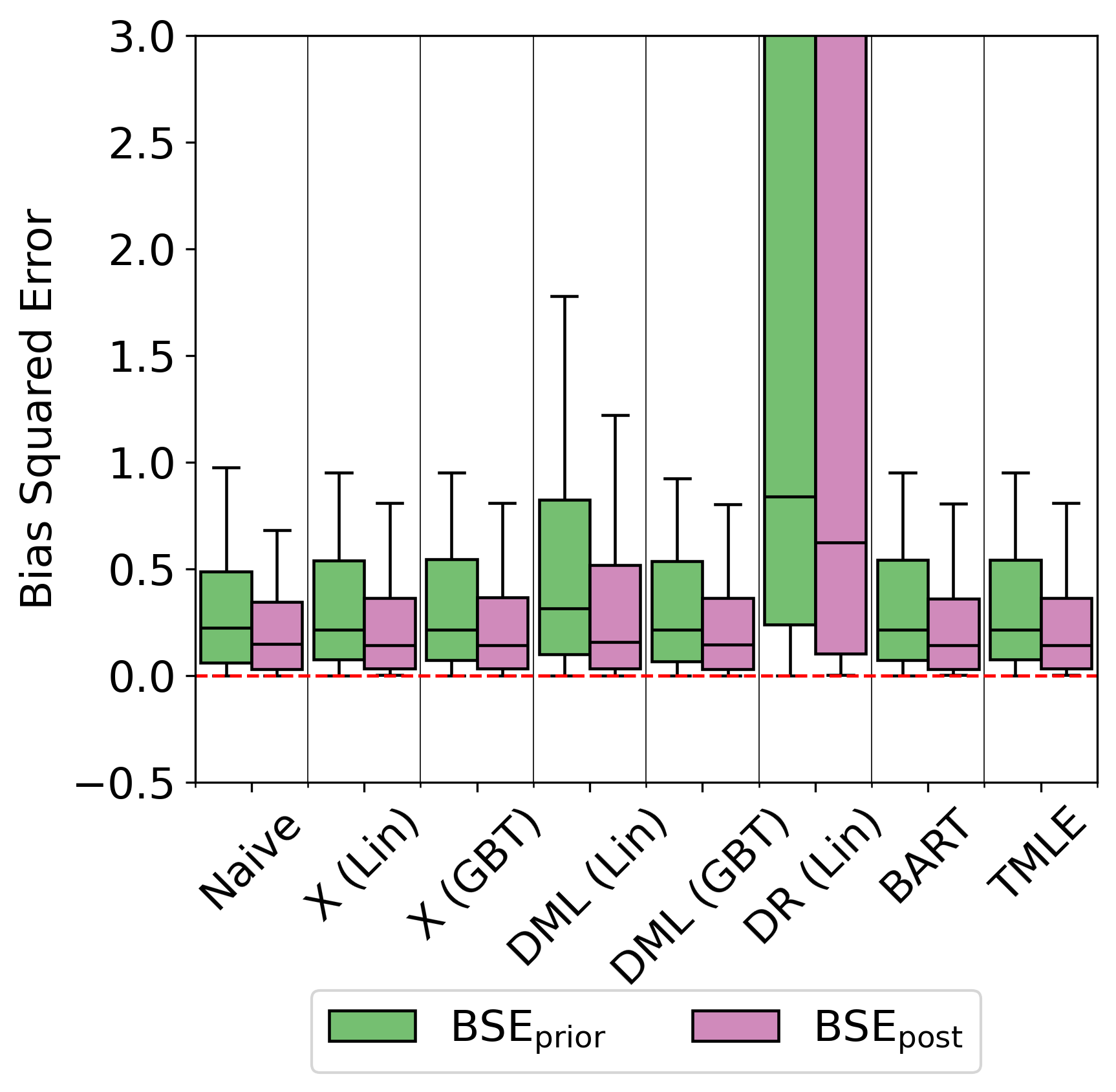}
\end{minipage}%
\begin{minipage}{0.47\textwidth}
    \centering
        \begin{tabular}{lcc}
        \toprule
        & Prior & Posterior \\ \midrule
        Diff. of Means & 0.315 & 0.229 \\
        X (Lin) & 0.311 & 0.229 \\
        X (GBT) & 0.311 & 0.230 \\
        DML (Lin) & 0.554 & 0.339 \\
        DML (GBT) & 0.306 & 0.231 \\
        DR (Lin)  & 14.914 & 0.033 \\
        BART & 0.311 & 0.229 \\
        TMLE & 0.311 & 0.229\\
        \bottomrule
        \end{tabular}
\end{minipage}
\caption{BSE for DGP: LinearParam DGP6 and Simulator: LinearParam Sim9.}
\label{fig:results-lp-dgp9}
\end{figure}

\begin{figure}[h]
    \centering
    \includegraphics[width=0.75\linewidth]{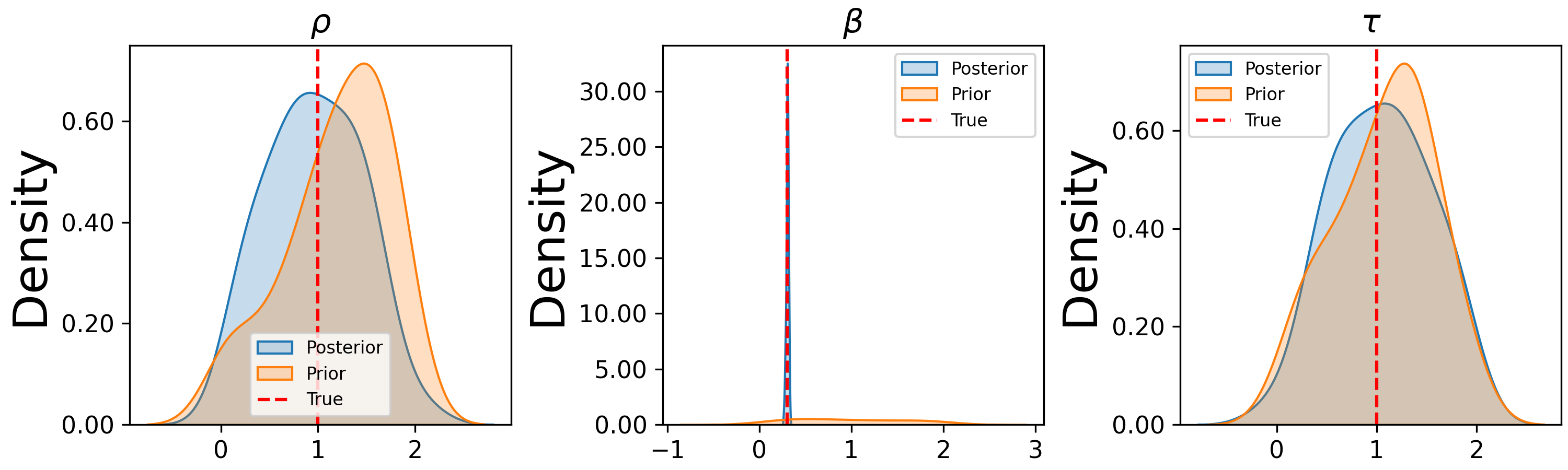}
    \caption{Posterior distribution for DGP: LinearParam DGP6, Simulator: LinearParam Sim9}
    \label{fig:post-lp-dgp9}
\end{figure}

\paragraph{Evaluation} Our hypothesis is that because the priors are informative and centered near the true values, the posterior will not offer a substantial improvement in the quality of datasets or causal estimates. The posterior and prior should lead to similar estimator performance. This hypothesis is confirmed by the results shown in Figure~\ref{fig:results-lp-dgp9} and ~\ref{fig:post-lp-dgp9}, where both the posterior and prior samples yield comparable ATE bias and parameter distributions. 

\paragraph{LinearParam Sim10: `Incorrect' priors} In practical scenarios, it is not uncommon to misspecify the range or directionality of plausible parameter values due to limited or misleading prior knowledge. This setting examines such cases where the prior distributions do not contain the true values of the {\knob}s that generated the observed data. When this occurs, we hypothesize that the posterior distribution does not converge toward the true parameter values, but instead toward those that produce simulations whose distributions are closest to the source data, as measured by our distance metric---here, the sliced-Wasserstein distance. To test this hypothesis, we run a controlled experiment with a linear, parametric simulator. The data-generating process (DGP) LinearParam DGP10 is defined as:

\paragraph{LinearParam DGP10}
\begin{equation}
    \begin{aligned}
        Z \sim \mathcal{N}(0, 1) \\
        X \sim \mathcal{N}(0, 1) \\
        T \sim \text{Binomial}(\rho Z + \beta X + \mathcal{N}(0, 0.1)) \\
        Y = \rho Z + \beta X + \tau T + \mathcal{N}(0, 0.1)
    \end{aligned}
\end{equation}

The true values of the {\knob}s are $\rho =1.0, \beta = -1.5, \tau= 1.5$. However, the priors are intentionally misspecified, such that none of the true values fall within their support:

\paragraph{Prior}
\begin{equation}
    \begin{aligned}
        \text{Prior}(\rho) \sim U[-2.0, 0.0] \\
        \text{Prior}(\beta) \sim U[0.0, 2.0] \\
        \text{Prior}(\tau) \sim U[-2.0, 0.0] \\
    \end{aligned}
\end{equation}

\begin{figure}[h]
\centering
\begin{minipage}{0.49\textwidth}
    \centering
    \includegraphics[scale=0.4]{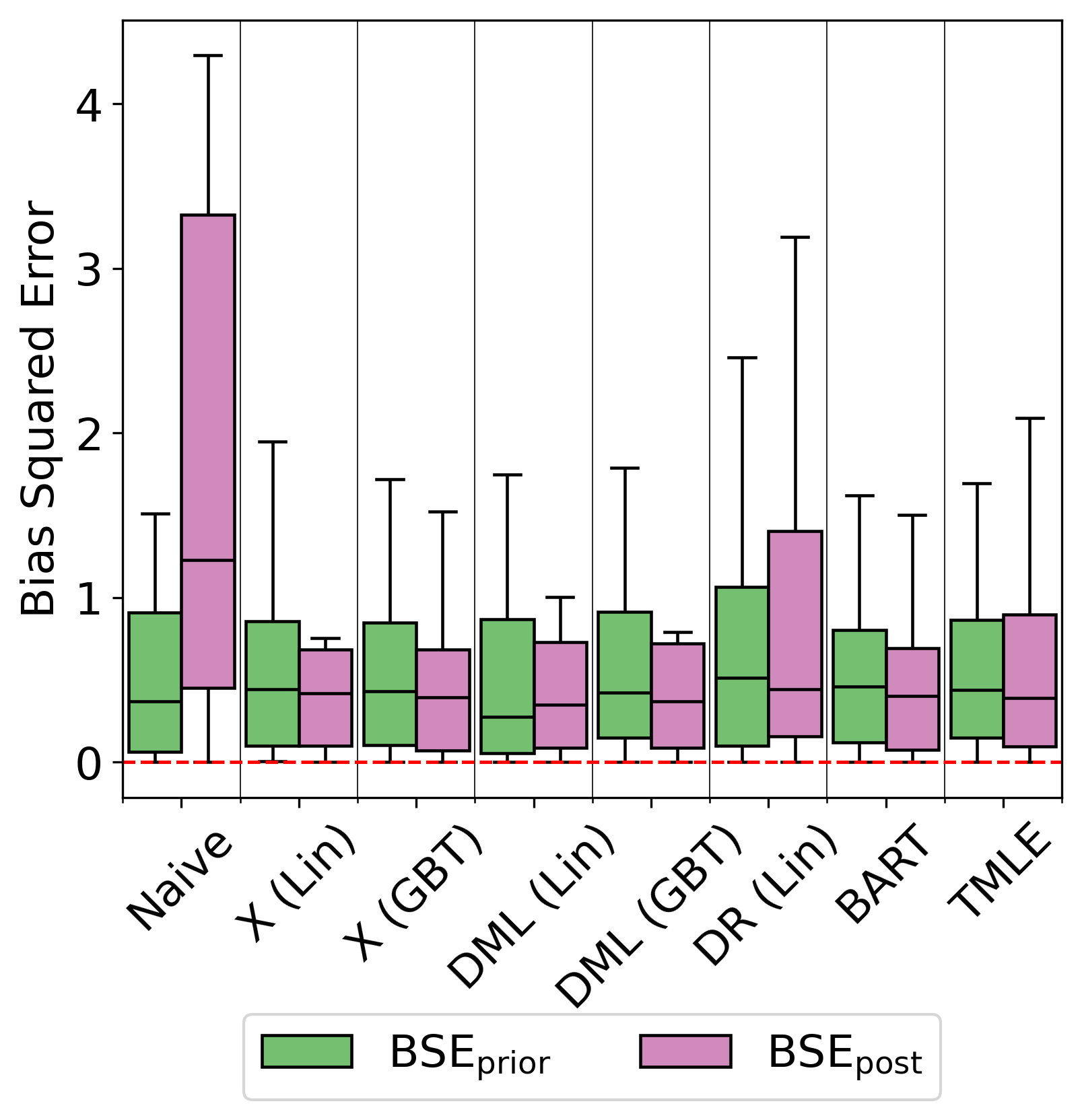}
\end{minipage}%
\begin{minipage}{0.47\textwidth}
    \centering
        \begin{tabular}{lcc}
        \toprule
        & \multicolumn{2}{c}{mean BSE} \\ \midrule
        & Prior & Posterior \\ \midrule
        Naive & 0.789 & 1.712 \\
        X Learner (Linear) & 0.627 & 0.756 \\
        X Learner (GBT) & 0.627 & 0.742 \\
        DML (Linear) & 0.649 & 0.866 \\
        DML (GBT) & 0.633 & 0.752 \\
        DR (Linear)  & 1e7 & 1e7 \\
        Causal BART & 0.612 & 0.742 \\
        TMLE & 0.666 & 0.784\\
        \bottomrule
        \end{tabular}
\end{minipage}
\caption{BSE for DGP: LinearParam DGP10 and Simulator: LinearParam Sim10.}
\label{fig:results-lp-dgp10}
\end{figure}

\begin{figure}[h]
    \centering
    \includegraphics[width=0.75\linewidth]{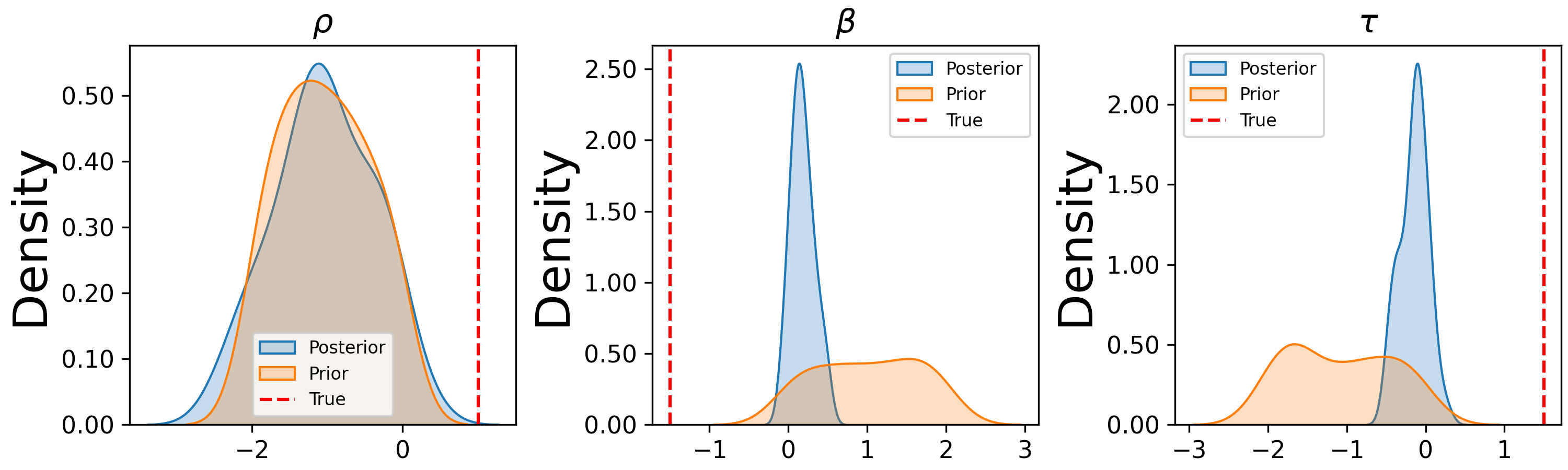}
    \caption{Posterior distribution for DGP: LinearParam DGP10, Simulator: LinearParam Sim10}
    \label{fig:post-lp-dgp10}
\end{figure}

\paragraph{Evaluation} The resulting mean BSE and posterior distributions are presented in Figures~\ref{fig:results-lp-dgp10} and ~\ref{fig:post-lp-dgp10}. We not that the posterior and prior exhibit similar performance across estimators. 

The specification of priors—and their potential misspecification—is a long-standing consideration in the Bayesian literature. While identifying the correct prior is often impractical, we recommend that practitioners using this framework iterate over a range of plausible prior distributions and manually inspect the datasets generated by the simulator. This process can surface misalignments between prior assumptions and observable data. As a practical guideline, we suggest beginning with heavy-tailed and broad priors to allow the posterior more flexibility during the initial exploration

\begin{tcolorbox}[colback=gray!5!white, colframe=gray!75!black, title=\textbf{Key takeaway}]
Using parametric, linear simulators we explored how different sources of uncertainty and modeling choices impact the effectiveness of {\ourmethod}. 
\begin{itemize}
    \item \textbf{Flexible simulators:} Even when the true functional form deviates from the simulator’s assumptions (e.g., added interaction or non-linear terms), the method remains robust as long as the simulator is expressive enough to capture distributional similarities. In these cases, the causal estimates on posterior datasets mirrored that of the source data.
    \item \textbf{Well-calibrated, wide priors:} In settings with limited domain knowledge, wide priors with heavy tails allow the posterior to concentrate near plausible values. Even when the priors are misspecified, if the simulator is accurate and the posterior matching is distributionally guided (e.g., via sliced-Wasserstein distance), the resulting datasets often lead to informative posteriors.
    \item \textbf{Low model misspecification:} In parametric settings where the simulator closely matches the true DGP, the method reliably sharpens inference. Posterior datasets exhibit lower bias squared error.
    \item \textbf{Joint posterior estimation for identifiable subspaces:} When parameters are non-identifiable individually but have identifiable combinations (e.g., fixed sum constraints), the method can still recover meaningful structure when the remaining parameters (like $\beta$) are well-informed by the data.
\end{itemize}
\end{tcolorbox}

\section{Evaluating {\ourmethod} on synthetic datasets generated using non-parametric data-generating processes}
\label{app:sbice-frugal-dgps}
We evaluate {\ourmethod} on synthetic datasets with high-dimensional covariates, generated using parametric models and frugal parameterization~\citep{evans2024parameterizing}. For each experiment, we report the mean BSE for a set  of causal estimators to assess how closely the generated datasets resemble the source distribution and how the inferred posterior affects estimator performance. For these datasets, we vary observable characteristics, including sample size, number of covariates and functional relationships between variables. 

We use two types of simulators for these experiments: (1) \textbf{FrugalParam}, which implements the data-generating process directly via frugal parameterization and serves as a correctly specified simulator; and (2) \textbf{FrugalFlows}~\citep{de2024marginal}, a non-parametric simulator based on normalizing flows and copulas. Note that the FrugalFlows simulators may not be able to accurately learn the source data distribution as they require large sample sizes for convergence. We use the labels Frugal DGP($x$) to denote these types of datasets.

\subsection{Frugal DGP1}

\paragraph{Frugal DGP1} This dataset includes two unobserved confounders $Z_1, Z_2$, and three observed confounders $X_1, X_2, X_3$, a binary treatment $T$ and outcome $Y$. We generate $3000$ samples from the following causal model. 

\begin{equation}
\begin{aligned}
    Z_1 \sim \text{Beta}(1.0, 1.0) \\
    Z_2 \sim \mathcal{N}(1.0, 0.5) \\
    X_1 \sim \mathcal{N}(-2.0, 2.0) \\
    X_2 \sim \text{Beta}(0.0, 0.25) \\
    X_3 \sim \text{t}(1.0, 1.0) \\
    T \sim \text{Binomial}(0.5 + 0.4 X_1 + 0.3 Z_1 + X_1 Z_1 + X_2 + 1.5 X_3 + 2.5 X_3 - 0.5 X_1 X_3 + Z_2) \\
    Y \mid \text{do(T)} \sim \mathcal{N}(3.0 T, 1.0)
\end{aligned}  
\label{dgp:frugal-dgp1}
\end{equation}

We use a multivariate Gaussian copula to introduce dependencies between the covariates and the causal effect distribution. The Spearman correlation matrix for the copula is 
\begin{equation}
\boldsymbol{R}_1 = 
\begin{pmatrix}
1.0 & 0.8 & 0.8 & 0.8 & 0.8 & 0.8 \\
0.8 & 1.0 & 0.8 & 0.8 & 0.8 & 0.8 \\
0.8 & 0.8 & 1.0 & 0.8 & 0.8 & 0.8 \\
0.8 & 0.8 & 0.8 & 1.0 & 0.8 & 0.8 \\
0.8 & 0.8 & 0.8 & 0.8 & 1.0 & 0.8 \\
0.8 & 0.8 & 0.8 & 0.8 & 0.8 & 1.0
\end{pmatrix}
\label{eq:dgp1-rho}
\end{equation}

\paragraph{FrugalFlows Sim1} We train a FrugalFlows simulator for this dataset, using the hyperparameters shown in Table~\ref{tab:frugal-sim1-hyp}. 

\begin{table}[htb]
    \caption{FrugalFlows hyperparameters for Frugal Sim1. \\}
    \label{tab:frugal-sim1-hyp}
    \centering
    \begin{tabular}{lc}
    \toprule
    Hyperparameter & Value \\ \midrule
      RQS knots & 2 \\
      Flow layers & 1 \\
      Learning rate & 0.0054 \\
      Network Depth & 3 \\
      Network Width & 9 \\
      Causal model (CM) & Location translation \\ 
      (CM) RQS knots & 7 \\
      (CM) Flow layers & 8 \\
      (CM) Network Depth & 6 \\
      (CM) Network Width & 6 \\
    \bottomrule
    \end{tabular}
\end{table}

\paragraph{Prior} We specify a single value for the causal effect parameter $\tau$ and $\rho$ while generating the datasets from the simulator. The prior is given by 

\begin{equation}
    \begin{aligned}
        \text{Prior}(\tau) \sim U[0.0, 10.0] \\
        \text{Prior}(\rho) \sim U[-1.0, 1.0] \\
    \end{aligned}
\label{eq:frugal-dgp1-prior}
\end{equation} 

\begin{figure}[ht]
\centering
\begin{minipage}{0.49\textwidth}
    \centering
    \includegraphics[scale=0.4]{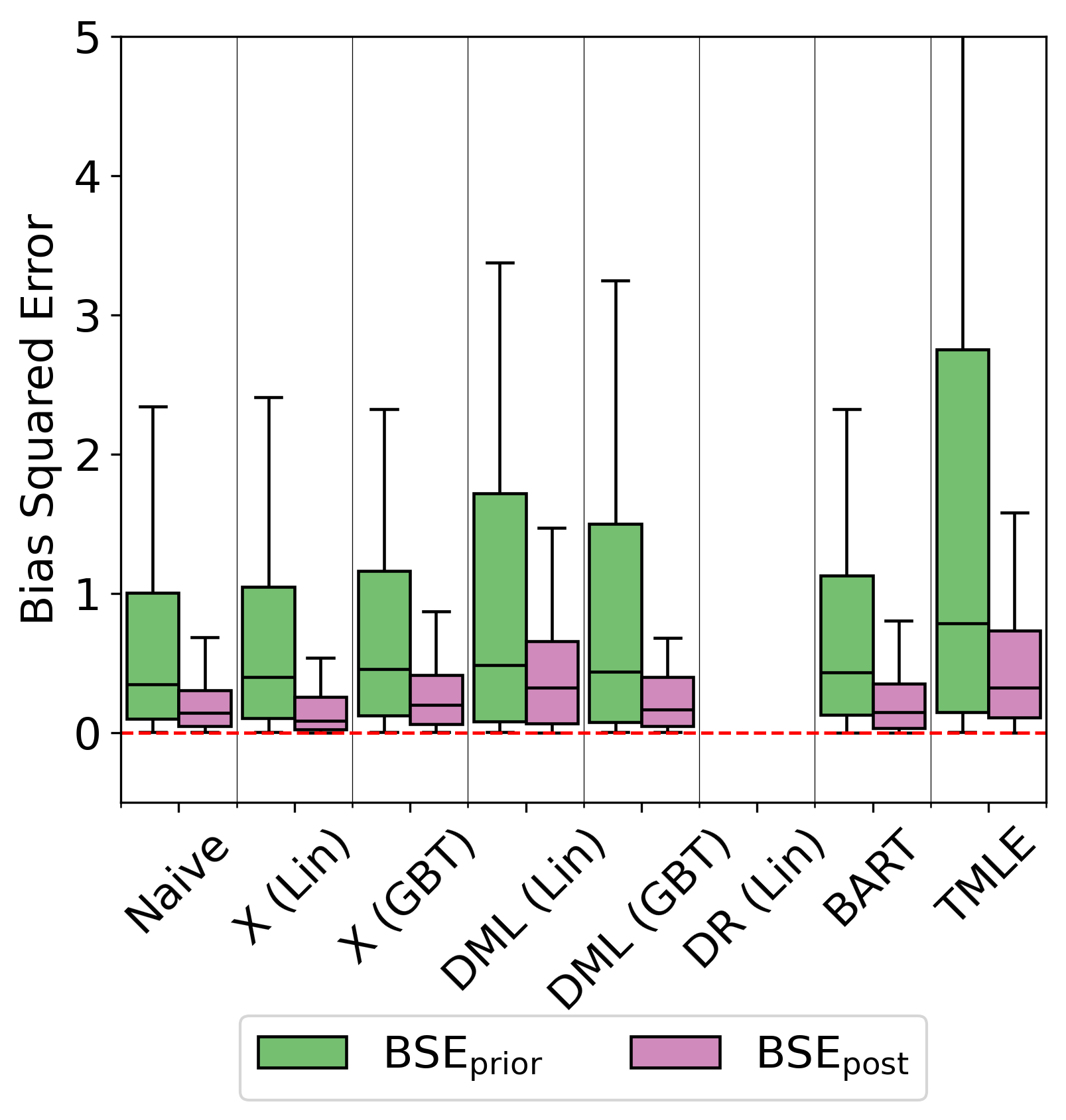}
\end{minipage}%
\begin{minipage}{0.47\textwidth}
    \centering
        \begin{tabular}{lcc}
        \toprule
        & Prior & Posterior \\ \midrule
        Diff. of Means & $0.748$ & $0.227$ \\
        X (Lin) & $0.857$ & $0.196$ \\
        X (GBT) & $1.027$ & $0.307$ \\
        DML (Lin) & $1.297$ & $0.461$ \\
        DML (GBT) & $1.007$ & $0.308$ \\
        DR (Lin) & $3e7$ & $3e7$ \\
        BART & $0.982$ & $0.262$ \\
        TMLE & $1.779$ & $0.620$\\
        \bottomrule
        \end{tabular}
\end{minipage}
\caption{BSE for DGP: FrugalParam DGP1 and Simulator: FrugalFlows Sim1. (Note the large bias squared error for the estimates of the DR (Lin) estimator). }
\label{fig:results-fp-dgp1}
\end{figure}

\paragraph{Evaluation} We include the BSE for the estimators in Figure~\ref{fig:results-fp-dgp1}. 

\subsection{Frugal DGP2} 

\paragraph{Frugal DGP2} We use a similar dataset as Frugal DGP1, but vary the correlation matrix and the dependence between the variables as follows

\begin{equation}
\boldsymbol{R}_2 = 
\begin{pmatrix}
1.0 & 0.8 & 0.2 & 0.3 & 0.2 & 0.7 \\
0.8 & 1.0 & 0.1 & 0.4 & 0.9 & 0.3 \\
0.2 & 0.1 & 1.0 & 0.5 & 0.8 & 0.1 \\
0.3 & 0.4 & 0.5 & 1.0 & 0.9 & 0.5 \\
0.2 & 0.9 & 0.8 & 0.9 & 1.0 & 0.6 \\
0.7 & 0.3 & 0.1 & 0.5 & 0.6 & 1.0
\end{pmatrix}
\label{eq:dgp2-rho}
\end{equation}

\paragraph{FrugalFlows Sim2} We simulate data using FrugalFlows with the hyperparameters shown in Table~\ref{tab:frugal-sim2-hyp}. 

\begin{table}[htb]
    \caption{FrugalFlows hyperparameters for Frugal Sim2 \\}
    \label{tab:frugal-sim2-hyp}
    \centering
    \begin{tabular}{lc}
    \toprule
    Hyperparameter & Value \\ \midrule
      RQS knots & 6 \\
      Flow layers & 6 \\
      Learning rate & 0.0027 \\
      Network Depth & 7 \\
      Network Width & 9 \\
      Causal model (CM) & Location translation \\ 
      (CM) RQS knots & 6 \\
      (CM) Flow layers & 5 \\
      (CM) Network Depth & 2 \\
      (CM) Network Width & 9 \\
    \bottomrule
    \end{tabular}
\end{table}

We introduce some model misspecification in this experiment, by using a single value for the {\knob} $\rho$ to approximate the dependence between all the covariates and the causal effect distribution. 

\paragraph{Prior}
\begin{equation}
    \begin{aligned}
        \text{Prior}(\tau) \sim U[0.0, 10.0] \\
        \text{Prior}(\rho) \sim U[-1.0, 1.0] \\
    \end{aligned}
\label{eq:frugal-dgp2-prior}
\end{equation}

\begin{figure}[ht]
\centering
\begin{minipage}{0.49\textwidth}
    \centering
    \includegraphics[scale=0.4]{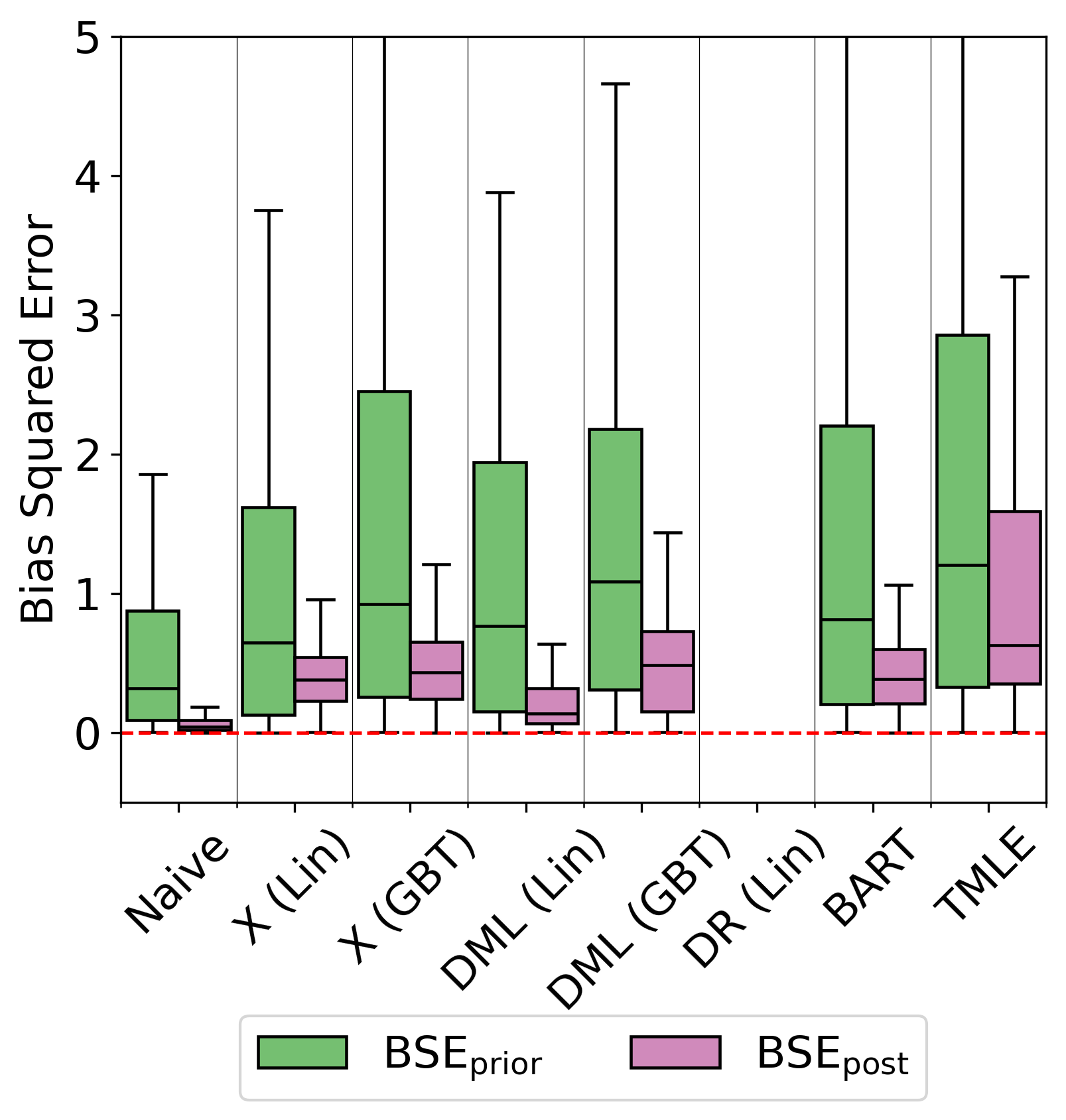}
\end{minipage}%
\begin{minipage}{0.47\textwidth}
    \centering
        \begin{tabular}{lcc}
        \toprule
        & Prior & Posterior \\ \midrule
        Diff. of Means & $0.516$ & $0.067$ \\
        X (Lin) & $1.071$ & $0.398$ \\
        X (GBT) & $1.529$ & $0.483$ \\
        DML (Lin) & $1.242$ & $0.240$ \\
        DML (GBT) & $1.504$ & $0.508$ \\
        DR (Lin) & $1e8$ & $9e7$ \\
        BART & $1.393$ & $0.411$ \\
        TMLE & $1.803$ & $1.236$\\
        \bottomrule
        \end{tabular}
\end{minipage}
\caption{BSE for DGP: Frugal DGP2 and Simulator: FrugalFlows Sim2.}
\label{fig:results-fp-dgp2}
\end{figure}

\paragraph{Evaluation} We include the mean BSE for the estimators in Figure~\ref{fig:results-fp-dgp2}. Note that $\text{BSE}_\text{post}$ is lower than the corresponding prior values, indicating that the posterior estimates are informative. 

\subsection{Frugal DGP3} 

\paragraph{Frugal DGP3} In this dataset, we have three covariates $X_1, X_2, X_3$, a binary treatment $T$ and outcome $Y$ for $5000$ samples defined using the following model 

\begin{equation}
\begin{aligned}
    X_1 \sim \mathcal{N}(2.0, 1.0) \\
    X_2 \sim \text{Gamma}(1.0, 1.0) \\
    X_3 \sim \mathcal{N}(3.0, 1.0) \\
    T \sim \text{Binomial}(-0.3 X_1 + 0.3 X_2 -0.4 X_3 - 0.1 X_1 X_2) \\
    Y \mid \text{do(T)} \sim \mathcal{N}(5.0 T, 1.5)
\end{aligned}  
\label{dgp:frugal-dgp3}
\end{equation}

The dependence between the covariates and the causal effect distribution is modeled using a multivariate Gaussian copula with the following Spearman correlation matrix 
\begin{equation}
\boldsymbol{R}_3 = 
\begin{pmatrix}
1.0 & 0.0 & 0.0 & -0.5 \\
0.0 & 1.0 & 0.0 & -0.3 \\ 
0.0 & 0.0 & 1.0 & 0.9 \\
-0.5 & -0.3 & 0.9 & 1.0 \\
\end{pmatrix}
\label{eq:dgp3-rho}
\end{equation}

\paragraph{FrugalFlows Sim3} We trained a FrugalFlows model on the source data using the hyperparameters stated in Table~\ref{tab:frugal-sim3-hyp}. We ran two version of the simulator, one without unobserved confounding, by setting the parameter $\rho = 0.0$, and another with unobserved confounding by dropping the covariate $X_2$ from the source data. We call the second setup FrugalFlows Sim3(u) and show the priors for both settings below. The hyperparameters for the dataset with unbserved confounding are also included in Table~\ref{tab:frugal-sim3-hyp}. 

\begin{table}[htb]
    \caption{FrugalFlows hyperparameters for FrugalFlows Sim3 and FrugalFlows Sim 3(u) \\}
    \label{tab:frugal-sim3-hyp}
    \centering
    \begin{tabular}{lcc}
    \toprule
    & FrugalFlows Sim3 & FrugalFlows Sim3(u) \\ \midrule
    Hyperparameter & \multicolumn{2}{c}{Value} \\ \midrule
      RQS knots & 2 & 8\\
      Flow layers & 1 & 11 \\
      Learning rate & 0.0061 & 0.0215\\
      Network Depth & 4 & 3\\
      Network Width & 39 & 89\\
      Causal model (CM) & Location translation & Location translation \\ 
      (CM) RQS knots & 15 & 6\\
      (CM) Flow layers & 15 & 4\\
      (CM) Network Depth & 53 & 62  \\
      (CM) Network Width & 75 & 62\\
    \bottomrule
    \end{tabular}
\end{table}

\paragraph{Prior: Sim3 }
\begin{equation}
    \begin{aligned}
        \text{Prior}(\tau) \sim U[0.0, 10.0] 
    \end{aligned}
\label{eq:frugal-dgp3-prior}
\end{equation} 

\paragraph{Prior: Sim3(u)}
\begin{equation}
    \begin{aligned}
        \text{Prior}(\tau) \sim U[0.0, 10.0] \\ 
        \text{Prior}(\rho) \sim U[-1.0, 1.0]
    \end{aligned}
\label{eq:frugal-dgp3u-prior}
\end{equation} 

\begin{figure}[htb]
\centering
\begin{minipage}{0.47\textwidth}
    \centering
    \includegraphics[scale=0.4]{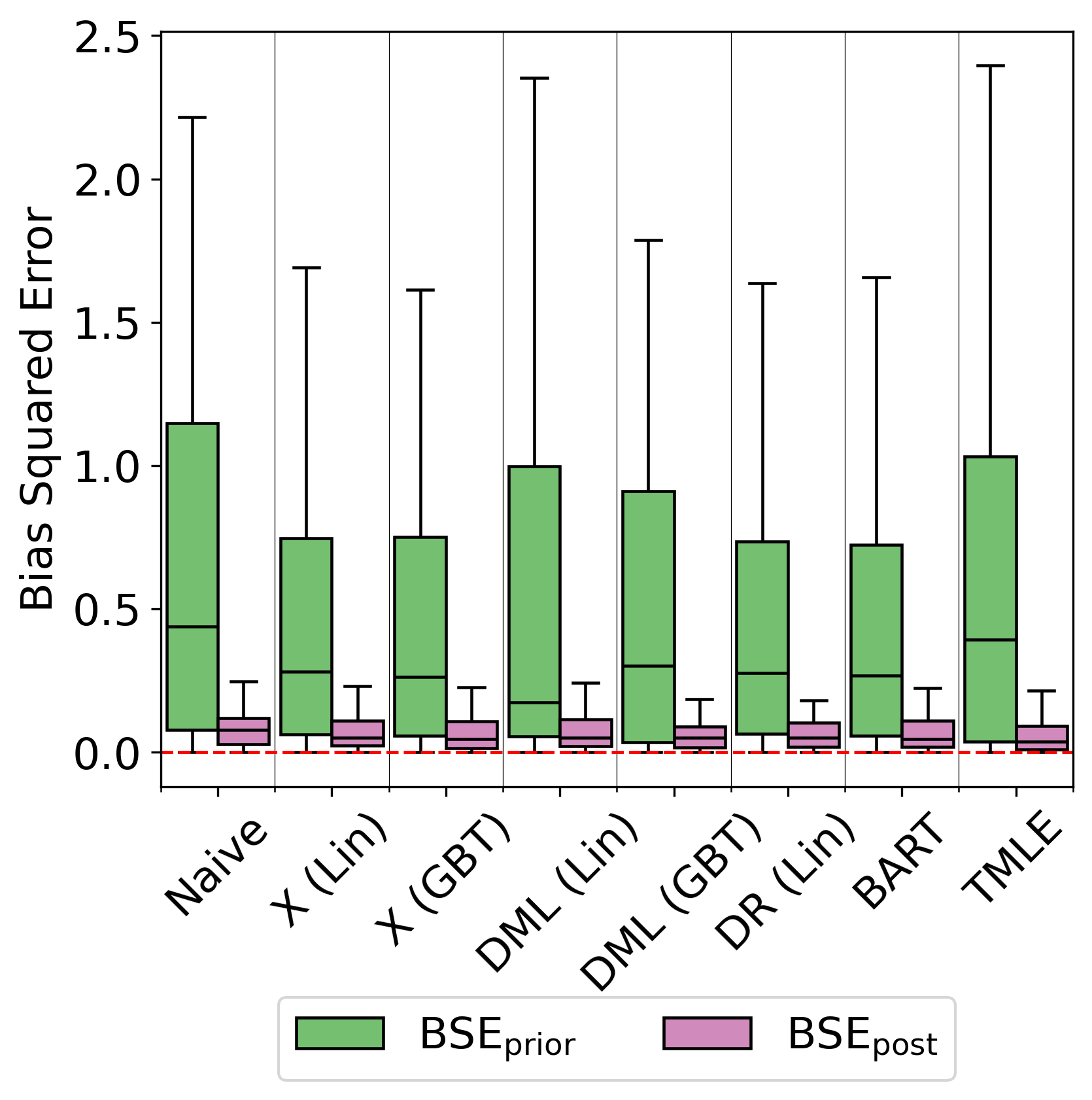}
\end{minipage}%
\begin{minipage}{0.47\textwidth}
    \centering
        \begin{tabular}{lcc}
        \toprule
        & Prior & Posterior \\ \midrule
        Diff. of Means & $0.868$ & $0.097$ \\
        X (Lin) & $0.633$ & $0.069$ \\
        X (GBT) & $0.610$ & $0.064$ \\
        DML (Lin) & $0.601$ & $0.077$ \\
        DML (GBT) & $0.695$ & $0.067$ \\
        DR (Lin) & $0.595$ & $0.067$ \\
        BART & $0.616$ & $0.064$ \\
        TMLE & $0.773$ & $0.062$\\
        \bottomrule
        \end{tabular}
\end{minipage}
\caption{BSE for DGP: Frugal DGP3 and Simulator: FrugalFlows Sim3.}
\label{fig:results-fp-dgp3}
\end{figure}

\begin{figure}[!h]
\centering
\begin{minipage}{0.47\textwidth}
    \centering
    \includegraphics[scale=0.4]{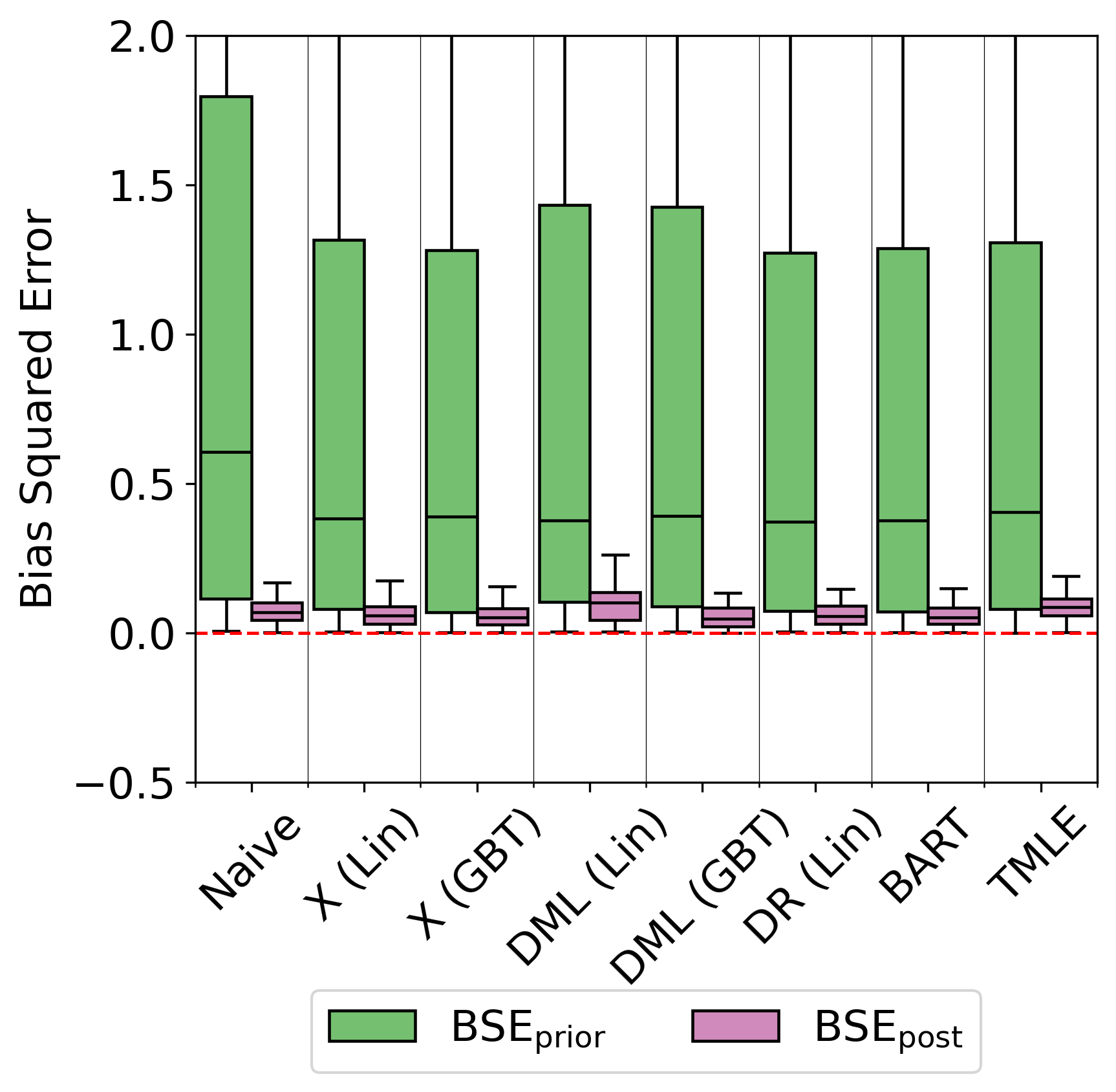}
\end{minipage}%
\begin{minipage}{0.47\textwidth}
    \centering
        \begin{tabular}{lcc}
        \toprule
        & Prior & Posterior \\ \midrule
        Diff. of Means & $1.175$ & $0.087$ \\
        X (Lin) & $0.968$ & $0.071$ \\
        X (GBT) & $0.923$ & $0.065$ \\
        DML (Lin) & $1.031$ & $0.108$ \\
        DML (GBT) & $0.972$ & $0.063$ \\
        DR (Lin) & $1.028$ & $0.072$ \\
        BART & $0.927$ & $0.066$ \\
        TMLE & $0.910$ & $0.098$\\
        \bottomrule
        \end{tabular}
\end{minipage}
\caption{BSE for DGP: Frugal DGP3 and Simulator: FrugalFlows Sim3(u).}
\label{fig:results-fp-dgp3u}
\end{figure}

\paragraph{Evaluation} We compute the mean BSE for causal estimators for FrugalFlows Sim3 in Figure~\ref{fig:results-fp-dgp3} and FrugalFlows Sim3(u) in Figure~\ref{fig:results-fp-dgp3u}. In both cases, the posterior estimates are informative and similar to the source data. 


\subsection{Frugal DGP4}

\paragraph{FrugalDGP4} This dataset is based on the simulator in the FrugalFlows~\citep{de2024marginal} paper (referred to as $M_1$ in the original paper). We have four covariates $X_1..X_4$, a binary treatment $T$, and outcome $Y$. We describe the full data generating process as shown below. 

\begin{equation}
\begin{aligned}
    X_1 \sim \text{Gamma}(\mu = 1, \phi=1) \\
    X_2 \sim \text{Gamma}(\mu = 1, \phi=1) \\
    X_3 \sim \text{Gamma}(\mu = 1, \phi=1) \\
    X_4 \sim \text{Gamma}(\mu = 1, \phi=1) \\
    T \sim \text{Binomial}(-2 + X_1 + X_2 + X_3 + X_4) \\
    Y \mid \text{do(T)} \sim \mathcal{N}(0.5 + 5T, 1)
\end{aligned}  
\label{dgp:frugal-dgp4}
\end{equation}

The gaussian dependency matrix $R_4$ is given by
\begin{equation}
\boldsymbol{R}_4 = 
\begin{pmatrix}
1.0 & 0.5 & 0.3 & 0.1 & 0.8 \\
0.5 & 1.0 & 0.4 & 0.1 & 0.8 \\
0.3 & 0.4 & 1.0 & 0.1 & 0.8 \\
0.1 & 0.1 & 0.1 & 1.0 & 0.8 \\
0.8 & 0.8 & 0.8 & 0.8 & 1.0 
\end{pmatrix}
\label{eq:dgp4-rho}
\end{equation}

\paragraph{FrugalParam Sim4(u)} For this dataset, we use a simulator that is based on the frugal parameterization method, called FrugalParam Sim4(u). This experiment is a sanity check to verify the informativeness of the selected {\knob}s. We drop the covariate $X_4$ from the dataset to simulate unobserved confounding, and treat the last column of matrix $R_4$ as the unobserved confounding bias. The simulator is as described below

\begin{equation}
    \begin{aligned}
    X_1 \sim \text{Gamma}(\mu = 1, \phi=1) \\
    X_2 \sim \text{Gamma}(\mu = 1, \phi=1) \\
    X_3 \sim \text{Gamma}(\mu = 1, \phi=1) \\
    X_4 \sim \text{Gamma}(\mu = 1, \phi=1) \\
    T \sim \text{Binomial}(-2 + X_1 + X_2 + X_3 + X_4) \\
    Y \mid \text{do(T)} \sim \mathcal{N}(0.5 + \tau T, 1)
\end{aligned}  
\label{dgp:frugalparam-sim4}
\end{equation}

\paragraph{Prior: FrugalParam Sim4(u)} The prior is given by
\begin{equation}
    \begin{aligned}
        \text{Prior}(\tau) \sim U[-20.0, 20.0] \\ 
        \text{Prior}(\rho) \sim U[-1.0, 1.0]
    \end{aligned}
\end{equation}

\begin{figure}[ht]
\centering
\begin{minipage}{0.49\textwidth}
    \centering
    \includegraphics[scale=0.4]{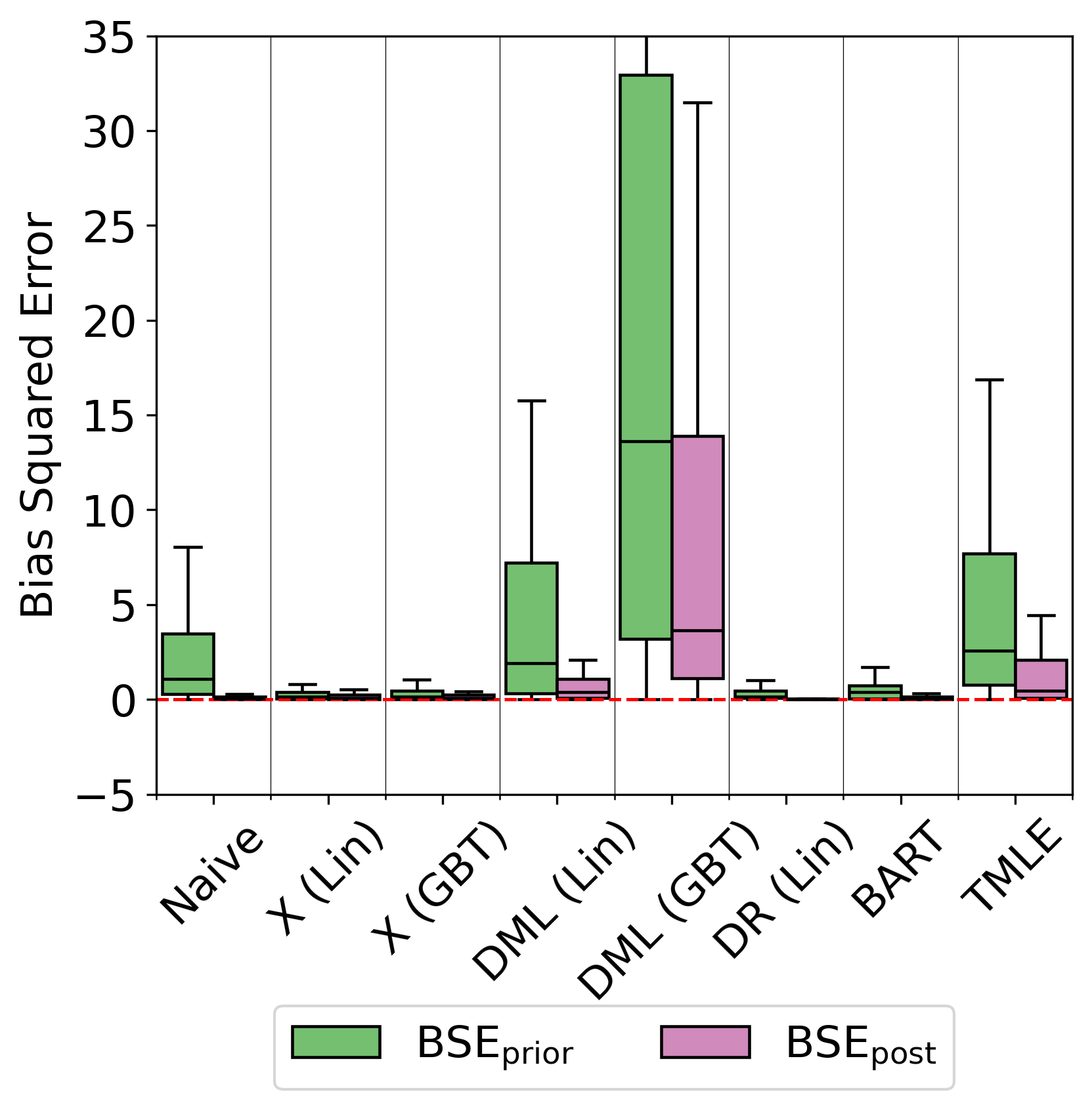}
\end{minipage}%
\begin{minipage}{0.47\textwidth}
    \centering
        \begin{tabular}{lcc}
        \toprule
        & Prior & Posterior \\ \midrule
        Diff. of Means & $1.992$ & $0.127$  \\
        X (Lin) & $0.277$ & $0.196$  \\
        X (GBT) & $0.304$ & $0.219$\\
        DML (Lin) & $5.144$ & $0.711$ \\
        DML (GBT) & $34.767$ & $10.531 $ \\
        DR (Lin) & $0.374 $& $0.040 $ \\
        BART & $0.585$ & $0.103$ \\
        TMLE & $4.809$ & $2.554$\\
        \bottomrule
        \end{tabular}
\end{minipage}
\caption{BSE for DGP: Frugal DGP4 and Simulator: FrugalParam Sim4(u).}
\label{fig:results-fp-causl-dgp4}
\end{figure}

\paragraph{Evaluation} The BSE is displayed in Figure~\ref{fig:results-fp-causl-dgp4}. 

\paragraph{FrugalFlows Sim4 and FrugalFlows Sim4(u)} For this experiment, we also train FrugalFlows as the simulator using two versions: with and without unobserved confounding. These simulators are described as FrugalFlows Sim4 and FrugalFlows Sim4(u) respectively. As stated earlier, to simulate unobserved confounding, we drop the covariate $X_4$ from the data. The hyperparameters for both simulators are shown in Table~\ref{tab:frugal-sim4-hyp}.

\begin{table}[htb]
    \caption{FrugalFlows hyperparameters for FrugalFlows Sim4 and FrugalFlows Sim4(u). \\}
    \label{tab:frugal-sim4-hyp}
    \centering
    \begin{tabular}{lcc}
    \toprule
    &  FrugalFlows Sim4 & FrugalFlows Sim4(u) \\ \midrule
    Hyperparameter & \multicolumn{2}{c}{Value} \\ \midrule
      RQS knots & 3 & 3\\
      Flow layers & 1 & 1\\
      Learning rate & 0.0077 & 0.0597\\
      Network Depth & 1 & 1\\
      Network Width & 98 & 58\\
      Causal model (CM) & Location translation & Location translation \\ 
      (CM) RQS knots & 5 & 4\\
      (CM) Flow layers & 3 & 2\\
      (CM) Network Depth & 78 & 68\\
      (CM) Network Width & 84 & 69\\
    \bottomrule
    \end{tabular}
\end{table}

\paragraph{Prior: FrugalFlows Sim4} The prior is given by
\begin{equation}
    \begin{aligned}
        \text{Prior}(\tau) \sim U[-20.0, 20.0] \\ 
    \end{aligned}
\end{equation}

\paragraph{Prior: FrugalParam Sim4(u)} The prior is given by
\begin{equation}
    \begin{aligned}
        \text{Prior}(\tau) \sim U[-20.0, 20.0] \\ 
        \text{Prior}(\rho) \sim U[-1.0, 1.0]
    \end{aligned}
\end{equation}

\begin{figure}[ht]
\centering
\begin{minipage}{0.49\textwidth}
    \centering
    \includegraphics[scale=0.4]{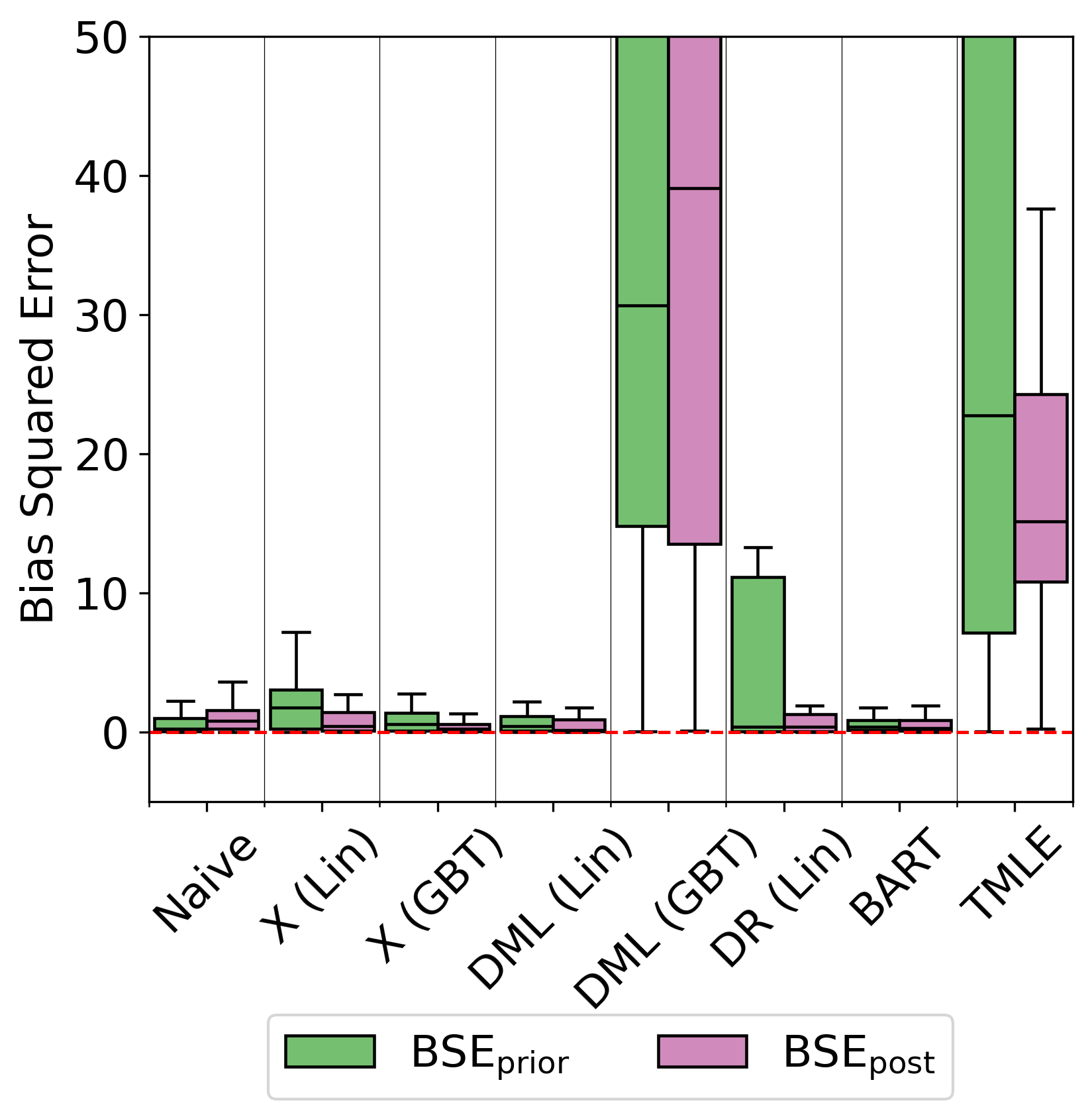}
\end{minipage}%
\begin{minipage}{0.47\textwidth}
    \centering
        \begin{tabular}{lcc}
        \toprule
        & Prior & Posterior \\ \midrule
        Diff. of Means & $0.605$ & $1.164$ \\
        X (Lin) & $2.182$ & $1.039$ \\
        X (GBT) & $1.009$ & $0.425$ \\
        DML (Lin) & $0.849$ & $0.669$ \\
        DML (GBT) & $38.579$ & $44.946$ \\
        DR (Lin) & $8e6$ & $1e6$ \\
        BART & $0.636$ & $0.506$ \\
        TMLE & $44.175$ & $17.503$\\
        \bottomrule
        \end{tabular}
\end{minipage}
\caption{BSE for DGP: Frugal DGP4 and Simulator: FrugalFlows Sim4.}
\label{fig:results-fp-dgp4}
\end{figure}

\begin{figure}[ht]
\centering
\begin{minipage}{0.49\textwidth}
    \centering
    \includegraphics[scale=0.4]{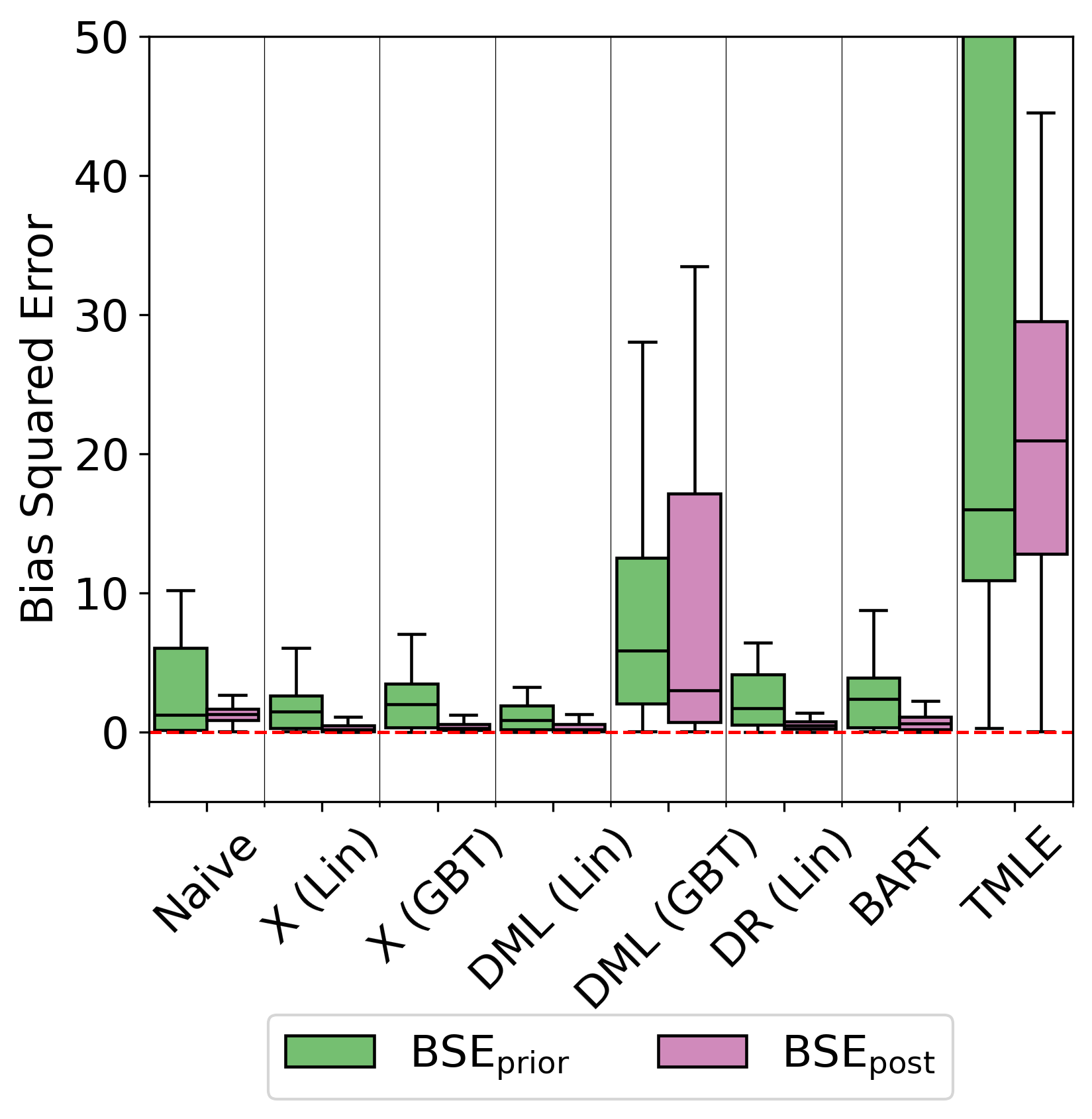}
\end{minipage}%
\begin{minipage}{0.47\textwidth}
    \centering
        \begin{tabular}{lcc}
        \toprule
        & Prior & Posterior \\ \midrule
        Diff. of Means & $2.961$ & $1.305$ \\
        X (Lin) & $1.971$ & $0.315$ \\
        X (GBT) & $2.187$ & $0.415$ \\
        DML (Lin) & $1.226$ & $0.439$ \\
        DML (GBT) & $16.569$ & $12.744$ \\
        DR (Lin) & $1e7$ & $0.628$ \\
        BART & $2.447$ & $0.709$ \\
        TMLE & $48.569$ & $21.788$\\
        \bottomrule
        \end{tabular}
\end{minipage}
\caption{BSE for DGP: Frugal DGP4 and Simulator: FrugalFlows Sim4(u).}
\label{fig:results-fp-dgp4u}
\end{figure}

\paragraph{Evaluation} We display the BSE in Figures~\ref{fig:results-fp-dgp4} and~\ref{fig:results-fp-dgp4u} respectively.

\subsection{Frugal DGP5}

\paragraph{Frugal DGP5} We replicate model $M_3$ in the FrugalFlows~\citep{de2024marginal} paper for this experiment. It has 10 covariates, which are combination of discrete and continuous covariates $X_1..X_{10}$, a binary treatment $T$ and outcome $Y$. The data generating process is as shown below

\begin{equation}
\begin{aligned}
    X_1 \sim \text{Gamma}(\mu = 1.3, \phi=1) \\
    X_2 \sim \text{Gamma}(\mu = 1.3, \phi=1) \\
    X_3 \sim \text{Gamma}(\mu = 1.3, \phi=1) \\
    X_4 \sim \text{Gamma}(\mu = 1.3, \phi=1) \\
    X_5 \sim \text{Gamma}(\mu = 1.3, \phi=1) \\
    X_6 \sim \text{Bernoulli}(p=0.5) \\
    X_7 \sim \text{Bernoulli}(p=0.5) \\
    X_8 \sim \text{Bernoulli}(p=0.5) \\
    X_9 \sim \text{Bernoulli}(p=0.5) \\
    X_{10} \sim \text{Bernoulli}(p=0.5) \\
    T \sim \text{Bernoulli}(-0.3 + 0.1 X_1 + 0.2 X_2 + 0.5  X_3 -0.2  X_4 + 1 X_5 \\
    + 0.3 X_6-0.4  X_7 + 0.7 X_8 - 0.1  X_9 + 0.9 X_{10}) \\
    Y \mid \text{do(T)} \sim \mathcal{N}(2.5 - 5T, 1)
\end{aligned}  
\label{dgp:frugal-dgp5}
\end{equation}

The gaussian dependency matrix $R_5$ is given by
\begin{equation}
\boldsymbol{R}_5 = 
\begin{pmatrix}
1.0 & 0.3 & 0.4 & 0.5 & 0.1 & -0.2 & -0.7 & 0.5 & -0.4 & 0.5 \\
0.3 & 1.0 & -0.3 & 0.6 & -0.3 & 0.4 & -0.4 & 0.6 & 0.3 & 0.2 \\
0.4 & -0.3 & 1.0 & -0.5 & 0.2 & -0.1 & -0.1 & 0.0 & -0.4 & -0.4 \\
0.5 & 0.6 & -0.5 & 1.0 & -0.2 & -0.2 & -0.5 & 0.5 & 0.3 & 0.4 \\
0.1 & -0.3 & 0.2 & -0.2 & 1.0 & -0.1 & -0.1 & -0.5 & -0.6 & -0.2 \\
-0.2 & 0.4 & -0.1 & -0.2 & -0.2 & 1.0 & 0.0 & 0.4 & 0.2 & 0.5 \\
-0.7 & -0.4 & -0.1 & -0.5 & -0.1 & 0.0 & 1.0 & -0.5 & 0.4 & -0.4 \\
0.5 & 0.6 & 0.0 & 0.5 & -0.5 & 0.5 & -0.5 & 1.0 & 0.4 & 0.4 \\
-0.4 & 0.3 & -0.4 & 0.3 & -0.6 & 0.2 & 0.4 & 0.4 & 1.0 & 0.4 \\
0.5 & 0.2 & -0.4 & 0.4 & -0.2 & 0.5 & -0.4 & 0.4 & 0.4 & 1.0 \\
\end{pmatrix}
\label{eq:dgp5-rho}
\end{equation}

\paragraph{FrugalFlows Sim5 and FrugalFlows Sim5(u)} We train two simulators FrugalFlows Sim5 and FrugalFlows Sim5(u). To simulate unobserved confounding, we drop the covariates $X_3$ and $X_7$. The hyperparameters for both simulators are displayed in Table~\ref{tab:frugal-sim5-hyp}.

\begin{table}[htb]
    \caption{FrugalFlows hyperparameters for FrugalFlows Sim5 and FrugalFlows Sim5(u). \\}
    \label{tab:frugal-sim5-hyp}
    \centering
    \begin{tabular}{lcc}
    \toprule
    & FrugalFlows Sim5 & FrugalFlows Sim5(u) \\ \midrule
    Hyperparameter & \multicolumn{2}{c}{Value} \\ \midrule
      RQS knots & 1 & 1\\
      Flow layers & 1 & 4 \\
      Learning rate & 0.0081 & 0.0043 \\
      Network Depth & 50 & 77 \\
      Network Width & 22 & 39\\
      Causal model (CM) & Location translation & Location translation \\ 
      (CM) RQS knots & 5 & 9 \\
      (CM) Flow layers & 5 & 3\\
      (CM) Network Depth & 31 & 56\\
      (CM) Network Width & 1 & 36\\
    \bottomrule
    \end{tabular}
\end{table}

\paragraph{Prior: FrugalFlows Sim5} The prior is given by
\begin{equation}
    \begin{aligned}
        \text{Prior}(\tau) \sim U[-20.0, 20.0] \\ 
    \end{aligned}
\end{equation}

\paragraph{Prior: FrugalParam Sim5(u)} The prior is given by
\begin{equation}
    \begin{aligned}
        \text{Prior}(\tau) \sim U[-20.0, 20.0] \\ 
        \text{Prior}(\rho) \sim U[-1.0, 1.0]
    \end{aligned}
\end{equation}

\begin{figure}[ht]
\centering
\begin{minipage}{0.49\textwidth}
    \centering
    \includegraphics[scale=0.4]{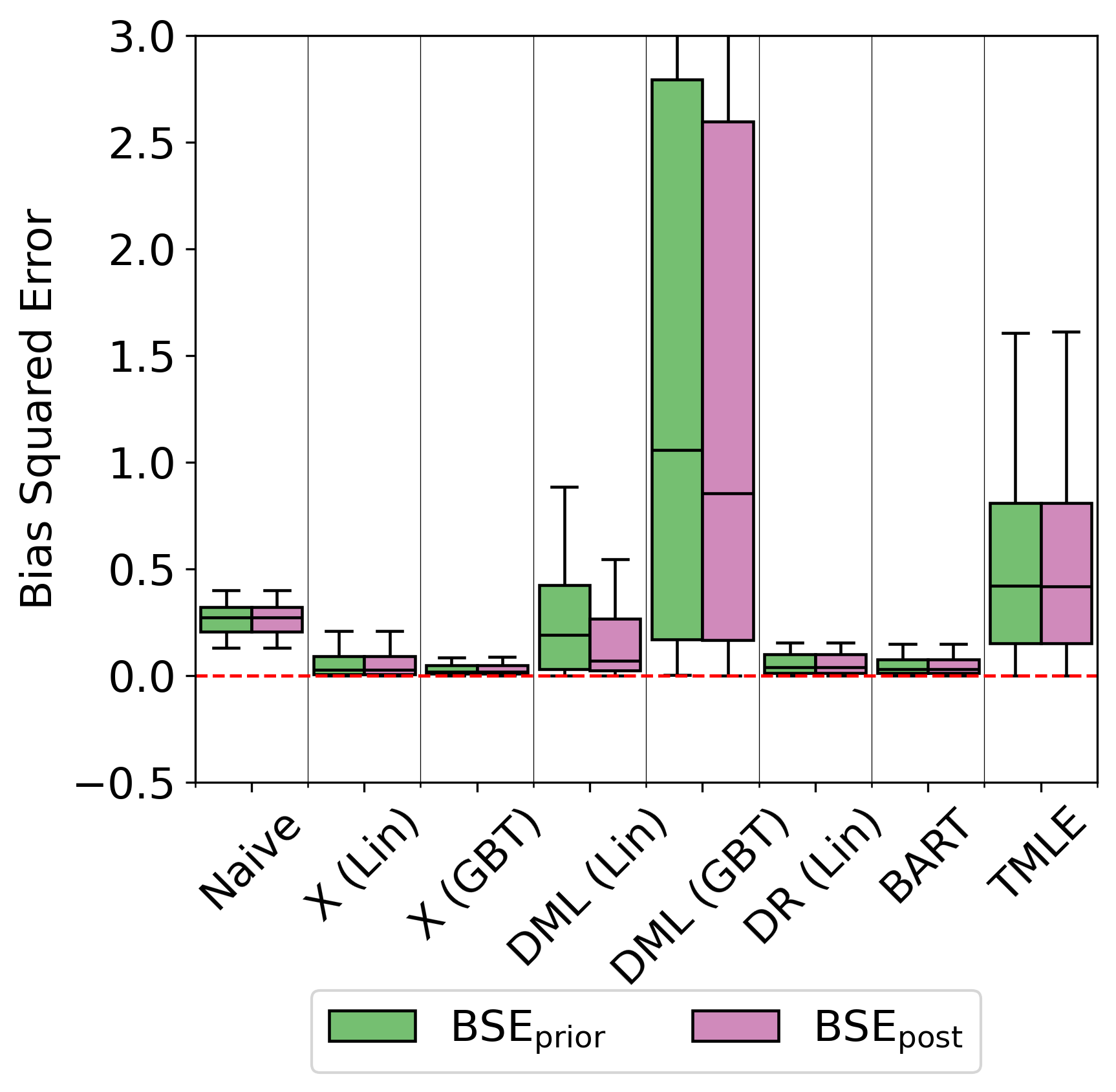}
\end{minipage}%
\begin{minipage}{0.47\textwidth}
    \centering
        \begin{tabular}{lcc}
        \toprule
        & Prior & Posterior \\ \midrule
        Diff. of Means & $0.264$ & $0.264$ \\
        X (Lin) & $0.065$ & $0.065$ \\
        X (GBT) & $0.040$ & $0.040$ \\
        DML (Lin) & $0.319$ & $0.178$ \\
        DML (GBT) & $2.173$ & $1.876$ \\
        DR (Lin) & $1249.365$ & $1249.374$ \\
        BART & $0.053$ & $0.053$ \\
        TMLE & $0.790$ & $0.789$\\
        \bottomrule
        \end{tabular}
\end{minipage}
\caption{BSE for DGP: Frugal DGP5 and Simulator: FrugalFlows Sim5.}
\label{fig:results-fp-dgp5}
\end{figure}

\begin{figure}[ht]
\centering
\begin{minipage}{0.49\textwidth}
    \centering
    \includegraphics[scale=0.4]{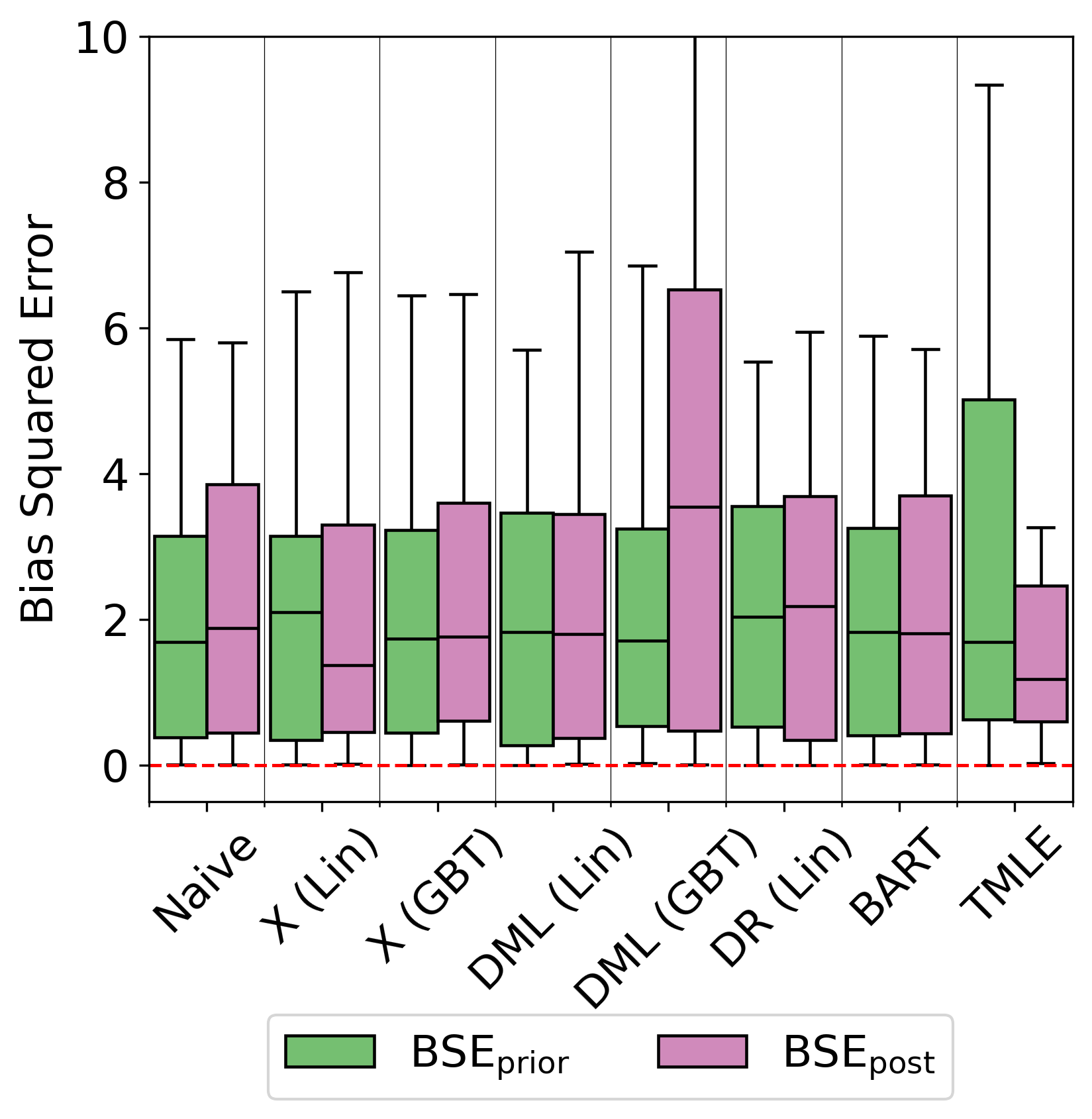}
\end{minipage}%
\begin{minipage}{0.47\textwidth}
    \centering
        \begin{tabular}{lcc}
        \toprule
        & Prior & Posterior \\ \midrule
        Diff. of Means & $1.955$ & $2.249$ \\
        X (Lin) & $2.157$ & $2.034$ \\
        X (GBT) & $2.033$ & $2.224$ \\
        DML (Lin) & $2.179$ & $2.195$ \\
        DML (GBT) & $2.880$ & $4.193$ \\
        DR (Lin) & $4e5$ & $5e5$ \\
        BART & $2.013$ & $2.149$ \\
        TMLE & $3.171$ & $2.114$\\
        \bottomrule
        \end{tabular}
\end{minipage}
\caption{BSE for DGP: Frugal DGP5 and Simulator: FrugalFlows Sim5(u).}
\label{fig:results-fp-dgp5u}
\end{figure}

\paragraph{Evaluation} The BSE is displayed in Figures~\ref{fig:results-fp-dgp5} and ~\ref{fig:results-fp-dgp5u}. 

\begin{tcolorbox}[colback=gray!5!white, colframe=gray!75!black, title=\textbf{Key takeaway}]
As long as the simulator is able to approximate the source data distribution, computing the posterior is informative, i.e. we may be able to eliminate values of the {\knob}s that are incompatible with the source data distribution. We verified this is true for non-parametric DGPs and flexible, neural-network based simulators (FrugalFlows).  
\end{tcolorbox}

\section{Evaluating {\ourmethod} on real-world, observational datasets}
\label{app:sbice-real-dgps}
Apart from synthetic datasets, we also evaluated {\ourmethod} on real-world, observational datasets described in Appendix~\ref{sec:comparing-gen-real}. We used the Lalonde (RCT), the Lalonde (observational) dataset, Postgres and the Twins dataset which all have a ground-truth estimate of the ATE. This allows us to compute the bias squared error (BSE). For the real-world datasets, we utilized two simulators: Realcause and Frugalflows, we found that datasets generated by Realcause exhibited lower SWD values when compared to FrugalFlows and was automatically selected by {\ourmethod}. We focused on three different prior parameters: treatment effect $\tau$, degree of heterogeneity $\xi$, and overlap $\pi = P(T =1 \mid X)$. We measured the bias squared error for a number of estimators and found that the posterior BSE was often lower than the prior BSE suggesting that our methods are useful in eliminating {\knob} values that are inconsistent with the source data.  

\subsection{Lalonde (RCT)} 
\label{app:sbice-lalonde-rct}

This dataset has 8 covariates (a combination of discrete and continuous values). It is based on the randomized controlled trial that examines the impact of an employment program on the income levels of participants. There is no unobserved confounding, as this is an RCT. 

\paragraph{Simulators} We include the details of the two simulators that were trained on the Lalonde (RCT) data, named FrugalFLows Lalonde (RCT) and Realcause Lalonde (RCT). The hyperparameters used are shown in Tables~\ref{tab:lalonde-exp-ff-hyp} and~\ref{tab:lalonde-exp-rc-hyp}. To ensure that the scales of the generated datasets are the same across the two simulators, we re-scaled the source data distribution prior to running {\ourmethod}. 

\begin{table}[htb]
    \begin{minipage}{.5\linewidth}
        \centering
        \begin{tabular}{lc}
        \toprule
        Hyperparameter & Value \\ \midrule
          RQS knots & 20 \\
          Flow layers & 14 \\
          Learning rate & 0.001 \\
          Network Depth & 22 \\
          Network Width & 11 \\
          Causal model (CM) & Location translation \\ 
          (CM) RQS knots & 3 \\
          (CM) Flow layers & 13 \\
          (CM) Network Depth & 25 \\
          (CM) Network Width & 28 \\
        \bottomrule
        \end{tabular}    
        \caption{FrugalFlows Lalonde (RCT)}
        \label{tab:lalonde-exp-ff-hyp}
    \end{minipage}
    \begin{minipage}{.49\linewidth}
        \centering
        \begin{tabular}{lc}
        \toprule
        Hyperparameter & Value \\ \midrule
          Model & TARNet \\
          Distribution & Sigmoidal Flow \\
          Dimensions & 2 \\
          Base distribution & Normal \\
          Hidden layers & 64 \\
          Hidden dimensions & 256 \\ 
          Treatment Kernel & RBFKernel \\
          Outcome Kernel & RBFKernel \\
          Learning rate & 0.0068 \\
          Batch size & 512 \\
          Max Epocs & 1000 \\
          W transform & Standardize \\
          Y transform & Normalize \\ 
        \bottomrule
        \end{tabular}
        \caption{Realcause Lalonde (RCT)}
        \label{tab:lalonde-exp-rc-hyp}
    \end{minipage}
    \caption{Hyperparameters for simulators trained on the Lalonde (RCT) dataset.}
    \label{tab:lalonde-exp-hyp}
\end{table}

\paragraph{Prior} The prior distribution is as follows

\begin{equation}
    \begin{aligned}
        \text{Prior}(\tau, \xi, \pi \mid G_k=\text{RC1}) = \{U[-1.0, 1.0], U[0.0,1.0], U[0.0, 1.0]\} \\
        \text{Prior}(\tau,\rho \mid G_k=\text{FF}1) = \{U[-1.0, 1.0], U[-1.0, 2.0]\} 
    \end{aligned}
\end{equation}

\begin{figure}[ht]
\centering
\begin{minipage}{0.49\textwidth}
    \centering
    \includegraphics[scale=0.4]{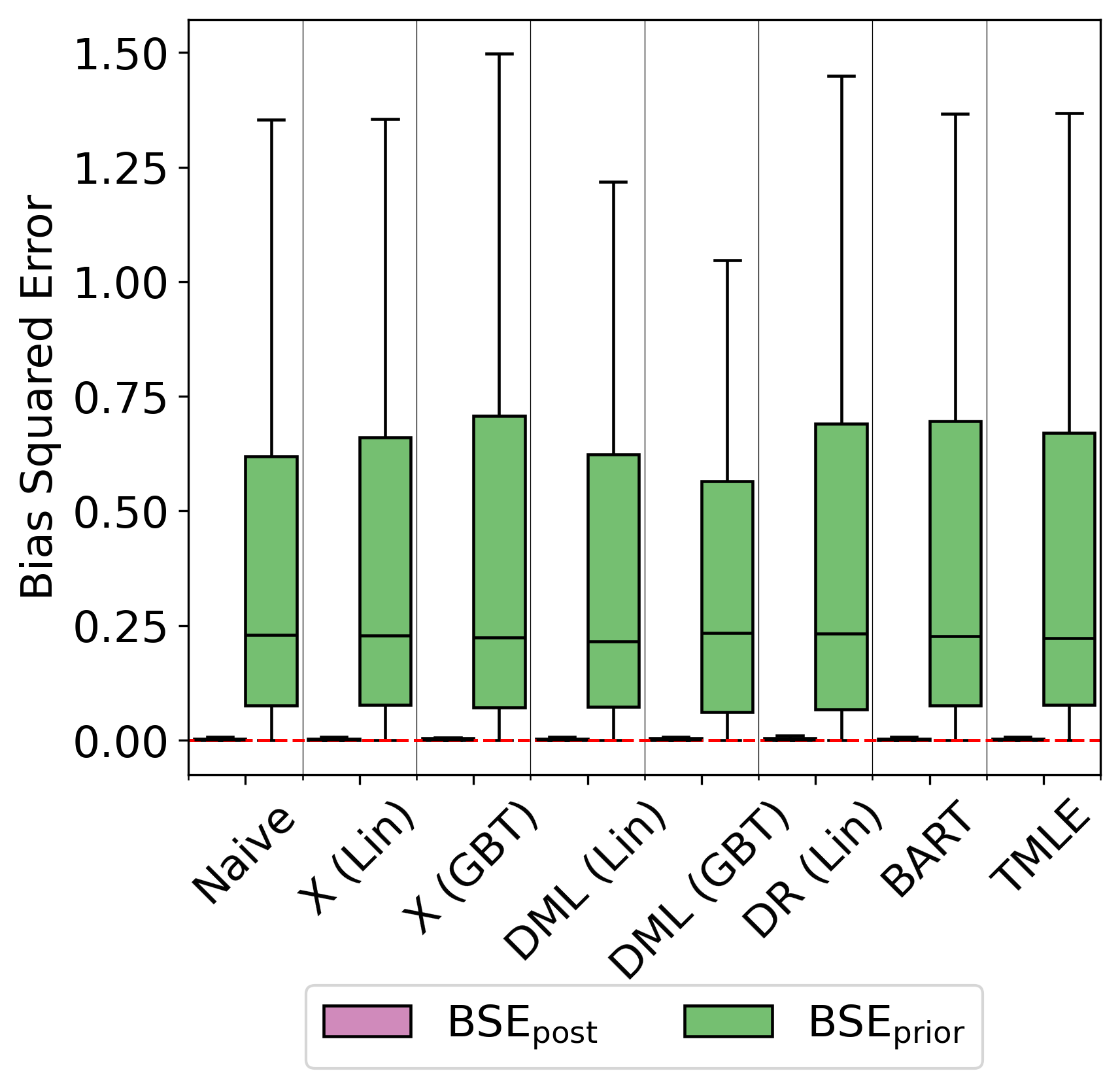}
\end{minipage}%
\begin{minipage}{0.47\textwidth}
    \centering
        \begin{tabular}{lcc}
        \toprule
        & Prior & Posterior \\ \midrule
        SWD & $0.51 \pm 0.28$ & $0.15\pm 0.01$   \\ \midrule
        \textbf{Mean BSE} & & \\
        X (Lin) & $0.373$ & $0.002$ \\
        X (GBT) & $0.396$ & $0.002$ \\
        DML (Lin) & $0.347$ & $0.002$ \\
        DML (GBT) & $0.400$ & $0.003$ \\
        DR (Lin) & $6e5$ & $160.43$ \\
        BART & $0.381$ & $0.002$ \\
        TMLE & $0.375$ & $0.002$\\
        \bottomrule
        \end{tabular}
\end{minipage}
\caption{SWD and Mean BSE of causal estimators for DGP: Lalonde (RCT) and Simulator: Realcause Lalonde (RCT).}
\label{fig:results-real-dgp1}
\end{figure}

\paragraph{Evaluation} Since this is an RCT, the difference of means is the ground-truth ATE $\tau^*$. We compute the mean SWD for the posterior and prior datasets as well as the BSE for a set of causal estimators and include the results in Figure~\ref{fig:results-real-dgp1}. We note that the model probability $P(G_k = \text{RC}) = 1.0$ for the last iteration of SMC-ABC. Note that the Doubly Robust estimator exhibits a large mean BSE value for both the posterior and the prior, primarily driven by outliers in the estimated ATE values (for certain datasets). We examined the median values and found them to be comparable to other estimators. 

\subsection{Lalonde} 
\label{app:sbice-lalonde-obs}

An observational counterpart to the RCT in Section~\ref{app:sbice-lalonde-rct} is the Population Survey of Income Dynamics (PSID) control sample. This dataset contains the treated units from the RCT and the control samples from the observational dataset. We use the same set of covariates, and use the ground-truth ATE as the value obtained in the RCT. 

\paragraph{Simulators} The hyperparameters for the FrugalFlows Lalonde and Realcause Lalonde simulators are shown in Table~\ref{tab:lalonde-obs-hyp}. 

\begin{table}[htb]
    \begin{minipage}{.5\linewidth}
        \centering
        \begin{tabular}{lc}
        \toprule
        Hyperparameter & Value \\ \midrule
          RQS knots & 1 \\
          Flow layers & 8 \\
          Learning rate & 0.0038 \\
          Network Depth & 2 \\
          Network Width & 6 \\
          Causal model (CM) & Location translation \\ 
          (CM) RQS knots & 9 \\
          (CM) Flow layers & 4 \\
          (CM) Network Depth & 3 \\
          (CM) Network Width & 6 \\
        \bottomrule
        \end{tabular}    
        \caption{FrugalFlows Lalonde}
        \label{tab:lalonde-obs-ff-hyp}
    \end{minipage}
    \begin{minipage}{.49\linewidth}
        \centering
        \begin{tabular}{lc}
        \toprule
        Hyperparameter & Value \\ \midrule
          Model & TARNet \\
          Distribution & Sigmoidal Flow \\
          Dimensions & 2 \\
          Base distribution & Uniform \\
          Hidden layers & 1 \\
          Hidden dimensions & 4 \\ 
          Treatment Kernel & RBFKernel \\
          Outcome Kernel & RBFKernel \\
          Learning rate & 0.01 \\
          Batch size & 25000 \\
          Max Epocs & 10000 \\
          W transform & Normalize \\
          Y transform & Normalize \\ 
        \bottomrule
        \end{tabular}
        \caption{Realcause Lalonde}
        \label{tab:lalonde-obs-rc-hyp}
    \end{minipage}
    \caption{Hyperparameters for simulators trained on the Lalonde dataset}
    \label{tab:lalonde-obs-hyp}
\end{table}

\paragraph{Prior} We set the prior as follows

\begin{equation}
    \begin{aligned}
        \text{Prior}(\tau, \xi, \pi \mid G_k=\text{RC2}) = \{U[-5.0, 5.0], U[0.0,1.0], U[0.0, 1.0]\} \\
        \text{Prior}(\tau,\rho \mid G_k=\text{FF}2) = \{U[-5.0, 5.0], U[-1.0, 2.0]\} 
    \end{aligned}
\end{equation}

\begin{figure}[ht]
\centering
\begin{minipage}{0.49\textwidth}
    \centering
    \includegraphics[scale=0.4]{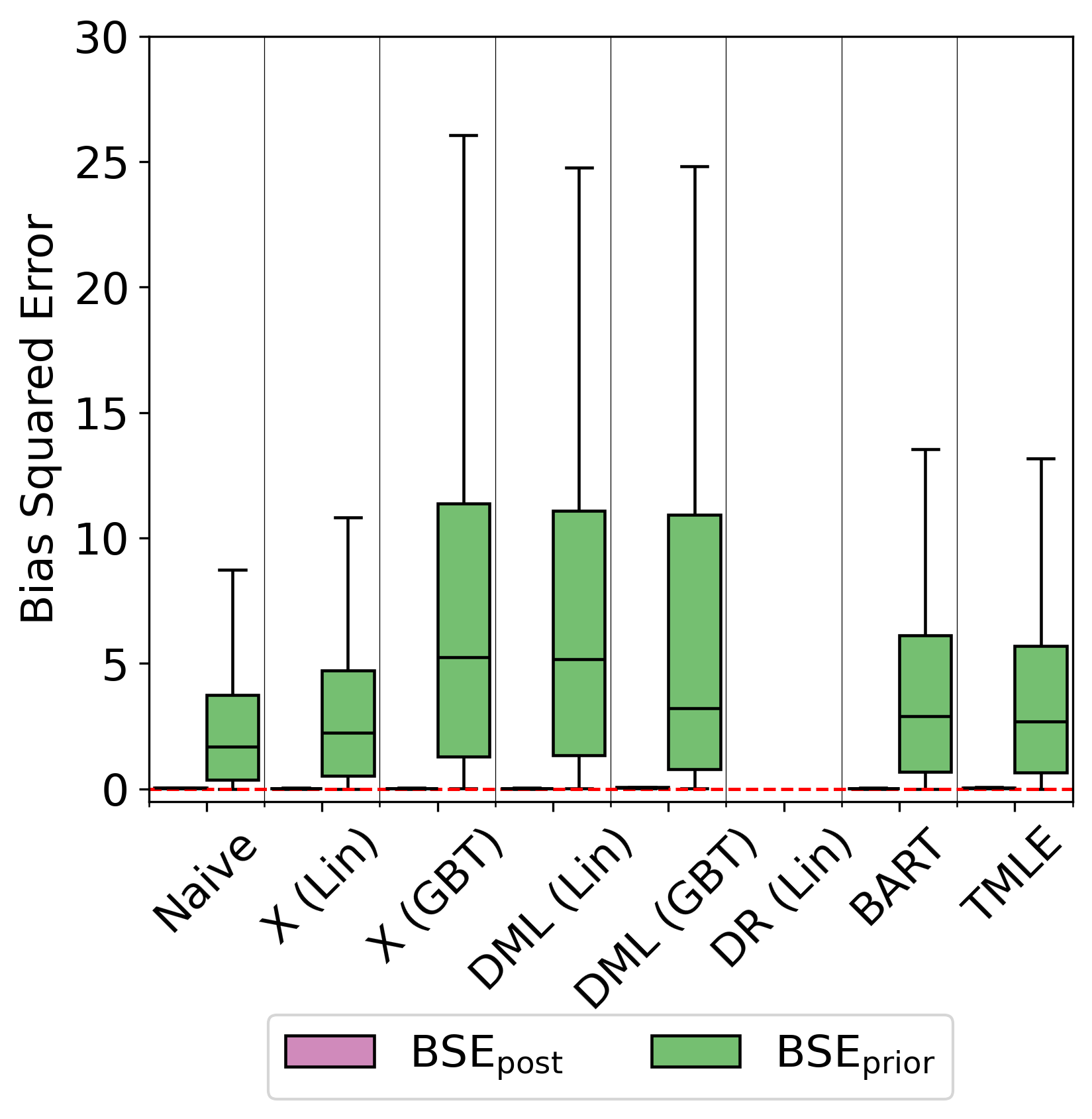}
\end{minipage}%
\begin{minipage}{0.47\textwidth}
    \centering
        \begin{tabular}{lcc}
        \toprule
        & Prior & Posterior \\ \midrule
        SWD & $1.87 \pm 1.20$ & $0.15\pm 0.01$   \\ \midrule
        \textbf{Mean BSE} & & \\
        X (Lin) & $3.520$ & $0.016$ \\
        X (GBT) & $8.580$ & $0.017$ \\
        DML (Lin) & $8.332$ & $0.022$ \\
        DML (GBT) & $6.193$ & $0.046$ \\
        DR (Lin) & $4e3$ & $4e3$ \\
        BART & $4.631$ & $0.017$ \\
        TMLE & $4.291$ & $0.035$\\
        \bottomrule
        \end{tabular}
\end{minipage}
\caption{SWD and Mean BSE of causal estimators for DGP: Lalonde and Simulator: Realcause Lalonde.}
\label{fig:results-real-dgp2}
\end{figure}

\paragraph{Evaluation} We include the bias squared error plot, the mean SWD and the BSE for each causal estimator in Figure~\ref{fig:results-real-dgp2}. 

\subsection{Postgres}
\label{app:sbice-postgres}

The Postgres dataset contains 8 covariates measuring the settings of a Postgres database. The treatment is a binary indicator of the index level and the outcome measures the run time for that setting.

\paragraph{Simulators} The hyperparameters for FrugalFlows Postgres and Realcause Postgres simulators are included in Table~\ref{tab:postgres-hyp}. 

\begin{table}[htb]
    \begin{minipage}{.5\linewidth}
        \centering
        \begin{tabular}{lc}
        \toprule
        Hyperparameter & Value \\ \midrule
          RQS knots & 45 \\
          Flow layers & 38 \\
          Learning rate & 0.0003 \\
          Network Depth & 2 \\
          Network Width & 14 \\
          Causal model (CM) & Location translation \\ 
          (CM) RQS knots & 46 \\
          (CM) Flow layers & 41 \\
          (CM) Network Depth & 23 \\
          (CM) Network Width & 48 \\
        \bottomrule
        \end{tabular}    
        \caption{FrugalFlows Postgres}
        \label{tab:postgres-ff-hyp}
    \end{minipage}
    \begin{minipage}{.49\linewidth}
        \centering
        \begin{tabular}{lc}
        \toprule
        Hyperparameter & Value \\ \midrule
          Model & TARNet \\
          Distribution & Sigmoidal Flow \\
          Dimensions & 3 \\
          Base distribution & Normal \\
          Hidden layers & 2 \\
          Hidden dimensions & 32 \\ 
          Treatment Kernel & RBFKernel \\
          Outcome Kernel & RBFKernel \\
          Learning rate & 0.007 \\
          Batch size & 16 \\
          Max Epocs & 1000 \\
          W transform & Standardize \\
          Y transform & Normalize \\ 
        \bottomrule
        \end{tabular}
        \caption{Realcause Postgres}
        \label{tab:postgres-rc-hyp}
    \end{minipage}
    \caption{Hyperparameters for simulators trained on the Postgres dataset.}
    \label{tab:postgres-hyp}
\end{table}

\paragraph{Prior} We set the prior as follows

\begin{equation}
    \begin{aligned}
        \text{Prior}(\tau, \xi, \pi \mid G_k=\text{Realcause}) = \{U[-5.0, 5.0], U[0.0,1.0], U[0.0, 1.0]\} \\
        \text{Prior}(\tau,\rho \mid G_k=\text{FrugalFlows}) = \{U[-5.0, 5.0], U[-1.0, 2.0]\} 
    \end{aligned}
\end{equation}

\begin{figure}[ht]
\centering
\begin{minipage}{0.49\textwidth}
    \centering
    \includegraphics[scale=0.4]{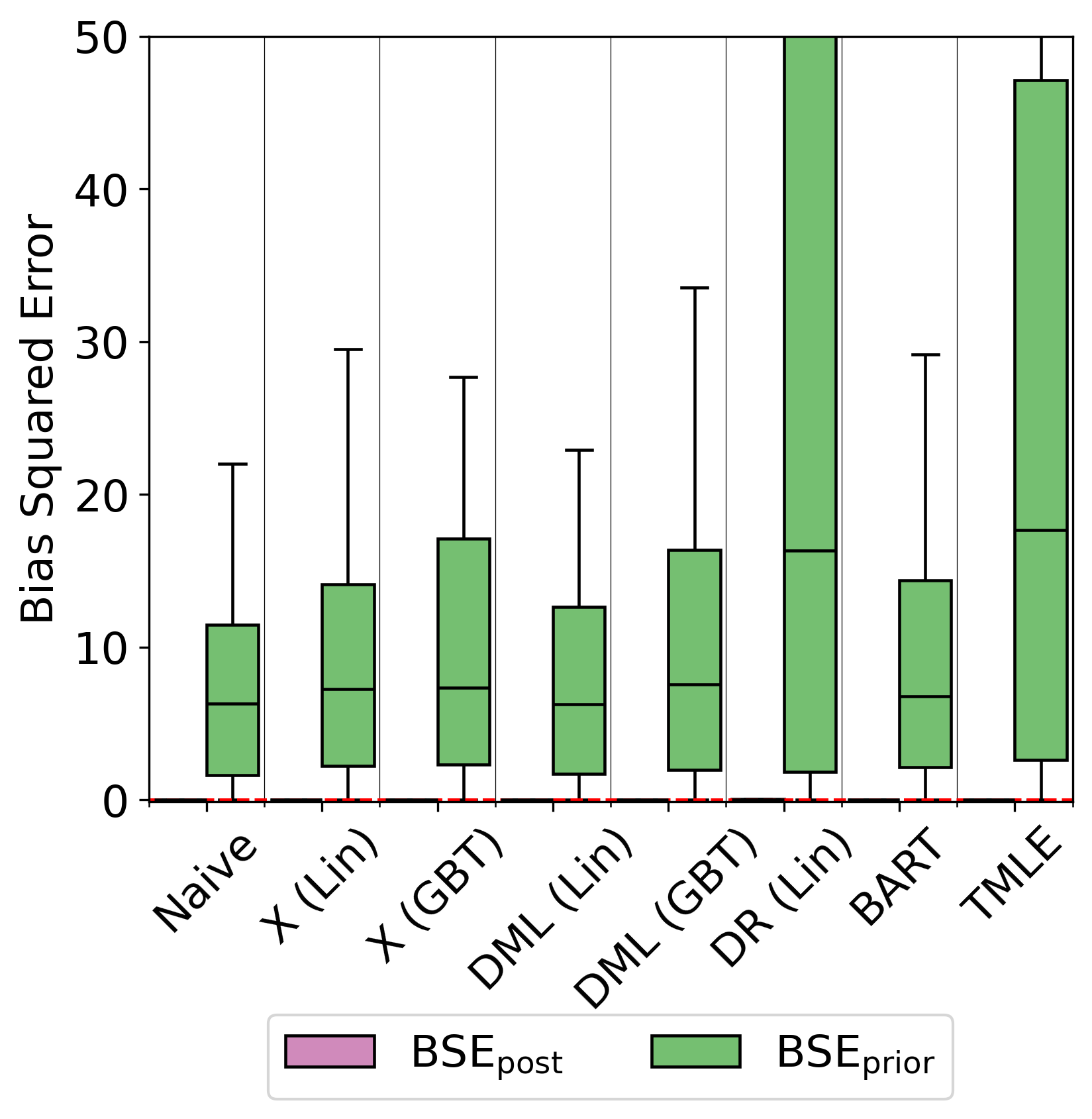}
\end{minipage}%
\begin{minipage}{0.47\textwidth}
    \centering
        \begin{tabular}{lcc}
        \toprule
        & Prior & Posterior \\ \midrule
        SWD & $1.62 \pm 1.03$ & $0.02\pm 0.001$   \\ \midrule
        \textbf{Mean BSE} & & \\
        X (Lin) & $9.130$ & $0.0003$ \\
        X (GBT) & $10.055$ & $0.0003$ \\
        DML (Lin) & $7.827$ & $0.0003$ \\
        DML (GBT) & $9.844$ & $0.0003$ \\
        DR (Lin) & $1e7$ & $590$ \\
        BART & $8.738$ & $0.0003$ \\
        TMLE & $27.045$ & $0.002$\\
        \bottomrule
        \end{tabular}
\end{minipage}
\caption{SWD and Mean BSE of causal estimators for DGP: Postgres and Simulator: Realcause Postgres.}
\label{fig:results-real-dgp3}
\end{figure}

\paragraph{Evaluation} We include the bias squared error plot, the mean SWD and the BSE for each causal estimator in Figure~\ref{fig:results-real-dgp3}. 

\subsection{Twins}
\label{app:sbice-twins}

We include another dataset with simulated treatment confounding (similar to the Postgres dataset)~\citep{louizos2017causal}. We find that the posterior distribution is closer to the source distribution as demonstrated by the lower BSE values. The experimental details are as listed below. 

\paragraph{Simulators} The hyperparameters for FrugalFlows Twins and Realcause Twins simulators are included in Table~\ref{tab:twins-hyp}. 

\begin{table}[htb]
    \begin{minipage}{.5\linewidth}
        \centering
        \begin{tabular}{lc}
        \toprule
        Hyperparameter & Value \\ \midrule
          RQS knots & 15 \\
          Flow layers & 26 \\
          Learning rate & 0.0055 \\
          Network Depth & 2 \\
          Network Width & 49 \\
          Causal model (CM) & Location translation \\ 
          (CM) RQS knots & 9 \\
          (CM) Flow layers & 43 \\
          (CM) Network Depth & 49 \\
          (CM) Network Width & 15 \\
        \bottomrule
        \end{tabular}    
        \caption{FrugalFlows Twins}
        \label{tab:twins-ff-hyp}
    \end{minipage}
    \begin{minipage}{.49\linewidth}
        \centering
        \begin{tabular}{lc}
        \toprule
        Hyperparameter & Value \\ \midrule
          Model & TARNet \\
          Distribution & Bernoulli \\
          Hidden layers & 1 \\
          Hidden dimensions & 64 \\ 
          Treatment Kernel & RBFKernel \\
          Outcome Kernel & RBFKernel \\
          Learning rate & 5e-5 \\
          Batch size & 256 \\
          Max Epocs & 10000 \\
          W transform & Normalize \\
          Y transform & Normalize \\ 
        \bottomrule
        \end{tabular}
        \caption{Realcause Twins}
        \label{tab:twins-rc-hyp}
    \end{minipage}
    \caption{Hyperparameters for simulators trained on the Twins dataset.}
    \label{tab:twins-hyp}
\end{table}

\paragraph{Prior} We set the prior as follows

\begin{equation}
    \begin{aligned}
        \text{Prior}(\tau, \xi, \pi \mid G_k=\text{Realcause}) = \{U[-3.0, 3.0], U[0.0,1.0], U[0.0, 1.0]\} \\
        \text{Prior}(\tau,\rho \mid G_k=\text{FrugalFlows}) = \{U[-3.0, 3.0], U[-1.0, 2.0]\} 
    \end{aligned}
\end{equation}

\begin{table}
    \centering
    \begin{tabular}{lcc}
    \toprule
    & Prior & Posterior \\ \midrule
    SWD & $1.01 \pm 0.62$ & $0.01\pm 0.00$   \\ \midrule
    \textbf{Mean BSE} & & \\
    X (Lin) & $0.044$ & $0.0003$ \\
    X (GBT) & $0.063$ & $0.0001$ \\
    DML (Lin) & $0.039$ & $0.0002$ \\
    DML (GBT) & $0.048$ & $0.0001$ \\
    DR (Lin) & $0.062$ & $0.002$ \\
    BART & $-$ & $-$ \\
    TMLE & $-$ & $-$\\
    \bottomrule
    \end{tabular}
\caption{SWD and Mean BSE of causal estimators for DGP: Twins and Simulator: Realcause Twins.}
\label{fig:results-real-dgp4}
\end{table} 

\paragraph{Evaluation} We include the mean SWD and the BSE for each causal estimator in Table~\ref{fig:results-real-dgp4}. We used the default settings of the estimators for BART and TMLE and found many errors for the generated datasets. For this reason, we omit their causal estimates and corresponding BSE values in the table.

\section{{\ourmethod} with alternate distance metrics}
\label{app:sbice-mmd}
We evaluate {\ourmethod} for the three real-world datasets using an alternate distance metric: maximum mean discrepancy (MMD). MMD is a kernel-based distance metric that measures the difference between two probability distributions by comparing their mean embeddings in a reproducing kernel Hilbert space (RKHS). MMD is based on the idea that if two distributions are equivalent, then, their moments match. While it is more appropriate for the mixed-type and high dimensional datasets, exact computation is of the order $O(n^2)$, when compared to the sliced-Wasserstein distance (SWD) which can be computed in $O(n \log n)$. We include the results of SBICE for three real-world datasets in this section. We find that the BSE for all estimators across both the posterior and the prior datasets is similar to that obtained using the SWD as seen in Appendix~\ref{app:sbice-real-dgps}. 

\subsection{Lalonde (RCT)} 
\label{app:sbice-lalonde-rct-mmd}

\begin{figure}[ht]
\centering
\begin{minipage}{0.49\textwidth}
    \centering
    \includegraphics[scale=0.4]{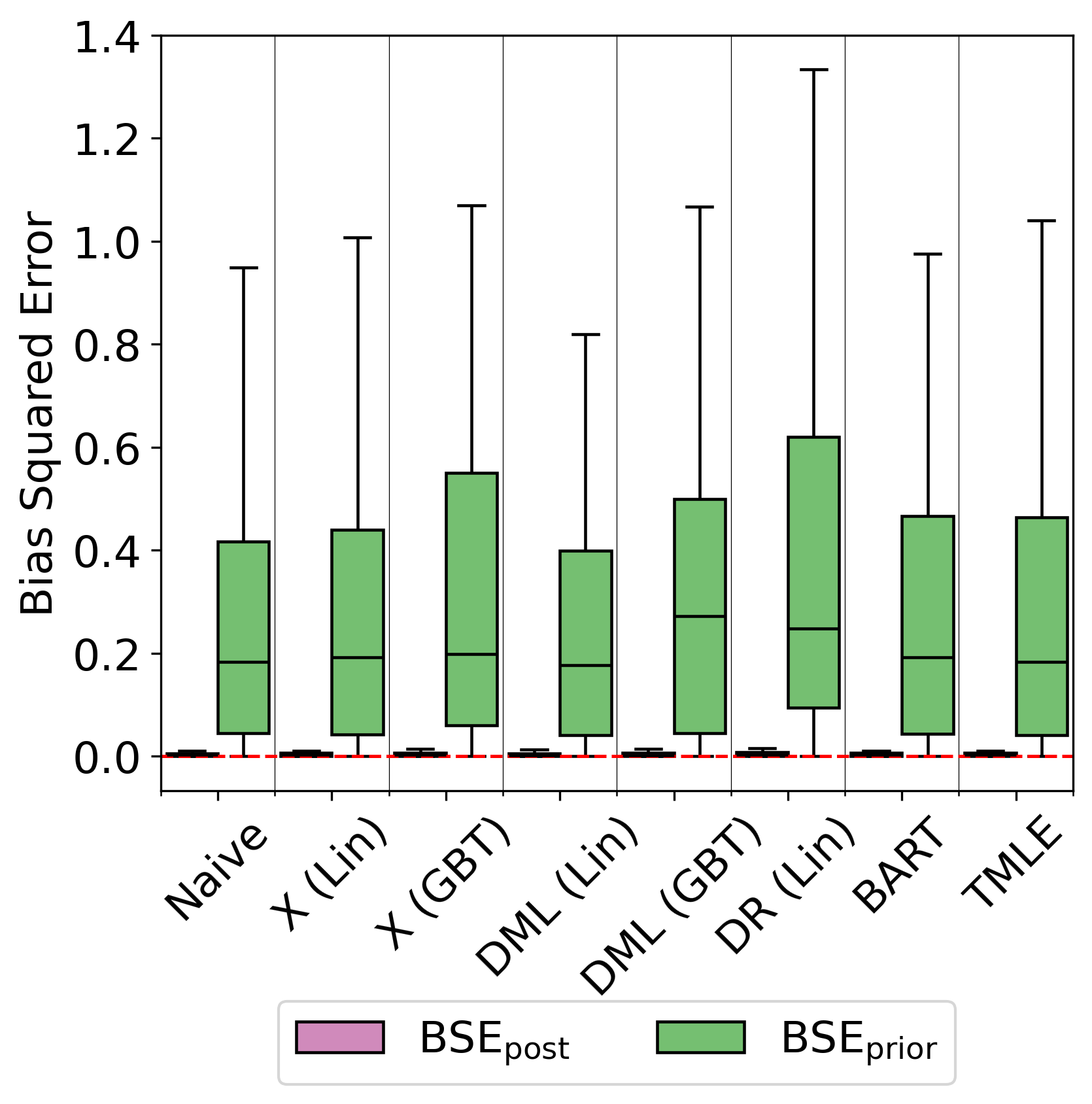}
\end{minipage}%
\begin{minipage}{0.47\textwidth}
    \centering
        \begin{tabular}{lcc}
        \toprule
        & Prior & Posterior \\ \midrule
        SWD & $0.50 \pm 0.28$ & $0.15\pm 0.02$   \\ 
        MMD & $0.028 \pm 0.016$ & $0.002 \pm 0.002$ \\
        \midrule
        \textbf{Mean BSE} & & \\
        X (Lin) & $0.191$ & $0.002$ \\
        X (GBT) & $0.198$ & $0.002$ \\
        DML (Lin) & $0.176$ & $0.001$ \\
        DML (GBT) & $0.271$ & $0.001$ \\
        DR (Lin) & $0.247$ & $0.002$ \\
        BART & $0.192$ & $0.002$ \\
        TMLE & $0.182$ & $0.002$\\
        \bottomrule
        \end{tabular}
\end{minipage}
\caption{SWD, MMD and Mean BSE of causal estimators for DGP: Lalonde (RCT) and Simulator: Realcause Lalonde (RCT) using MMD as the distance metric.}
\label{fig:results-real-dgp1-mmd}
\end{figure}

Figure~\ref{fig:results-real-dgp1-mmd} displays the BSE for each estimator with the posterior and prior datasets drawn using MMD as the distance metric. We also include the mean BSE values, mean SWD and mean MMD distances for all 50 datasets of the posterior and prior. As with the SWD metric, we find that the posterior BSE is less than the prior BSE across all estimators. 

\subsection{Lalonde}
\label{app:sbice-lalonde-obs-mmd}

\begin{figure}[ht]
\centering
\begin{minipage}{0.49\textwidth}
    \centering
    \includegraphics[scale=0.4]{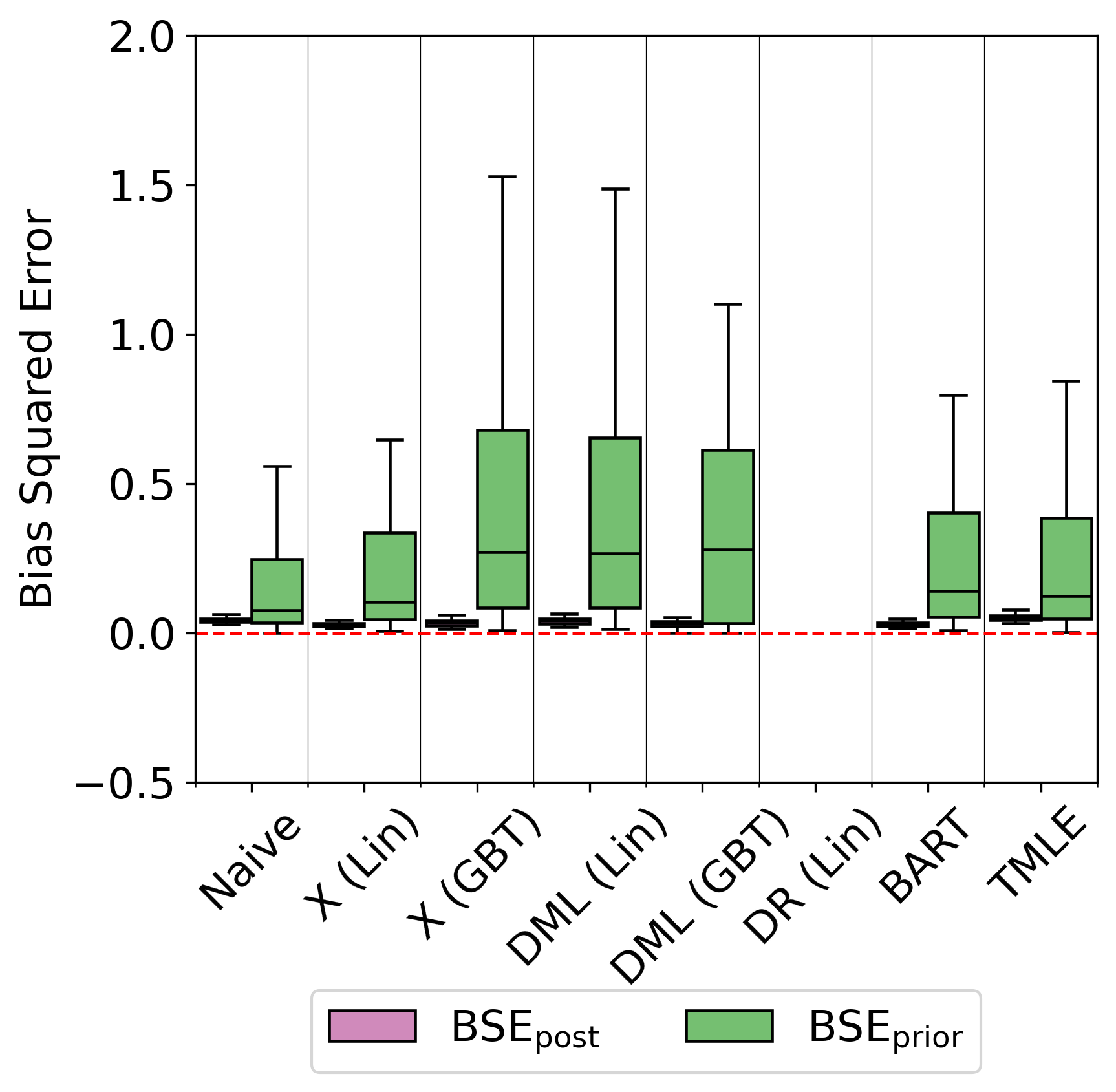}
\end{minipage}%
\begin{minipage}{0.47\textwidth}
    \centering
        \begin{tabular}{lcc}
        \toprule
        & Prior & Posterior \\ \midrule
        SWD & $0.43 \pm 0.20$ & $0.16\pm 0.01$   \\ 
        MMD & $0.159 \pm 0.055$ & $0.083 \pm 0.004$ \\
        \midrule
        \textbf{Mean BSE} & & \\
        X (Lin) & $0.187$ & $0.025$ \\
        X (GBT) & $0.441$ & $0.032$ \\
        DML (Lin) & $0.427$ & $0.039$ \\
        DML (GBT) & $0.498$ & $0.028$ \\
        DR (Lin) & $4e3$ & $4e3$ \\
        BART & $0.235$ & $0.027$ \\
        TMLE & $0.238$ & $0.049$\\
        \bottomrule
        \end{tabular}
\end{minipage}
\caption{SWD, MMD and Mean BSE of causal estimators for DGP: Lalonde and Simulator: Realcause Lalonde using MMD as the distance metric}
\label{fig:results-real-dgp2-mmd}
\end{figure}

Figure~\ref{fig:results-real-dgp2-mmd} displays the BSE for each estimator with the posterior and prior datasets drawn using MMD as the distance metric. 

\subsection{Postgres}
\label{app:sbice-postgres-mmd}

\begin{figure}[ht]
\centering
\begin{minipage}{0.49\textwidth}
    \centering
    \includegraphics[scale=0.4]{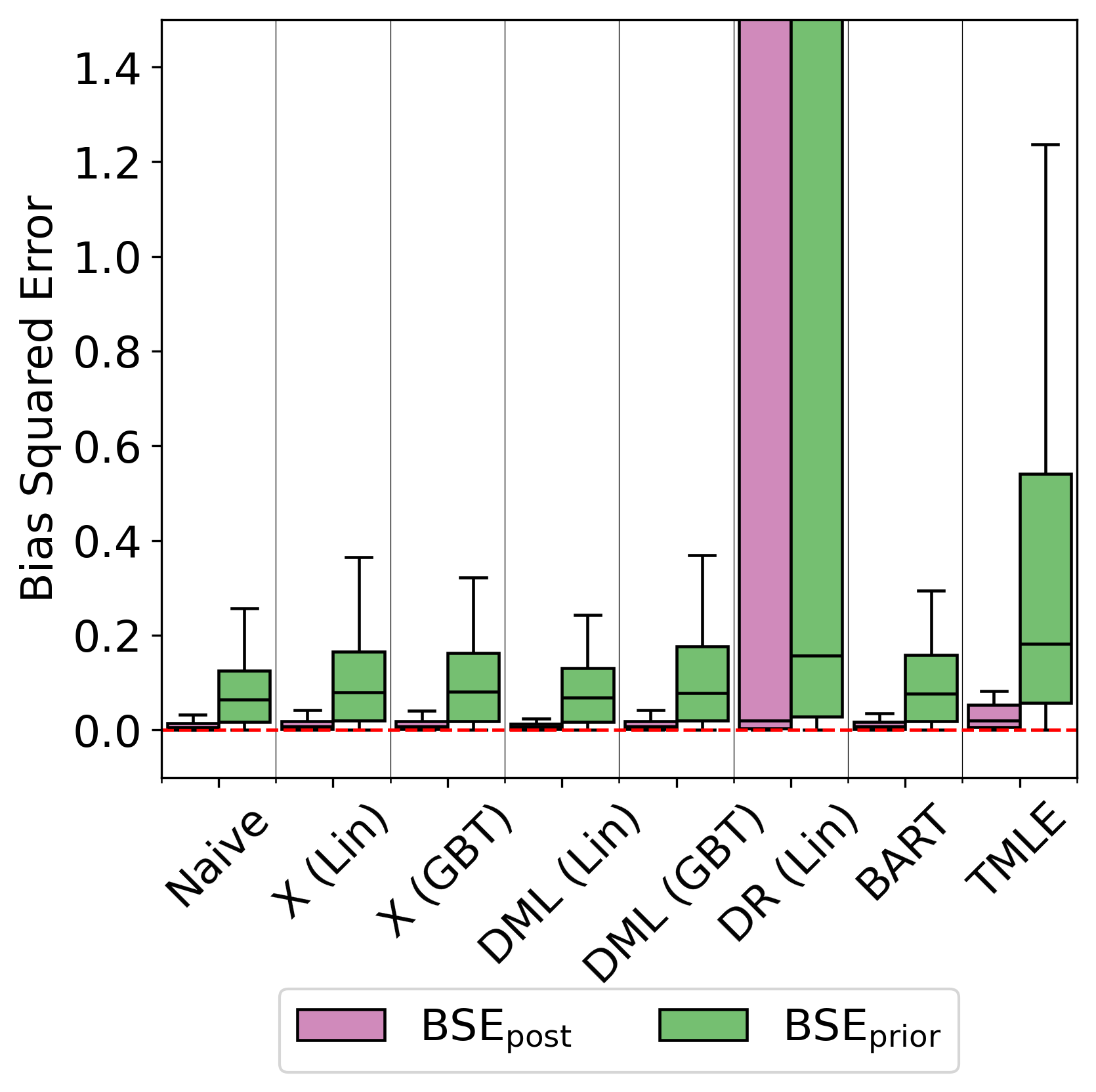}
\end{minipage}%
\begin{minipage}{0.47\textwidth}
    \centering
        \begin{tabular}{lcc}
        \toprule
        & Prior & Posterior \\ \midrule
        SWD & $0.12 \pm 0.07$ & $0.04\pm 0.02$   \\ 
        MMD & $0.0019 \pm 0.0014$ & $0.0005 \pm 0.0004$ \\ 
        \midrule
        \textbf{Mean BSE} & & \\
        X (Lin) & $0.100$ & $0.011$ \\
        X (GBT) & $0.109$ & $0.011$ \\
        DML (Lin) & $0.080$ & $0.009$ \\
        DML (GBT) & $0.113$ & $0.012$ \\
        DR (Lin) & $4e5$ & $2e5$ \\
        BART & $0.095$ & $0.011$ \\
        TMLE & $0.317$ & $0.039$\\
        \bottomrule
        \end{tabular}
\end{minipage}
\caption{SWD, MMD and Mean BSE of causal estimators for DGP: Postgres and Simulator: Realcause Postgres with MMD as the distance metric.}
\label{fig:results-real-dgp3-mmd}
\end{figure}

Figure~\ref{fig:results-real-dgp3-mmd} displays the BSE for each estimator with the posterior and prior datasets drawn using MMD as the distance metric for the Postgres dataset. We observe a similar trend as when using the SWD metric. 

\end{document}